\documentclass{article}

\usepackage{arxiv}

\usepackage{float}
\usepackage[utf8]{inputenc} 
\usepackage[T1]{fontenc}    
\usepackage{hyperref}       
\usepackage{url}            
\usepackage{booktabs}       
\usepackage{siunitx}        
\usepackage{amsfonts}       
\usepackage{nicefrac}       
\usepackage{microtype}      
\usepackage{graphicx}
\usepackage[numbers]{natbib}
\usepackage{doi}
\usepackage{amsmath}
\usepackage{cleveref}
\usepackage{subcaption}
\usepackage{algorithm}
\usepackage{algpseudocode}
\newcommand{\incfig}[2][]{%
  \IfFileExists{#2}{%
    \includegraphics[#1]{#2}%
  }{%
    \fbox{\begin{minipage}[c][0.5\textwidth][c]{\textwidth}\centering\small
      Placeholder (missing figure):\\\texttt{#2}
    \end{minipage}}%
  }%
}

\renewcommand{\headeright}{}
\renewcommand{\undertitle}{}

\title{E-PINNs: Epistemic Physics-Informed Neural Networks}

\author{
        \href{https://orcid.org/0009-0001-5361-3105}{\includegraphics[scale=0.06]{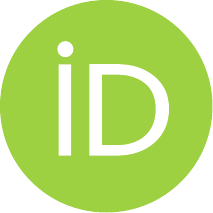}\hspace{1mm}Bruno Jacob}\\
	Pacific Northwest National Laboratory\\
	Richland, WA 99354 \\
	\texttt{bruno.jacob@pnnl.gov} \\
    \And
        Ashish S. Nair\\
	University of Notre Dame\\
	South Bend, IN 46637 \\
	\texttt{anair@nd.edu} \\
      \And  
        \href{https://orcid.org/0000-0002-6411-6198}{\includegraphics[scale=0.06]{Figures/orcid.pdf}\hspace{1mm}Amanda A. Howard}\\
	Pacific Northwest National Laboratory\\
	Richland, WA 99354 \\
	\texttt{amanda.howard@pnnl.gov} \\
     \And  
        \href{https://orcid.org/0000-0003-1223-208X}{\includegraphics[scale=0.06]{Figures/orcid.pdf}\hspace{1mm}Jan Drgona}\\
	Pacific Northwest National Laboratory\\
	Richland, WA 99354 \\
	\texttt{jan.drgona@pnnl.gov} \\
     \And  
        \href{https://orcid.org/0000-0002-9928-5637}{\includegraphics[scale=0.06]{Figures/orcid.pdf}\hspace{1mm}Panos Stinis}\\
	Pacific Northwest National Laboratory\\
	Richland, WA 99354 \\
	\texttt{panagiotis.stinis@pnnl.gov} \\
}

\begin{document}

\maketitle
\begin{abstract}
Physics-informed neural networks (PINNs) have demonstrated promise as a framework for solving forward and inverse problems involving partial differential equations. Despite recent progress in the field, it remains challenging to quantify uncertainty in these networks. While techniques such as Bayesian PINNs (B-PINNs) provide a principled approach to capturing epistemic uncertainty through Bayesian inference, they can be computationally expensive for large-scale applications. In this work, we propose Epistemic Physics-Informed Neural Networks (E-PINNs), a framework that uses a small network, the \emph{epinet}, to efficiently quantify epistemic uncertainty in PINNs. The proposed approach works as an add-on to existing, pre-trained PINNs with a small computational overhead. We demonstrate the applicability of the proposed framework in various test cases and compare the results with B-PINNs using Hamiltonian Monte Carlo (HMC) posterior estimation and dropout-equipped PINNs (Dropout-PINNs). In our experiments, E-PINNs achieve calibrated coverage with competitive sharpness at substantially lower cost. We demonstrate that when B-PINNs produce narrower bands, they under-cover in our tests. E-PINNs also show better calibration than Dropout-PINNs in these examples, indicating a favorable accuracy-efficiency trade-off.
\end{abstract}

\keywords{Physics-informed neural networks \and Uncertainty quantification \and Epistemic neural networks}

\section{Introduction}

Uncertainty quantification is a critical component in the deployment of machine learning models for scientific and engineering applications. Physics-Informed Neural Networks (PINNs) have shown promise in solving partial differential equations (PDEs) by incorporating physical laws directly into the learning process \cite{raissi2019physics}. PINNs have been successfully applied to a diverse range of scientific and engineering problems, including fluid dynamics and turbulence \cite{jin2021nsfnets, cai2021physics, hanrahan2023studying, patel2024turbulence}, heat transfer \cite{laubscher2021simulation, xu2023physics, cai2021bphysics,jalili2024physics}, biomedical applications \cite{sahli2020physics, kissas2020machine, sel2023physics}, electromagnetic systems \cite{khan2022physics, baldan2023physics}, among many other applications. When employing neural networks to physics-informed tasks, it is crucial to ensure that the models are not only accurate but also provide insights into the confidence of their predictions. However, traditional PINNs do not inherently quantify uncertainty \cite{yang2021b}.

Uncertainty is typically classified into two types: aleatoric or epistemic uncertainty. Aleatoric refers to the uncertainty inherent in the data, often due to noise or stochasticity in the physical process being modeled. This type of uncertainty cannot be reduced by collecting more data. In contrast, epistemic uncertainty arises due to limited data or lack of knowledge about the model itself and can be reduced with more data or a better model \cite{der2009aleatory, hullermeier2021aleatoric}. Understanding and distinguishing between these types of uncertainty is crucial for robust modeling, particularly in scientific applications where decision-making based on model predictions can have significant consequences.

Classical uncertainty quantification techniques for parametric PDEs, such as polynomial chaos expansion (PCE) \cite{kontolati2022survey, sharma2024physics}, have been widely used for forward propagation of parametric uncertainty and surrogate modeling. These methods expand solutions in orthogonal polynomials of random input parameters and, with sparsity and dimension reduction, can address high-dimensional inputs more efficiently. Recent developments have incorporated physics constraints into PCE-based surrogates \cite{sharma2024physics}, reducing model evaluations and promoting physically consistent predictions. These approaches typically assume sufficiently smooth responses and a prescribed set of random input parameters. By design, PCE propagates parametric uncertainty through known PDE operators; it does not address epistemic uncertainty in learned PDE surrogates that arises from limited data and model capacity.

The field of uncertainty quantification in deep learning has seen substantial growth in recent years. Bayesian Neural Networks (BNNs) provide a principled approach to uncertainty quantification by placing prior distributions over model parameters and inferring posterior distributions through Bayesian inference \cite{mackay1992practical, graves2011practical, neal2012bayesian,OLIVIER2021114079}. However, they often face scalability challenges when applied to complex models or high-dimensional data \cite{blundell2015weight}.

Ensemble methods represent another prominent approach to uncertainty quantification, where multiple models are trained, and their predictions aggregated to estimate uncertainty \cite{lakshminarayanan2017simple}. While effective in practice, these methods can be computationally expensive as they require training multiple independent models. Deep ensembles, in particular, have demonstrated strong performance in capturing uncertainty by training neural networks with different random initializations \cite{fort2019deep}. However, previous studies have shown that naively pooling ensemble predictions can sometimes lead to under-confidence in the model's predictions \cite{ashukha2020pitfalls}.

As an alternative to the computationally intensive Bayesian and ensemble methods, dropout methods offer a computationally lighter approach for uncertainty quantification. Dropout approximates Bayesian inference through random deactivation of neurons during both training and inference, effectively sampling from a posterior distribution over the network parameters without explicitly performing Bayesian sampling \cite{hinton2012improving,gal2016dropout,gal2017concrete,li2017dropout}. This method has been widely adopted due to its simplicity and effectiveness, although its theoretical underpinnings and optimal configuration remain areas of active research \cite{hiraoka2021dropout}.

In the context of scientific machine learning, recent approaches have focused on integrating physical principles into the uncertainty quantification process. Physics-informed Bayesian Neural Networks have been proposed to combine the expressiveness of neural networks with the physical constraints imposed by governing equations \cite{yang2020physics, zou2024neuraluq}. These methods aim to respect physical laws while providing probabilistic predictions, though they often face challenges related to the balance between data-driven learning and physical consistency. For PINNs \cite{raissi2019physics} specifically, several uncertainty quantification methods have been developed. Bayesian PINNs (B-PINNs) incorporate prior knowledge about physical systems while quantifying both aleatoric and epistemic uncertainties through a principled Bayesian treatment \cite{yang2021b}. However, they often struggle with scalability due to the high computational cost of sampling from the posterior distribution over the model parameters, and may suffer from high variance in the predictive distribution, particularly in high-dimensional spaces or when data is sparse \cite{zhang2019quantifying}.

Dropout has also been applied to PINNs (Dropout-PINNs), providing a more computationally efficient alternative to B-PINNs \cite{zhang2019quantifying, yang2021b}. However, the choice of dropout rate significantly impacts the quality of uncertainty quantification, which may limit the applicability of Dropout-PINNs in scenarios involving complex physical phenomena or sparse training data \cite{yang2021b}.

In the broader context of operator learning, uncertainty quantification for neural operators, including deep operator networks (DeepONets) \cite{lu2019deeponet}, has gained attention. These models aim to learn mappings between function spaces. Approaches such as randomized prior methods \cite{yang2022scalable}, ensemble methods \cite{pensoneault2025uncertainty}, and Bayesian formulations \cite{garg2023vb, zhang2024bayesian} have been proposed to quantify uncertainty in operator learning settings. These methods are particularly important when the input-output mappings are complex or when dealing with noisy or incomplete data.

A recent promising direction involves combining conformal prediction methods with PINNs \cite{podina2024conformalized} and DeepONets \cite{moya2025conformalized}. Conformal methods construct distribution-free uncertainty intervals that offer rigorous finite-sample coverage guarantees without relying on Bayesian assumptions or computationally intensive sampling. For example, conformalized DeepONets utilize split conformal prediction to generate robust confidence intervals directly from neural operator predictions \cite{moya2025conformalized}. However, both approaches inherit fundamental conformal prediction limitations: they require representative calibration datasets that match future test distributions, and they sacrifice some training data for the calibration process \cite{angelopoulos2020uncertainty}.

\subsection{Contributions and Novelty of the Present Work}

In the present work, we propose epistemic physics-informed neural networks (E-PINNs), a novel extension of the traditional PINN framework designed to quantify epistemic uncertainty. Our approach builds upon the concept of epistemic neural networks \cite{osband2023epistemic}, which provides a flexible framework for modeling uncertainty by introducing a source of randomness into the model parameters or inputs. We provide comprehensive empirical evaluations comparing E-PINNs with B-PINNs and Dropout-PINNs on forward initial-boundary value problems and an inverse problem. Additionally, we conduct ablation studies investigating the sensitivity of E-PINNs to various hyperparameters including network size, number of training epochs and magnitude of noise. These analyses provide insights into the robustness and practical applicability of our approach.

The key innovation of our E-PINN framework lies in its architecture: we augment a base PINN with an epinet, a neural network that takes random perturbations alongside features from the base model as input. This design offers several advantages over existing methods. First, E-PINNs maintain computational efficiency even for complex models; unlike B-PINNs, which face scalability challenges when sampling from high-dimensional posterior distributions. Second, unlike dropout-based methods that randomly deactivate neurons throughout the network, our approach introduces uncertainty through a dedicated architectural component that can be trained separately from the base model, thus enabling uncertainty quantification as a post-processing step. This latter capability is particularly valuable for practitioners who already have trained PINNs and wish to add uncertainty quantification without retraining the entire model, while introducing only a small computational overhead. 

This work is organized as follows: In Section~2, we present three frameworks for uncertainty quantification in PINNs: B-PINNs, Dropout-PINNs and E-PINNs. In Section~3, we report results grouped into forward problems, inverse problems, and an ablation study. Finally, a summary of the findings is presented in Section~4.

\section{Methods}
In this section, we detail the methodologies employed for uncertainty quantification in PINNs. We explore three primary approaches: B-PINNs, Dropout-PINNs and E-PINNs. Each method offers distinct mechanisms for capturing and quantifying uncertainty in model predictions.

In all methods, we consider a general PDE of the form
\begin{subequations}
\label{eq:pde_general}
\begin{align}
\mathcal{N}_x(u; \boldsymbol{\lambda}) &= f(x), \quad x \in \Omega, \label{eq:pde_general_pde} \\
\mathcal{B}_x(u; \boldsymbol{\lambda}) &= b(x), \quad x \in \partial \Omega, \label{eq:pde_general_bc}
\end{align}
\end{subequations}

where $\mathcal{N}_x$ represents a general differential operator, $\mathcal{B}_x$ denotes a general boundary operator, $u(x)$ is the solution to the PDE, $\boldsymbol{\lambda}$ is the (possibly vector-valued) set of PDE parameters, $f(x)$ is a source term, and $b(x)$ specifies the boundary conditions. The domain $\Omega \subset \mathbb{R}^d$ is a $d$-dimensional spatial domain, and $\partial \Omega$ denotes its boundary.

\subsection{Bayesian physics-informed neural networks}

B-PINNs \cite{yang2021b} extend the traditional PINN framework by incorporating BNNs \cite{mackay1992practical, neal2012bayesian} to quantify uncertainty in the model predictions. This approach treats the neural network parameters as random variables with specified prior distributions, allowing for the estimation of posterior distributions given observed data and physical constraints \cite{yang2021b,meng2021multi, lin2022multi,pensoneault2024efficient}.

In this framework, the solution $u(x)$ is approximated by a surrogate model $u_\theta(x)$, where $\theta$ denotes the neural network parameters, with a prior $P(\theta)$. Following the original B-PINN formulation, we 
evaluate the physics operators on the surrogate, i.e., $\mathcal{N}_x(u_\theta(x);\boldsymbol{\lambda})$ and $\mathcal{B}_x(u_\theta(x);\boldsymbol{\lambda})$, and compare them to (noisy) measurements when available. For readability we occasionally use residual shorthands 
$r_\theta(x) = \mathcal{N}_x(u_\theta(x);\boldsymbol{\lambda})-f(x)$ and $g_\theta(x) = \mathcal{B}_x(u_\theta(x);\boldsymbol{\lambda})-b(x)$.

We assume a dataset $\mathcal{D}$ containing independently distributed noisy measurements of $u$, $f$, and $b$, denoted as $\mathcal{D} = \mathcal{D}_u \cup \mathcal{D}_f \cup \mathcal{D}_b$, where
\begin{subequations}
\label{eq:datasets}
\begin{alignat}{2}
    \mathcal{D}_u &= \{(x_i^u, \overline{u}_i)\}_{i=1}^{N_u}, &\quad \overline{u}_i &= u(x_i^u) + \epsilon_i^u, \label{eq:datasets_u} \\
    \mathcal{D}_f &= \{(x_i^f, \overline{f}_i)\}_{i=1}^{N_f}, &\quad \overline{f}_i &= f(x_i^f) + \epsilon_i^f, \label{eq:datasets_f} \\
    \mathcal{D}_b &= \{(x_i^b, \overline{b}_i)\}_{i=1}^{N_b}, &\quad \overline{b}_i &= b(x_i^b) + \epsilon_i^b, \label{eq:datasets_b}
\end{alignat}
\end{subequations}
and $N_u, N_f, N_b$ denote the number of measurements, and $\epsilon_i^u, \epsilon_i^f, \epsilon_i^b$ are independent Gaussian noises with zero mean and known standard deviations $\sigma_i^u, \sigma_i^f, \sigma_i^b$, i.e.,  $\epsilon_i^u \sim \mathcal{N}(0, (\sigma_i^u)^2)$, $\epsilon_i^f \sim \mathcal{N}(0, (\sigma_i^f)^2)$, and $\epsilon_i^b \sim \mathcal{N}(0, (\sigma_i^b)^2)$.

In general, with independent Gaussian noise on the available sensors, the unnormalized likelihood factorizes as
\begin{equation}
    P(\mathcal{D}\mid\theta) \;\propto\; P(\mathcal{D}_u\mid\theta)\,P(\mathcal{D}_f\mid\theta)\,P(\mathcal{D}_b\mid\theta),
\end{equation}
with
\begin{subequations}
\label{eq:log_likelihoods}
\begin{align}
    \ln P(\mathcal{D}_u\mid\theta) &\;=\; -\frac{1}{2}\sum_{i=1}^{N_u} \frac{\big(u_\theta(x_i^u)-\overline u_i\big)^2}{(\sigma_i^u)^2} + \text{const}, \label{eq:log_likelihoods_u} \\
    \ln P(\mathcal{D}_f\mid\theta) &\;=\; -\frac{1}{2}\sum_{i=1}^{N_f} \frac{\big(\mathcal{N}_x(u_\theta(x_i^f);\boldsymbol{\lambda})-\overline f_i\big)^2}{(\sigma_i^f)^2} + \text{const}, \label{eq:log_likelihoods_f} \\
    \ln P(\mathcal{D}_b\mid\theta) &\;=\; -\frac{1}{2}\sum_{i=1}^{N_b} \frac{\big(\mathcal{B}_x(u_\theta(x_i^b);\boldsymbol{\lambda})-\overline b_i\big)^2}{(\sigma_i^b)^2} + \text{const}. \label{eq:log_likelihoods_b}
\end{align}
\end{subequations}
If a subset of sensors is unavailable (e.g., no $f$ measurements), the corresponding factor is omitted. In the present work, we add observation noise only to interior $u$ measurements; we do not add noise to $f$ or to boundary values when generating data. In our B-PINN experiments all available channels are modeled with Gaussian likelihoods: the $u$-data likelihood uses the known noise level, and the physics and boundary residual channels use small fixed noise levels for HMC stability. Unless otherwise noted, we set this residual noise standard deviation to 0.01. We report B-PINN results for the cases indicated in Section~3.

With a prior $P(\theta)$ on network parameters and the data likelihood $P(\mathcal{D}\mid\theta)$ defined above, Bayes' theorem gives the posterior:
\begin{equation}\label{eq:posterior}
    P(\theta \mid \mathcal{D}) \propto P(\mathcal{D} \mid \theta)\,P(\theta),
\end{equation}
where $P(\mathcal{D})$ denotes the marginal likelihood, often intractable to compute directly. For inverse problems with unknown PDE parameters $\boldsymbol{\lambda}$, we place a prior $P(\boldsymbol{\lambda})$ and infer the joint posterior
\begin{equation}\label{eq:joint_posterior}
    P(\theta,\boldsymbol{\lambda}\mid \mathcal{D})\;\propto\; P(\mathcal{D}\mid \theta,\boldsymbol{\lambda})\,P(\theta)\,P(\boldsymbol{\lambda}).
\end{equation}
Sampling or approximation methods (e.g., Hamiltonian Monte Carlo, variational inference) can then be employed to approximate the posterior distribution. We draw weight samples $\theta^{(i)}\sim P(\theta\mid\mathcal{D})$ (or joint samples $(\theta^{(i)},\boldsymbol{\lambda}^{(i)})$) and compute \emph{posterior predictive} realizations $u_{\theta^{(i)}}(x)$.

This Bayesian formulation captures both aleatoric (through sensor noise) and epistemic (through the posterior over $\theta$) uncertainty. In our experiments we do not inject noise at prediction time; as a result, the reported predictive bands reflect \emph{epistemic} variability only, i.e., $u$ noise enters only through the likelihood during training. Given the low performance of B-PINN-VI reported by \cite{yang2021b}, in this work we will use only B-PINN-HMC when referring to B-PINNs in the tests and comparisons presented in Section~3.

\subsection{Dropout-based physics-informed neural networks}

Dropout-PINNs incorporate dropout layers within the neural network architecture as a means to quantify uncertainty \cite{zhang2019quantifying}. In general, Monte Carlo dropout can reflect aspects of total uncertainty; however, in this paper we report \emph{epistemic-only} uncertainty, i.e., the variability induced by stochastic dropout masks with no additive observation noise at prediction time. Dropout, originally introduced as a regularization technique to prevent overfitting \cite{srivastava2014dropout}, randomly deactivates a subset of neurons during training, effectively sampling from an ensemble of sub-networks.

In the context of PINNs, dropout is applied during both training and inference phases. During training, it regularizes the network by preventing co-adaptation of neurons, while during inference, it enables uncertainty quantification through Monte Carlo sampling. By performing multiple forward passes with different dropout masks, one can obtain a distribution of predictions that reflects the uncertainty in the model's output. The predictive mean and variance are then computed as:
\begin{subequations}
\label{eq:dropout_stats}
\begin{align}
    \mathbb{E}[u_\theta(x)] &= \frac{1}{M} \sum_{m=1}^{M} u_{\theta^{(m)}}(x), \label{eq:dropout_mean} \\
    \operatorname{Var}[u_\theta(x)] &= \frac{1}{M} \sum_{m=1}^{M} \left(u_{\theta^{(m)}}(x) - \mathbb{E}[u_\theta(x)]\right)^2, \label{eq:dropout_var}
 \end{align}
 \end{subequations}
where $u_{\theta^{(m)}}$ denotes the output of the model's $m$-th forward pass, $\theta^{(m)}$ represents the network parameters sampled with dropout for the $m$-th forward pass, and $M$ is the total number of Monte Carlo samples.

While dropout provides a computationally efficient method for uncertainty quantification, it has inherent limitations. As reported in \cite{yang2021b}, the choice of dropout rate significantly influences the quality of the uncertainty estimates, with higher rates potentially leading to underconfident predictions and lower rates risking overfitting \cite{gal2016dropout}. Moreover, dropout-based methods do not adequately represent the full uncertainty spectrum, especially in regions with sparse or no training data \cite{yang2021b, yao2019quality}.

Nevertheless, when compared to Bayesian approaches like B-PINNs, dropout-based methods offer a simpler implementation with lower computational overhead. The method requires minimal modifications to the existing PINN architecture and can be easily integrated into existing frameworks. Furthermore, the computational cost of uncertainty estimation scales linearly with the number of Monte Carlo samples $M$, making it particularly attractive for large-scale applications where full Bayesian inference might be computationally prohibitive. In Section~3, we use dropout to compare the quality of uncertainty and computational time in various examples.

\subsection{Epistemic physics-informed neural networks}

Epistemic neural networks are a generalization of ensemble-based uncertainty estimation methods proposed by \cite{osband2023epistemic}. This framework defines a parameterized function $u$, approximated by a neural network with learnable parameters $\theta$ as $u_\theta$, and a reference distribution $P_z$, typically chosen as a standard normal distribution $\mathcal{N}(0, I)$. An \emph{epistemic index}, denoted by $z$, is sampled from $P_z$ and used to express the uncertainty of $u$. Thus, for an input $x$ and epistemic index $z$, we obtain a realization of the output, $u_\theta(x, z)$.

\begin{figure}[H]
    \centering
    \includegraphics[width=0.5\linewidth]{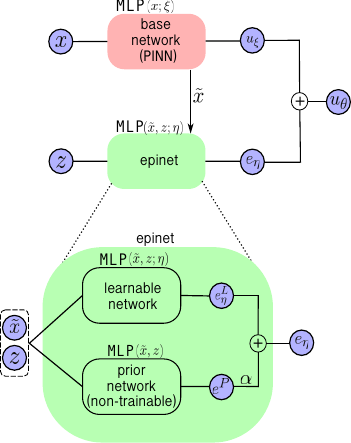}
    \caption{Schematic diagram of the E-PINN architecture. The base PINN (red) produces a deterministic prediction $u_\xi$. The epinet (green) receives as input the augmented feature vector $\tilde{x} = \texttt{sg}\big([\,x,\; h_{\text{penultimate}}(x;\xi)\,]\big)$, where \texttt{sg} blocks gradients to the base network, together with a random epistemic index $z \sim P_z$. Internally, the epinet comprises two neural networks: a trainable network producing $e_\eta^L$ and a non-trainable prior network with fixed parameters producing $e^P$. The final epinet output is $e_\eta = e_\eta^L + \alpha\, e^P$, a weighted sum controlled by the hyperparameter $\alpha$. The sum of the epinet output and the base PINN output yields the stochastic E-PINN prediction $u_\theta$.}
    \label{fig:Epinet_diagram}
\end{figure}

In the present work, we introduce epistemic PINNs (E-PINNs): a modification of the original PINN architecture that supplements the base PINN with an \emph{epinet}, a companion neural network designed to quantify uncertainty in the base network. A diagram of the framework is shown in Fig.~\ref{fig:Epinet_diagram}, and a pseudocode of the algorithm is shown in Algorithm \ref{alg:E-PINN}.

Formally, given an input feature $x$ and an epistemic index $z \sim P_z$, the E-PINN prediction is given by:

\begin{equation}
\label{eq:epinn_output}
\underbrace{u_\theta(x, z)}_{\text{E-PINN output}} = \underbrace{u_{\xi}(x)}_{\text{base PINN output}} + \underbrace{e_{\eta}(\tilde{x}, z)}_{\text{epinet output}},
\end{equation}

where $u_{\xi}(x)$ is the output of the base PINN with trainable parameters $\xi$, and $e_{\eta}(\tilde{x}, z)$ is the output of the epinet. We decompose the epinet as a trainable component with parameters $\eta^L$ and a frozen prior component with parameters $\eta^P$ (not trained). The set of trainable parameters is $\theta = (\xi, \eta^L)$. Throughout this work we adopt a decoupled training procedure: a base PINN is pre-trained and then held fixed while the epinet is trained. We define the augmented feature vector
\[
\tilde x \,\equiv\, \texttt{sg}\big([\,x,\; h_{\text{penultimate}}(x;\xi)\,]\big)\;\in\;\mathbb{R}^{d_x+H_{\text{base}}},
\]
where $h_{\text{penultimate}}(x;\xi)$ are the last hidden-layer activations of the base PINN and $\texttt{sg}$ denotes a stop-gradient operator that blocks backpropagation from the epinet to the base network.

The epinet consists of two components: a learnable network, with output $e_{\eta^L}^L(\tilde{x}, z)$, and a prior network with frozen weights, with output $e^{P}(\tilde{x}, z)$. Specifically, given inputs $\tilde{x}$ and epistemic index $z$, the output of the epinet is:

\begin{equation}
\label{eq:Epinet_output_Epinet}
e_{\eta}(\tilde{x}, z) = e_{\eta^L}^L(\tilde{x}, z) + \alpha\, e^{P}(\tilde{x}, z),
\end{equation}

where $\alpha$ is a hyperparameter used to scale the effect of the prior epinet. In the experiments we treat $\alpha$ as a tunable scalar and, unless stated otherwise, set $\alpha=0.05$. We design $e^{P}$ as a multilayer perceptron (MLP) with initial weights and biases computed using Xavier initialization; its parameters $\eta^P$ remain frozen during training and are not learned. Unless stated otherwise, $z\sim\mathcal{N}(0,I_{d_z})$ with default $d_z=8$.

Intuitively, the two-component epinet design can be understood through the lens of corruption and correction. The frozen prior network $e^P$, with its randomly initialized weights, introduces a form of stochastic perturbation to the deterministic base PINN prediction, essentially acting as a source of epistemic variability that prevents the model from collapsing to overconfident predictions. The trainable network $e_{\eta^L}^L$ then learns to correct and calibrate this baseline variability by adapting to the observed data and physics constraints. The choice of initialization distribution for $e^P$ (Xavier, He, orthogonal, etc.) implicitly determines the prior structure of the epistemic uncertainty, with different distributions yielding different baseline uncertainty patterns.

The E-PINN framework quantifies uncertainty through Monte Carlo sampling over the epistemic index $z$. For uncertainty estimation, we sample multiple realizations $z_1, z_2, \ldots, z_M$ from $P_z$ and compute the corresponding predictions $u_\theta(x, z_i)$. The mean prediction and the epistemic uncertainty are then computed as:
\begin{subequations}
\label{eq:epinn_stats}
\begin{align}
\mu_u(x) &= \mathbb{E}_{Z}\!\left[u_\theta(x, Z)\right] \;\approx\; \frac{1}{M} \sum_{i=1}^{M} u_\theta(x, z_i), \label{eq:overall_mean} \\
\sigma_u^2(x) &= \operatorname{Var}_{Z}\!\left[u_\theta(x, Z)\right] \;\approx\; \frac{1}{M} \sum_{i=1}^{M} \left(u_\theta(x, z_i) - \mu_u(x)\right)^2, \label{eq:total_variance}
\end{align}
\end{subequations}
where $Z\sim P_z$ is the epistemic index and $\{z_i\}_{i=1}^{M}$ are Monte Carlo samples. This approach provides a computationally efficient method for uncertainty quantification that captures the variability in model predictions due to the stochastic nature of the epistemic index.

\begin{algorithm}
\caption{Epistemic Physics-Informed Neural Networks (E-PINNs)}
\label{alg:E-PINN}
\begin{algorithmic}[1]
\Require 
\State PDE operators $\mathcal{N}_x$, $\mathcal{B}_x$, and parameters $\boldsymbol{\lambda}$
\State Training data $\{(x_i^u, \overline{u}_i)\}_{i=1}^{N_u}$, $\{(x_i^f, \overline{f}_i)\}_{i=1}^{N_f}$, $\{(x_i^b, \overline{b}_i)\}_{i=1}^{N_b}$
\State Reference distribution $P_z$ for epistemic index sampling

\Procedure{TrainEPINN}{}
    \State Use a pre-trained base PINN with fixed parameters $\xi$
    \State Initialize epinet learnable parameters $\eta^L$ and fix the prior network weights ($\eta^P$, not trained)
    \Repeat
        \State Sample collocation points, boundary points, and (if available) data points
        \State Sample epistemic index $z \sim P_z$ \Comment{One $z$ per minibatch}
        \State $u_{\xi}(x) \gets \text{BasePINN}(x; \xi)$, \quad $\tilde{x} \gets \texttt{sg}([\,x,\; h_{\text{penultimate}}(x;\xi)\,])$ \Comment{Stop gradient}
        \State $e_{\eta^L}^L(\tilde{x}, z) \gets \text{LearnableNetwork}(\tilde{x}, z; \eta^L)$, \quad $e^{P}(\tilde{x}, z) \gets \text{PriorNetwork}(\tilde{x}, z)$
        \State $u_{\theta}(x, z) \gets u_{\xi}(x) + e_{\eta^L}^L(\tilde{x}, z) + \alpha\, e^{P}(\tilde{x}, z)$
    \State $\mathcal{L}_{\text{pde}} \gets \frac{1}{N_f}\sum_{i=1}^{N_f} |\mathcal{N}_x(u_{\theta}(x_i^f, z); \boldsymbol{\lambda}) - \overline{f}_i|^2$, \quad $\mathcal{L}_{\text{boundary}} \gets \frac{1}{N_b}\sum_{i=1}^{N_b} |\mathcal{B}_x(u_{\theta}(x_i^b, z); \boldsymbol{\lambda}) - \overline{b}_i|^2$
        \State $\mathcal{L}_{\text{data}} \gets \frac{1}{N_u}\sum_{i=1}^{N_u} |u_{\theta}(x_i^u, z) - \overline{u}_i|^2$ (if data are available)
        \State Minimize $\mathcal{L}_{\text{total}} = \mathcal{L}_{\text{data}} + \mathcal{L}_{\text{pde}} + \mathcal{L}_{\text{boundary}}$ w.r.t. $\eta^L$
    \Until{convergence}
\EndProcedure

\Procedure{EstimateUncertainty}{$x$, $M$}
    \State For $i=1,\dots,M$: sample $z_i \sim P_z$ and set $u_i \gets u_{\xi}(x) + e_{\eta}(\tilde{x}, z_i)$
    \State $\mu_u(x) \gets \frac{1}{M}\sum_{i=1}^M u_i$ \Comment{Mean prediction}
    \State $\sigma_u^{2}(x) \gets \frac{1}{M}\sum_{i=1}^M (u_i - \mu_u(x))^2$ \Comment{Variance}
    \State \Return $\mu_u(x)$, $\sigma_u^{2}(x)$
\EndProcedure
\end{algorithmic}
\end{algorithm}

\subsection{Loss function and training}
For the optimization-based approaches in this paper (Dropout-PINN and E-PINN), the underlying training process involves minimizing a composite loss function that incorporates both data observations and physical constraints:

\begin{equation}
\mathcal{L}_{\text{total}} = w_{\text{data}} \,\mathcal{L}_{\text{data}} + w_{\text{pde}} \,\mathcal{L}_{\text{pde}} + w_{\text{bc}} \,\mathcal{L}_{\text{boundary}},
\end{equation}

where $w_{\text{data}}$, $w_{\text{pde}}$, and $w_{\text{bc}}$ are weighting coefficients that balance the different components of the loss, and:

\begin{subequations}
\label{eq:loss_components}
\begin{align}
\mathcal{L}_{\text{data}} &= \frac{1}{N_u}\sum_{i=1}^{N_u} |u_\theta(x_i^u) - \overline{u}_i|^2, \label{eq:loss_data} \\
\mathcal{L}_{\text{pde}} &= \frac{1}{N_f}\sum_{i=1}^{N_f} |\mathcal{N}_x(u_\theta(x_i^f); \boldsymbol{\lambda}) - \overline{f}_i|^2, \label{eq:loss_pde} \\
\mathcal{L}_{\text{boundary}} &= \frac{1}{N_b}\sum_{i=1}^{N_b} |\mathcal{B}_x(u_\theta(x_i^b); \boldsymbol{\lambda}) - \overline{b}_i|^2. \label{eq:loss_bc}
\end{align}
\end{subequations}

The weighting parameters can be adjusted manually or in a self-adaptive manner \cite{mcclenny2023self, anagnostopoulos2024residual, chen2024self}. Unless otherwise specified, we set $w_{\text{data}} = w_{\text{pde}} = 1$. For problems with boundary conditions, $w_{\text{bc}}$ is typically set to 10 to ensure boundary conditions are satisfied accurately.

The term $\mathcal{L}_{\text{data}}$ represents the mean squared error between the network predictions and the available data measurements, $\mathcal{L}_{\text{pde}}$ enforces the PDE residual at the collocation points within the domain, and $\mathcal{L}_{\text{boundary}}$ ensures that the solution satisfies the prescribed boundary conditions.

For B-PINN, we use Gaussian likelihoods over the same residual channels: $u$-data, PDE residual $\mathcal{N}_x(u_\theta)-f$, and boundary residual $\mathcal{B}_x(u_\theta)-b$. With fixed noise levels, the negative log-likelihood recovers the weighted squared terms above up to constants. In HMC, we do not minimize this loss; we sample the posterior over weights using these likelihoods and standard Gaussian parameter priors.

For E-PINNs, the predictive mapping is $u_\theta(x,z)$, where $z\sim P_z$ is the epistemic index. We therefore minimize the expected composite loss over $z$,
\begin{equation}
\mathcal{L}_{\text{total}}(\theta)\;=\;\mathbb{E}_{z\sim P_z}\Big[w_{\text{data}}\,\mathcal{L}_{\text{data}}(u_\theta(\cdot,z))\; +\; w_{\text{pde}}\,\mathcal{L}_{\text{pde}}(u_\theta(\cdot,z))\; +\; w_{\text{bc}}\,\mathcal{L}_{\text{boundary}}(u_\theta(\cdot,z))\Big],
\end{equation}
approximated in practice by sampling one $z$ per minibatch. In the present work our quantity of interest is the latent, noise-free field $u(x)$. We report epistemic uncertainty as the variance estimated by Monte Carlo sampling over $z$,
\begin{equation}
\sigma_u^2(x)\;=\;\operatorname{Var}_{Z}[u_\theta(x,Z)].
\end{equation}

The network parameters $\theta$ are optimized by solving the minimization problem:
\begin{equation}
\theta^* = \arg\min_{\theta} \mathcal{L}_{\text{total}}(\theta).
\end{equation}

This minimization problem is solved approximately using the Adam optimization algorithm. All networks in this work are trained with the Adam optimizer, with learning rate $10^{-3}$, exponential decay rates for the first and second moment estimates $\beta_1=0.9$ and $\beta_2 = 0.999$, respectively. Unless otherwise noted, the base PINN is trained for 100,000 epochs and the epinet for 10,000 epochs in a decoupled fashion; early stopping is not used.

\paragraph{Notation}
- Domain and operators: $\Omega\subset\mathbb R^d$, $\partial\Omega$; $\mathcal N_x(\cdot;\boldsymbol{\lambda})$, $\mathcal B_x(\cdot;\boldsymbol{\lambda})$; source $f(x)$.
- Base PINN: parameters $\xi$; output $u_\xi(x)$; last hidden activations $h_{\text{penultimate}}(x;\xi)$.
- Features: $\tilde x = \texttt{sg}([\,x,\;h_{\text{penultimate}}(x;\xi)\,])$.
- epinet: $e^L_{\eta^L}(\tilde x,z)$ (trainable), $e^P(\tilde x,z)$ (frozen); scale $\alpha$.
- E-PINN output: $u_\theta(x,z)=u_\xi(x)+e^L_{\eta^L}(\tilde x,z)+\alpha e^P(\tilde x,z)$, with $\theta=(\xi,\eta^L)$.
- Epistemic index: $z\sim\mathcal N(0,I_{d_z})$, default $d_z=8$.
- Physics operators on network: $\mathcal N_x(u_\theta(x);\boldsymbol{\lambda})$, $\mathcal B_x(u_\theta(x);\boldsymbol{\lambda})$; optional residual shorthands $r_\theta(x)=\mathcal N_x(u_\theta(x);\boldsymbol{\lambda})-f(x)$ and $g_\theta(x)=\mathcal B_x(u_\theta(x);\boldsymbol{\lambda})-b(x)$.
- Loss weights: $w_{\text{data}}, w_{\text{pde}}, w_{\text{bc}}$ (defaults $w_{\text{data}}=w_{\text{pde}}=1$, $w_{\text{bc}}=10$).
- Uncertainty: $\mu_u(x)=\mathbb E_Z[u_\theta(x,Z)]$, $\sigma_u^2(x)=\mathrm{Var}_Z[u_\theta(x,Z)]$.

Implementation details (common across experiments): residuals and required derivatives are computed with reverse-mode automatic differentiation (PyTorch autograd). Dropout-PINNs apply dropout to all hidden layers after the activation; input and output layers have no dropout; the dropout rate $p$ is shared across hidden layers. Inputs and outputs are left unnormalized unless explicitly stated. We enforce boundary conditions softly via $\mathcal{L}_{\text{boundary}}$ with weight $w_{\text{bc}}=10$ unless stated otherwise. For reproducibility, experiments use fixed random seeds recorded in the code.

\section{Results}
We compare the three uncertainty quantification methods introduced in Section~2. The goal is to illustrate the applicability of E-PINNs for estimating epistemic uncertainty, to evaluate computational costs, and to characterize uncertainty bands.

\subsection{Evaluation Metrics}

We evaluate two standard metrics: \emph{sharpness} and \emph{coverage}. Sharpness refers to the width of predictive intervals, while coverage indicates how often these intervals contain the true values \cite{gneiting2007probabilistic, pearce2018high}. As noted by \cite{pearce2018high}, prediction intervals should ideally be narrow while still capturing the expected proportion of observations.

We quantify \textit{sharpness} as the average width of the $\pm 2 \sigma$ predictive intervals \cite{gneiting2007probabilistic}:
\begin{equation}
    \mu^w_{2\sigma} = \frac{4}{N}\sum_{i=1}^{N}\sqrt{\sigma_u^2(x_i)},
\end{equation}
where $N$ denotes the number of points at which the epistemic variance $\sigma_u^2(x_i)$ is evaluated. Smaller values of $\mu^w_{2\sigma}$ indicate sharper uncertainty estimates.

In addition to sharpness, we evaluate the quality of uncertainty estimates using \textit{empirical coverage}, which measures the percentage of true values that fall within the predicted uncertainty bounds \cite{khosravi2011comprehensive}. Specifically, for a given confidence level $\gamma$ (typically 95\% corresponding to $\pm 2\sigma$ bounds), the empirical coverage is calculated as:

\begin{equation}
    \text{C}_{\gamma} = \frac{1}{N}\sum_{i=1}^{N} \mathbf{1}\Big\{|u(x_i) - \mu_u(x_i)| \leq \Phi^{-1}\!\big(\frac{1+\gamma}{2}\big)\, \sqrt{\sigma_u^2(x_i)}\Big\},
\end{equation}
where $\mathbf{1}\{\cdot\}$ is the indicator function, $u(x_i)$ is the true (exact) value, $\mu_u(x_i)$ and $\sigma_u^2(x_i)$ are the predicted mean and epistemic variance at point $x_i$, and $\Phi^{-1}((1+\gamma)/2)$ is the appropriate quantile of the standard normal distribution (approximately 1.96 for $\gamma= 0.95$). An ideal uncertainty quantification method should achieve an empirical coverage close to the specified confidence level: a value too low indicates overconfidence, while too high suggests overly conservative models \cite{khosravi2011comprehensive}.

We also report the root-mean-square error (RMSE) computed with the predictive mean $\mu_u(x)$ against the exact solution on dense validation grids.

\subsection{Benchmarks and Computational Setup}

To ensure a fair comparison of these metrics across methods, we implement the three uncertainty quantification approaches with consistent settings. We focus exclusively on epistemic uncertainty: all reported variances arise from either sampling the HMC posterior (B-PINN), Monte Carlo dropout (Dropout-PINN), or sampling the epistemic index $z$ (E-PINN). In physics-only settings, losses use deterministic PDE residuals with known $f(x)$ and exact boundary values; no interior $u$ measurements are used. In data-augmented settings, we add noisy interior $u$ observations with zero-mean Gaussian noise of standard deviation $\sigma = \rho\,\lVert u\rVert_{\infty}$, where $\lVert u\rVert_{\infty}$ denotes the sup norm on the evaluation domain and $\rho\in\{0.10, 0.30\}$. Equivalently, we can view this as a homoscedastic model with $\sigma_i^u \equiv \sigma$ for all $i$, whereas the indexed form $\sigma_i^u$ would allow known heteroscedastic noise if desired.

All experiments were executed on a system with an AMD EPYC 7502 CPU. Each job ran independently using 4 CPU cores, and no GPU acceleration was used. The "Time (s)" columns in the tables report wall-clock elapsed time as follows: E-PINN totals base and epinet training time, Dropout-PINN sums training and Monte Carlo inference time, and B-PINN reports Hamiltonian Monte Carlo sampling plus inference time.

\paragraph{Note on uncertainty plots} Within each multi-panel figure in Section~3, the vertical axis (scale) for uncertainty plots is fixed across methods. This can make E-PINN bands appear visually narrower relative to the other methods. For quantitative comparisons, refer to the tables reporting sharpness and coverage. The ablation study (Section~\ref{sec:ablation}) shows that the E-PINN standard deviation varies systematically with the noise level and other hyperparameters, indicating that E-PINN does not underestimate uncertainty under the tested settings.

For the B-PINN results (evaluated only in data-augmented regimes), we use Hamiltonian Monte Carlo with an identity mass matrix. Unless otherwise noted, the HMC parameters are: step size $5\times 10^{-5}$, $L=50$ leapfrog steps, $1{,}000$ burn-in steps, and $M=11{,}000$ total samples. For the 2D nonlinear Poisson case, to enhance stability, we use a smaller step size $1\times 10^{-5}$ with the same $L$, burn-in, and total samples. We place standard zero-mean Gaussian priors on weights and biases (e.g., $W\sim\mathcal N(0,1)$, $b\sim\mathcal N(0,1)$) and tune the step size to achieve the stated acceptance. In B-PINN runs we enforce boundary and physics residual channels with small noise tolerances (0.01) for HMC stability; in physics-only settings we do not evaluate B-PINN.

Boundary condition targets are generated without added noise. In B-PINN likelihoods we assign a small residual noise level to the physics and boundary channels (standard deviation $0.01$) for HMC stability; optimization-based methods (Dropout-PINN and E-PINN) use squared residual losses without injected noise on these channels. For Dropout-PINNs, the same dropout rate is used during training and inference. Uncertainty bands are computed from $M=10{,}000$ stochastic forward passes. For E-PINNs, we adopt a decoupled setup in which a pre-trained PINN provides features to an epinet that is trained subsequently. Unless otherwise specified, the base PINN uses three hidden layers with 32 neurons each (32, 32, 32), the epinet uses three hidden layers with 32 neurons each (32, 32, 32) and an epistemic index of dimension $d_z=8$, and the prior network contribution is scaled by the hyperparameter $\alpha$.

\subsection{Forward problems}

\subsubsection{1D Poisson equation}

In this first case we consider the 1D Poisson equation, given by 
\begin{equation}
\label{eq:1DPoissons}
    \lambda \, \frac{\partial ^2 u}{\partial x ^2} = f(x), \quad x \in [-1,1],
\end{equation}
with exact solution
\begin{equation}
\label{eq:1DPoissons_exact_sol}
    u(x) = \sin^3(6x).
\end{equation}

The analytical form for $f$ follows by substituting Eq.~\ref{eq:1DPoissons_exact_sol} into Eq.~\ref{eq:1DPoissons}. The base PINN is trained using 100 uniformly spaced collocation points in $x\in[-1,1]$ together with exact Dirichlet boundary conditions at the endpoints. We consider Dropout-PINN (Monte Carlo dropout) and E-PINN (epinet on a pre-trained PINN) in the physics-only setting, and additionally include B-PINN (HMC posterior over weights) for the cases with noise fractions $0.10$ and $0.30$.

Two training regimes are examined. In a physics-only regime, models are fitted solely to the PDE residual together with the exact boundary conditions; no interior observations are used. In this setting, the reported uncertainty is epistemic, reflecting the model's uncertainty over the interior induced by the absence of interior data. In a data-augmented regime, the physics loss is complemented with 32 interior observations of $u(x)$ corrupted with Gaussian noise at relative amplitudes $0.10$ and $0.30$ of $\lVert u \rVert_{\infty}$. Across both regimes, we report sharpness (mean width of the $\pm 2\sigma$ band), empirical 95\% coverage, and RMSE on $u$ evaluated on a dense validation grid. We set the diffusion coefficient to $\lambda=0.01$, $N_\text{colloc}=100$, $N_\text{sensors}=32$, boundary-loss weight $w_{\text{bc}}=10$, and evaluate uncertainty with $M=10{,}000$ Monte Carlo samples.

We instantiate the base PINN as an MLP with three hidden layers of 32 neurons each (32, 32, 32). The epinet also uses three hidden layers with 32 neurons each (32, 32, 32), an epistemic index dimension $d_z=8$, and a fixed prior network with hidden widths $(5,5)$ scaled by the factor $\alpha$. In the experiments, we set $\alpha=0.05$. For Dropout-PINNs we report results for dropout rates $p\in\{0.05,0.10\}$.

\begin{figure}[H]
    \centering
    \begin{subfigure}[b]{0.45\textwidth}
        \centering
        \incfig[width=\textwidth]{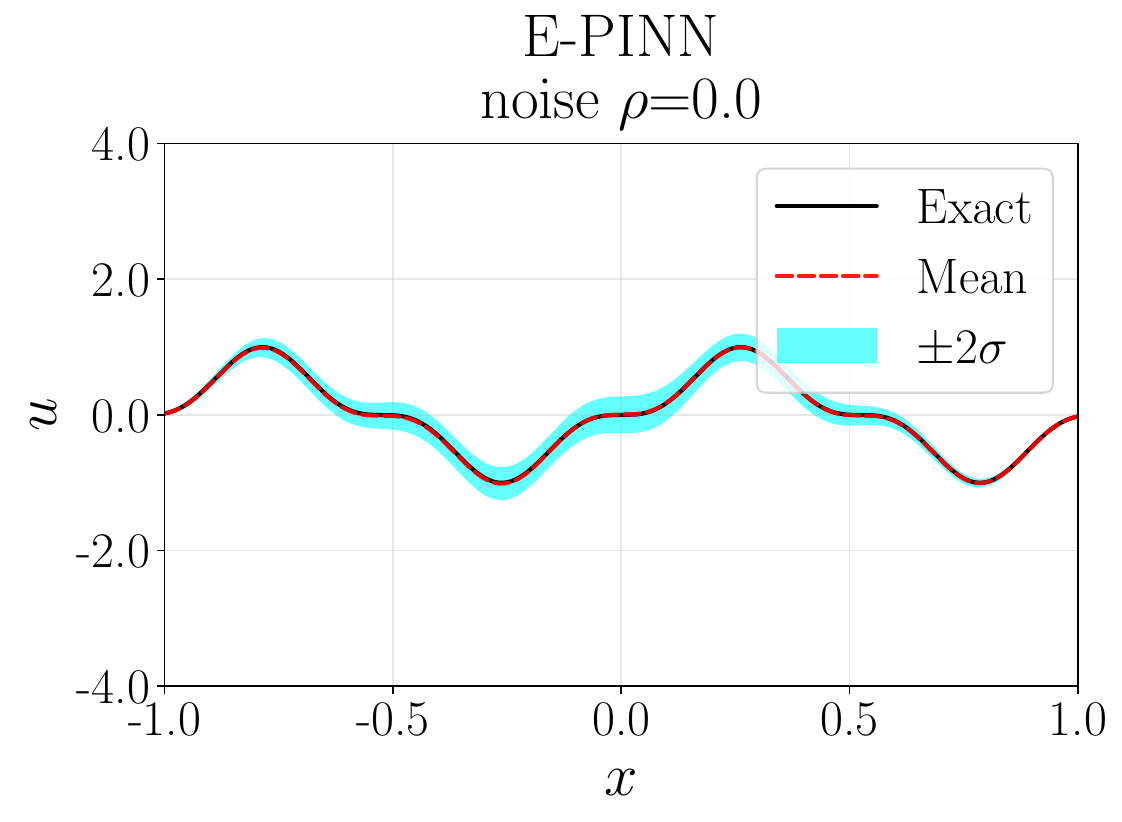}
        \caption{}
    \end{subfigure}
    \par\vspace{0.5em}
    \begin{subfigure}[b]{0.45\textwidth}
        \centering
        \incfig[width=\textwidth]{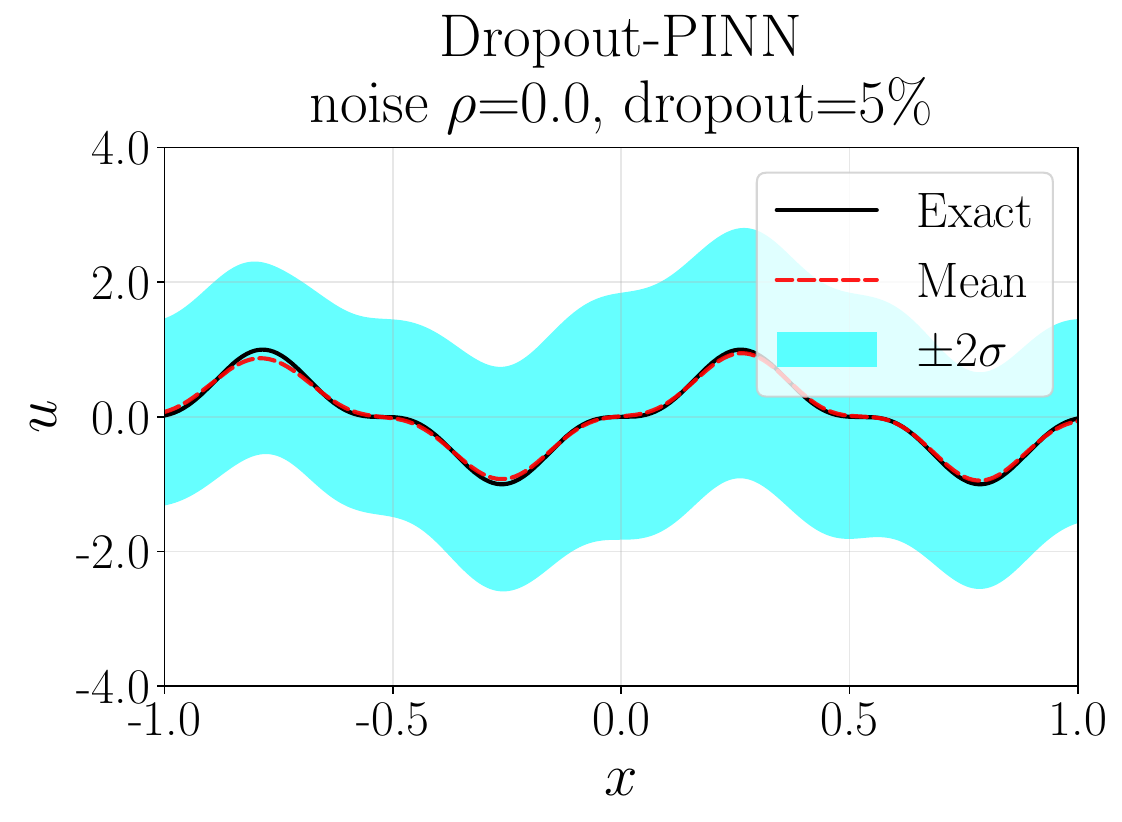}
        \caption{}
    \end{subfigure}
    \hfill
    \begin{subfigure}[b]{0.45\textwidth}
        \centering
        \incfig[width=\textwidth]{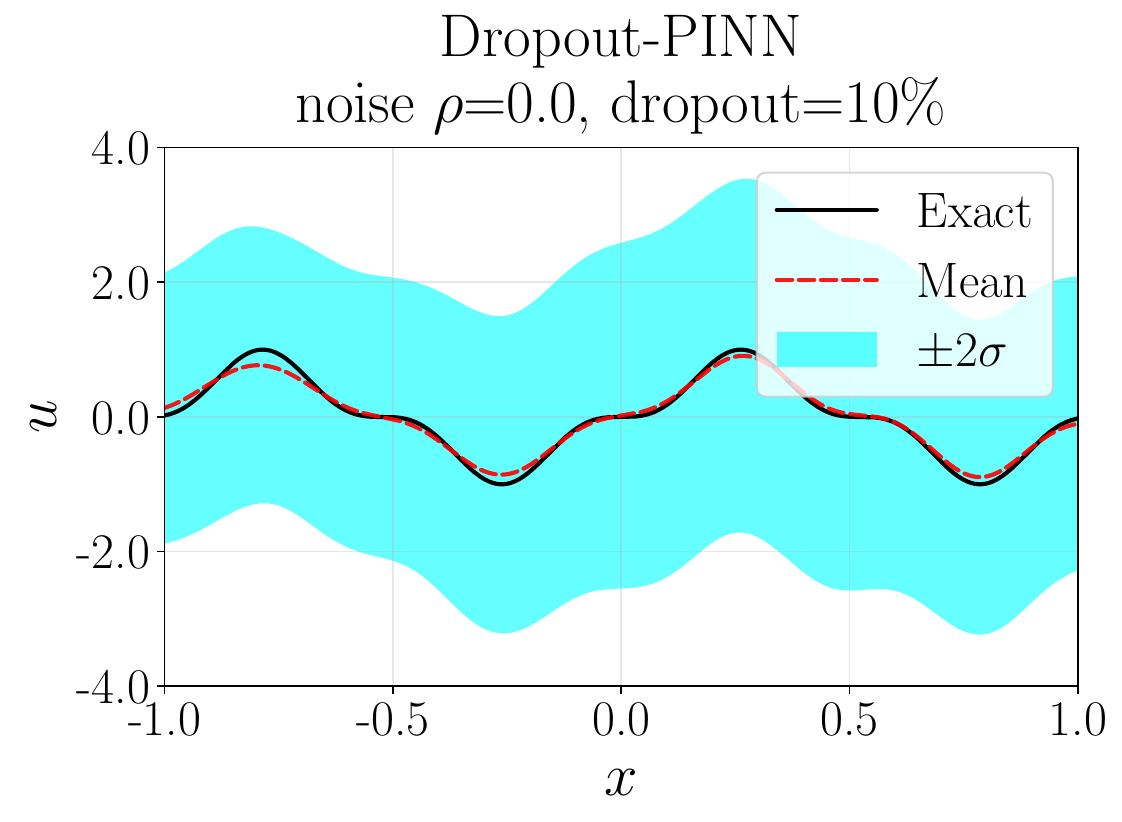}
        \caption{}
    \end{subfigure}
    \caption{Mean and uncertainty predictions for $u$ in the 1D Poisson equation (physics-only) using (a) E-PINN (top row) and (b,c) Dropout-PINN (bottom row; 5\% and 10\%).}
    \label{fig:1DPoisson_0.0}
\end{figure}

In the physics-only setting (Fig.~\ref{fig:1DPoisson_0.0}) the methods recover the overall profile, with the largest uncertainty in the interior where no data are provided. E-PINN yields comparatively narrow bands and the lowest error; Dropout-PINN produces substantially wider intervals that depend on the dropout rate (5\% narrower, 10\% broader).

\begin{figure}[H]
    \centering
    \begin{subfigure}[b]{0.45\textwidth}
        \centering
        \incfig[width=\textwidth]{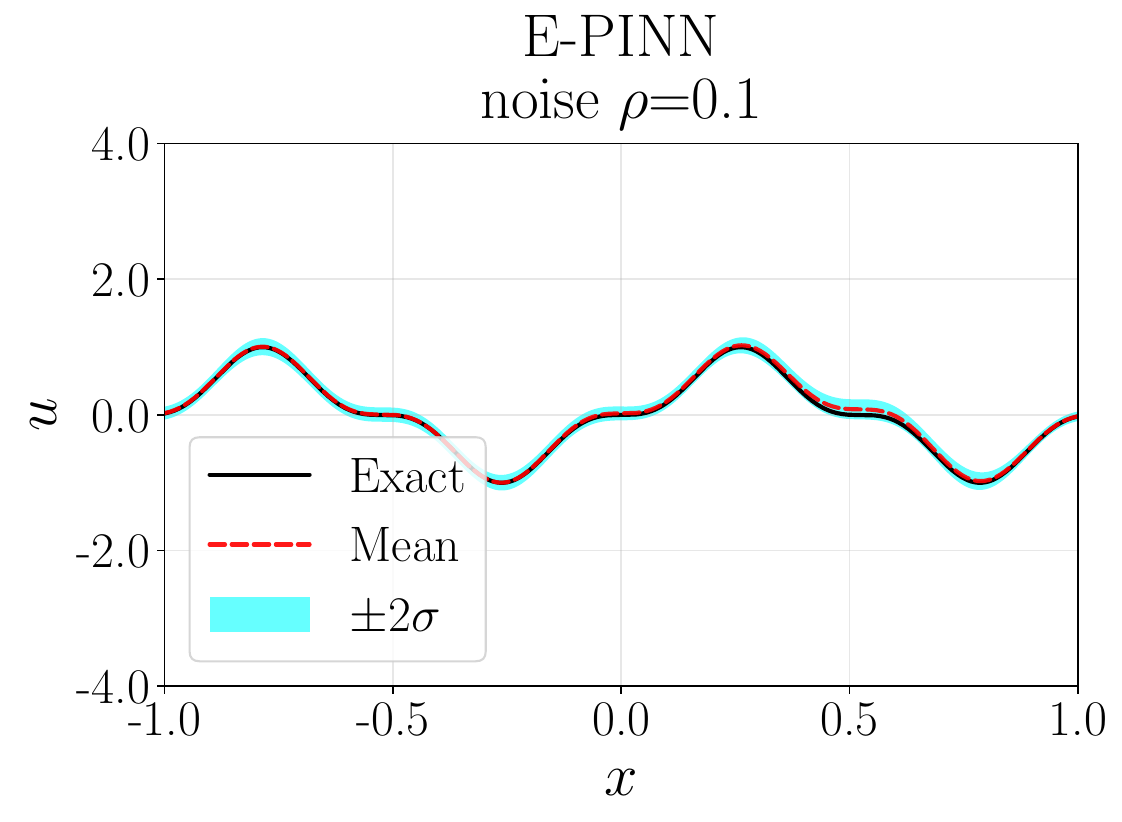}
        \caption{}
    \end{subfigure}
    \begin{subfigure}[b]{0.45\textwidth}
        \centering
        \incfig[width=\textwidth]{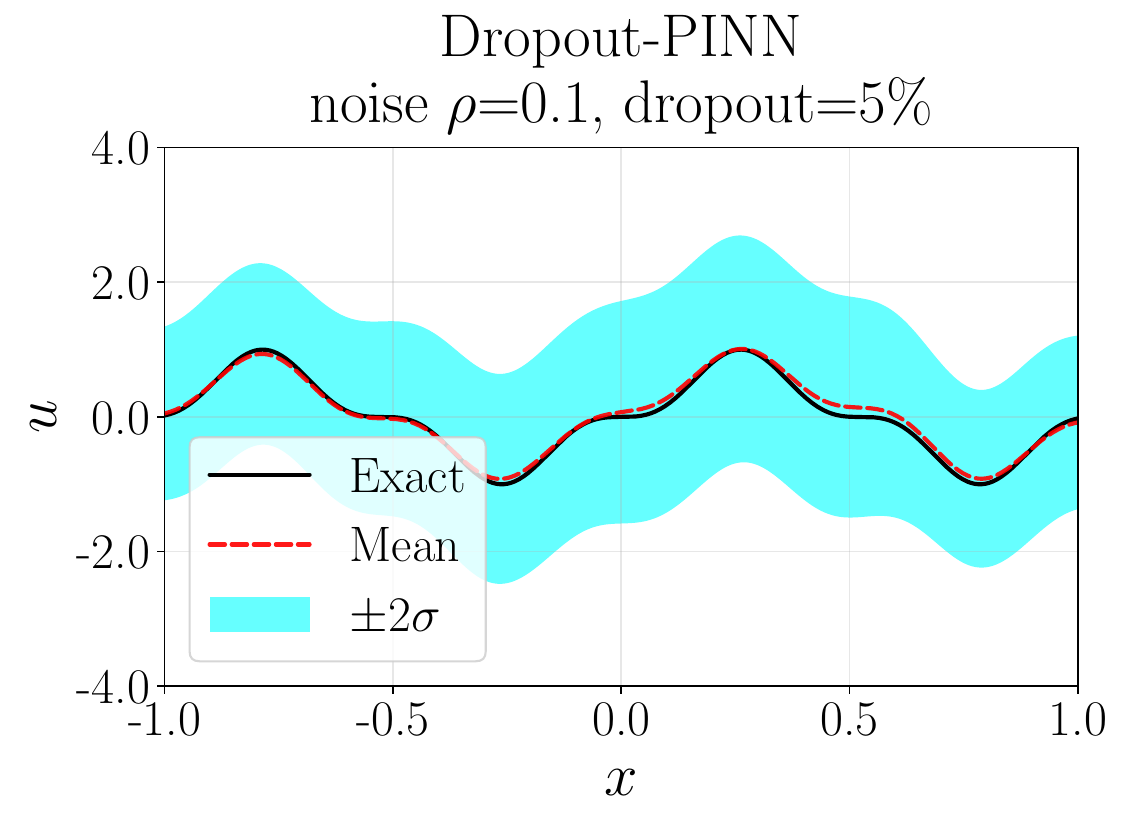}
        \caption{}
    \end{subfigure}
    \hfill
    \begin{subfigure}[b]{0.45\textwidth}
        \centering
        \incfig[width=\textwidth]{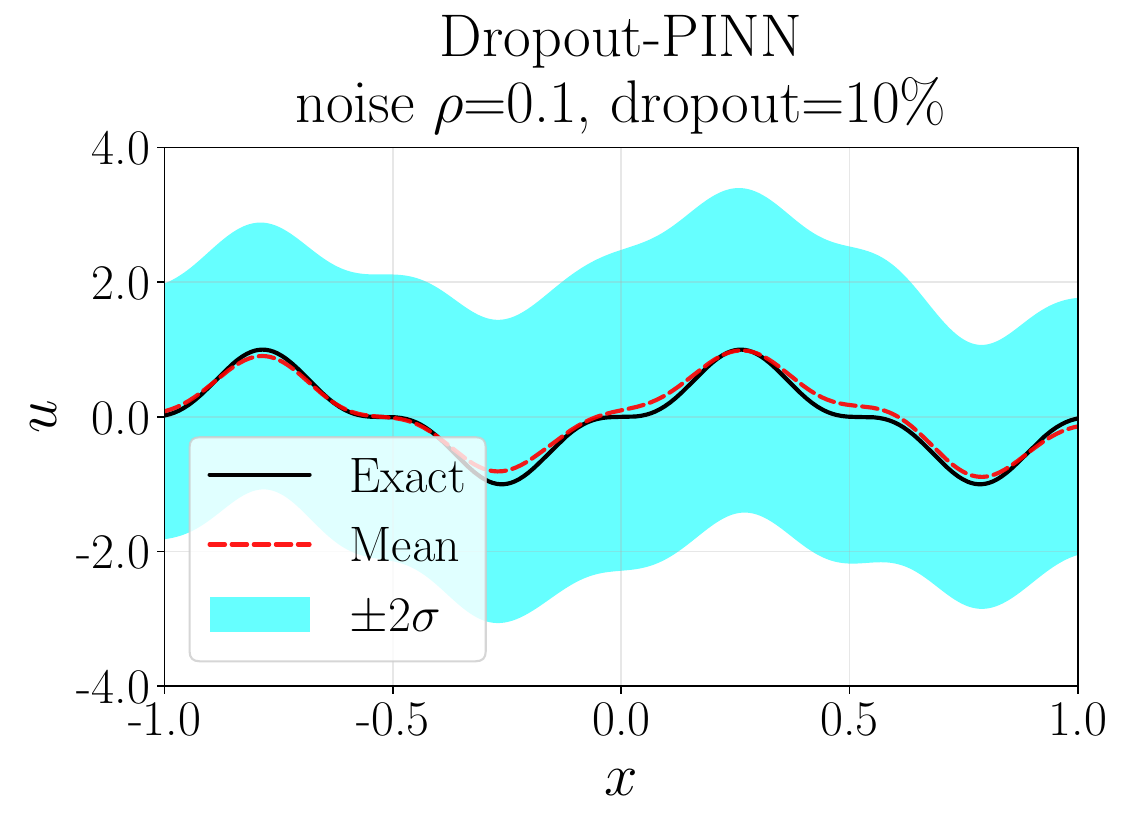}
        \caption{}
    \end{subfigure}
    \begin{subfigure}[b]{0.45\textwidth}
        \centering
        \incfig[width=\textwidth]{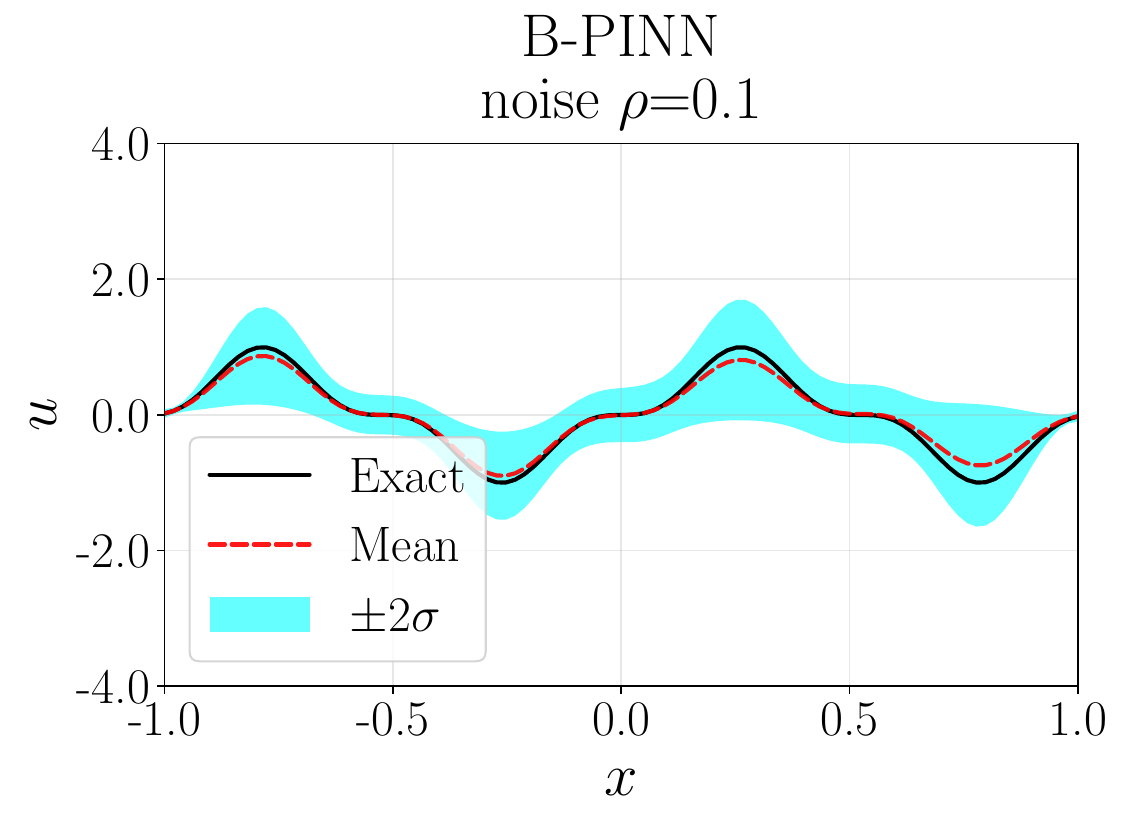}
        \caption{}
    \end{subfigure}
\caption{Mean and uncertainty predictions for $u$ in the 1D Poisson equation (physics + interior data, Gaussian noise with $\sigma = 0.10\,\lVert u\rVert_{\infty}$) using (a) E-PINN (b) Dropout-PINN (5\%) (c) Dropout-PINN (10\%) (d) B-PINN.}
    \label{fig:1DPoisson_0.1}
\end{figure}

With noisy interior observations (Fig.~\ref{fig:1DPoisson_0.1}), uncertainty bands widen and mean accuracy degrades across methods. At $\rho\in\{0.10,0.30\}$, E-PINN remains sharper and attains the smallest errors in these runs. Dropout-PINN is widest in all cases with full coverage. In these B-PINN runs, coverage equals 1.00 while bands are much wider than E-PINN and errors are larger; total runtime is higher due to HMC sampling.

In Table~\ref{tab:poisson_all}, physics-only ($\rho=0$) shows E-PINN with narrow bands (sharpness $\mu^w_{2\sigma}\approx0.33$) and the smallest RMSE ($\approx7.4\times10^{-3}$) with coverage equal to 1.00. Dropout-PINN yields much wider intervals (about 3.31 to 4.72) and larger RMSE. B-PINN is not evaluated in this regime. With interior $u$ data at $\rho=0.10$ and $\rho=0.30$, E-PINN remains sharper (about 0.23 and 0.34) and more accurate (RMSE about $3.4\times10^{-2}$ and $9.0\times10^{-2}$) than B-PINN (about 0.97 and 1.84; RMSE about 0.0996 and 0.1638) while requiring less time (about $5.3\times10^2$ to $7.2\times10^2$ s vs about $1.7\times10^3$ s). In all settings Dropout-PINN produces the widest intervals and larger RMSE than E-PINN; in the data-augmented cases, Dropout-PINN RMSE is comparable to or slightly below B-PINN. The last column reports the E-PINN epinet-only training time (about 114 to 168 s), which applies when a base PINN is already trained.

At $\rho\in\{0.10,0.30\}$, E-PINN remains sharper than the alternatives and attains the smallest errors in these runs. B-PINN coverage equals 1.00 in these cases with wider bands and larger RMSE than E-PINN; Dropout-PINN produces the widest intervals with full coverage.

\begin{table}[H]
\centering
\caption{1D Poisson: metrics vs.\ noise scale $\rho$ (standard deviation $\sigma=\rho\,\lVert u\rVert_{\infty}$). RMSE is computed with the predictive mean $\mu_u$. Time (s) is total wall-clock time; Time (epinet) (s) reports the E-PINN epinet-only training time (only populated for E-PINN rows). For $\rho=0$ (physics-only), B-PINN is not evaluated. Lower is better for sharpness, RMSE, and time; higher is better for coverage.}
\label{tab:poisson_all}
\sisetup{detect-weight=true,detect-family=true,mode=text}
\begin{tabular*}{\textwidth}{@{\extracolsep{\fill}}
S[table-format=1.2]
l
S[table-format=1.2]
S[table-format=1.2]
S[table-format=1.4]
S[table-format=4.2]
 c
}
\toprule
\multicolumn{1}{c}{$\rho$} & \multicolumn{1}{c}{Method} &
\multicolumn{1}{c}{Sharpness $\mu^w_{2\sigma}$} &
\multicolumn{1}{c}{Coverage (95\%)} &
\multicolumn{1}{c}{RMSE} &
\multicolumn{1}{c}{Time (s)} & \multicolumn{1}{c}{Time (epinet) (s)} \\
\midrule
0.00 & E-PINN        & \bfseries 0.33 & \bfseries 1.00 & \bfseries 0.0074 & 510.75 & 113.83 \\
0.00 & Dropout 5\%   & 3.31             & \bfseries 1.00 & 0.0443              & 420.34 & -- \\
0.00 & Dropout 10\%  & 4.72             & \bfseries 1.00 & 0.0830              & \bfseries 306.02  & -- \\
\midrule
0.10 & E-PINN        & \bfseries 0.23 & \bfseries 1.00 & \bfseries 0.0337 & 526.85 & 129.19 \\
0.10 & B-PINN        & 0.97             & 1.00              & 0.0996              & 1672.45 & -- \\
0.10 & Dropout 5\%   & 3.00             & \bfseries 1.00 & 0.0700              & 537.27 & -- \\
0.10 & Dropout 10\%  & 4.37             & \bfseries 1.00 & 0.0993              & \bfseries 452.17 & -- \\
\midrule
0.30 & E-PINN        & \bfseries 0.34 & \bfseries 1.00 & \bfseries 0.0901 & 715.71 & 168.11 \\
0.30 & B-PINN        & 1.84             & 1.00              & 0.1638              & 1670.35 & -- \\
0.30 & Dropout 5\%   & 2.55             & \bfseries 1.00 & 0.1576              & \bfseries 366.70 & -- \\
0.30 & Dropout 10\%  & 3.69             & \bfseries 1.00 & 0.1611              & 431.00 & -- \\
\bottomrule
\end{tabular*}
\end{table}

\subsubsection{1D nonlinear Poisson equation}

In this section we consider the 1D nonlinear Poisson equation, given by,
\begin{equation}
    \lambda\, \frac{\partial^2 u}{\partial x^2} + k \tanh(u) = f(x), \quad x \in [-0.7,0.7],
\label{eq:Non-linPoissons}
\end{equation}
with exact solution given by,
\begin{equation}
    u(x) = \sin^3(6x),
\label{eq:Non-linPoissonsExact}
\end{equation}

with constants $\lambda=0.01$ and $k = 0.7$. The analytical form for $f$ can be computed by substituting Eq.~\ref{eq:Non-linPoissonsExact} into Eq.~\ref{eq:Non-linPoissons}. The base PINN is trained using 100 collocation points uniformly distributed in $x \in [-0.7,0.7]$ with exact boundary conditions at the endpoints. We evaluate two scenarios: (i) physics-only (no interior data) and (ii) physics + noisy $u$ data with 32 interior sensors (fractions $0.10$ and $0.30$ of $\lVert u \rVert_{\infty}$). We do not add noise to the PDE residual or boundary values in either case. The epinet prior scaling is set to $\alpha=0.05$.

\begin{figure}[H]
    \centering
    \begin{subfigure}[b]{0.45\textwidth}
        \centering
        \incfig[width=\textwidth]{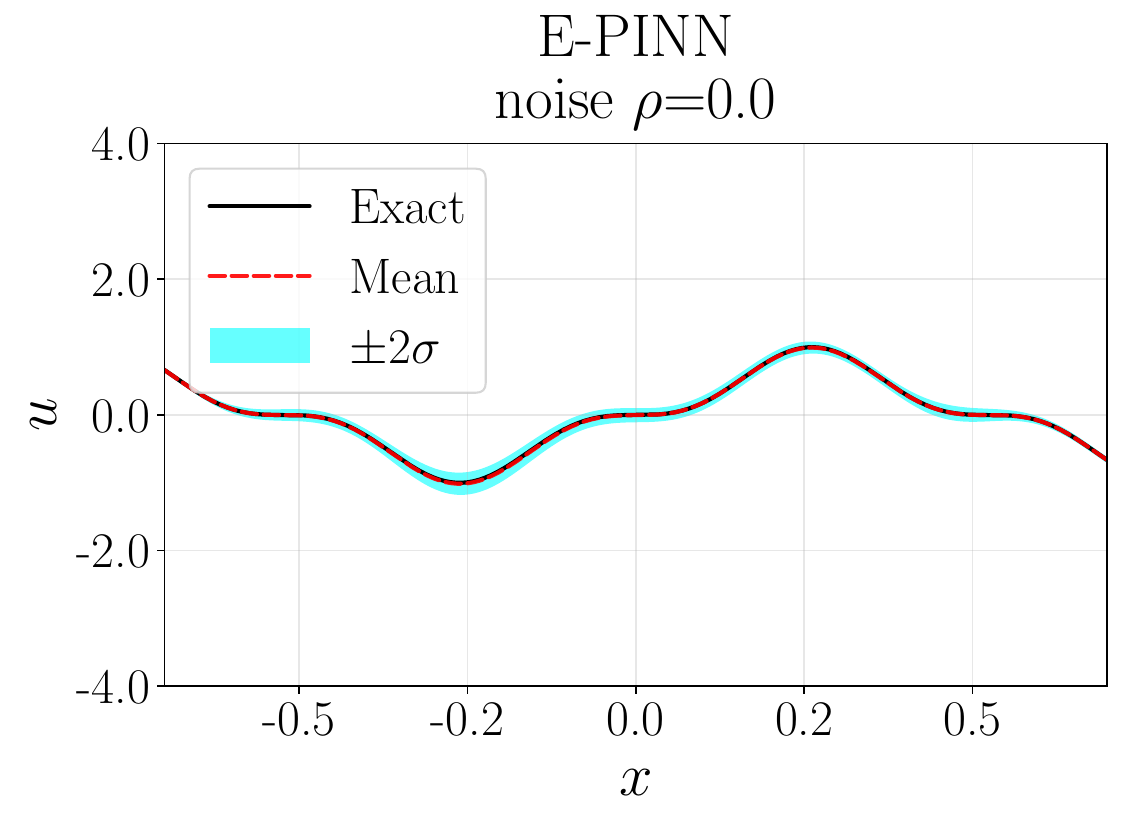}
        \caption{}
    \end{subfigure}
    \hfill
    \begin{subfigure}[b]{0.45\textwidth}
        \centering
        \incfig[width=\textwidth]{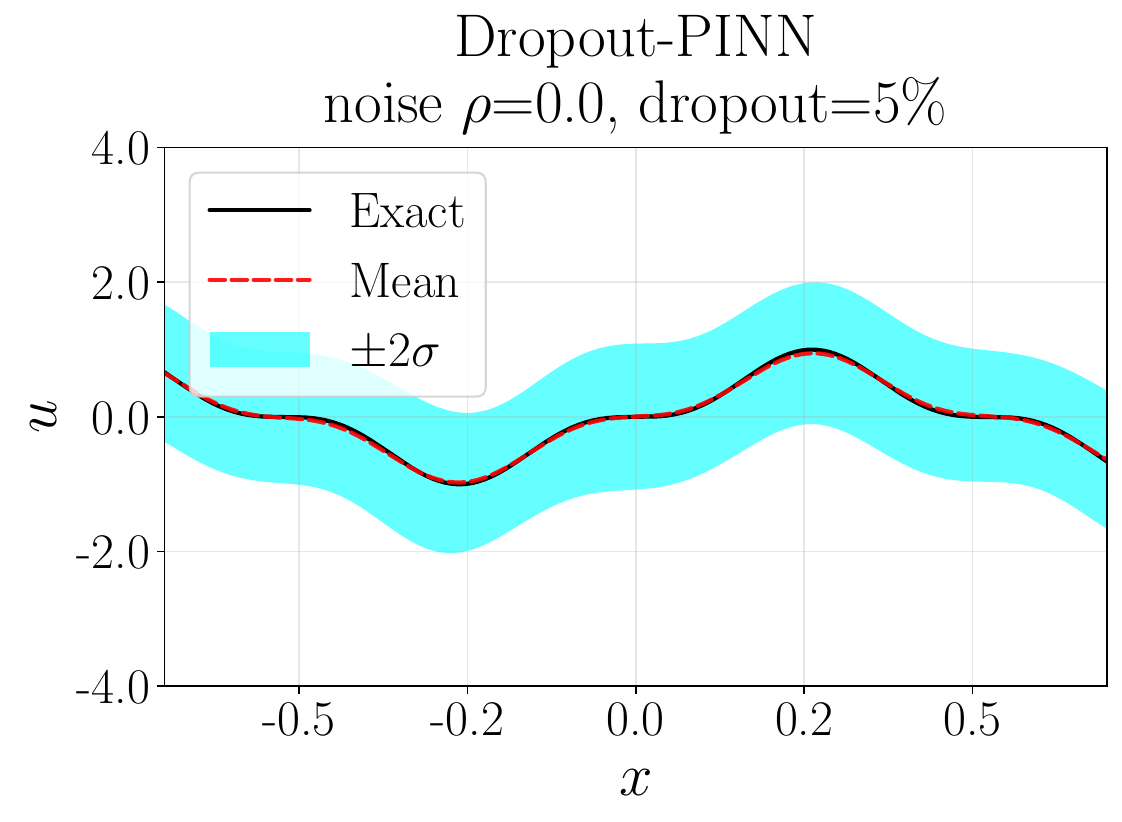}
        \caption{}
    \end{subfigure}
    \begin{subfigure}[b]{0.45\textwidth}
        \centering
        \incfig[width=\textwidth]{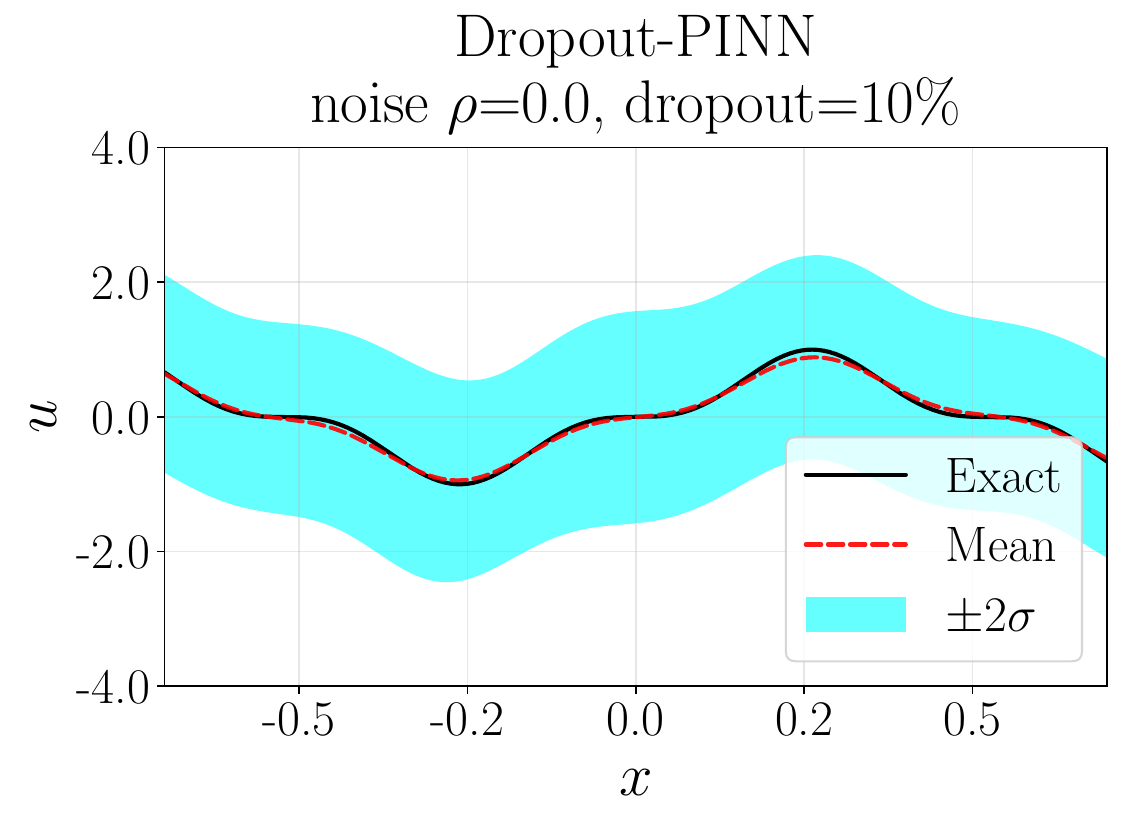}
        \caption{}
    \end{subfigure}
    
\caption{Mean and uncertainty predictions for $u$ in 1D nonlinear Poisson (physics-only, no interior data) using (a) E-PINN (b) Dropout-PINN (5\%) (c) Dropout-PINN (10\%). B-PINN is not evaluated in physics-only.}
    \label{fig:1DNon-linPoissons_0.0}
\end{figure}

\begin{figure}[H]
    \centering
    \begin{subfigure}[b]{0.45\textwidth}
        \centering
        \incfig[width=\textwidth]{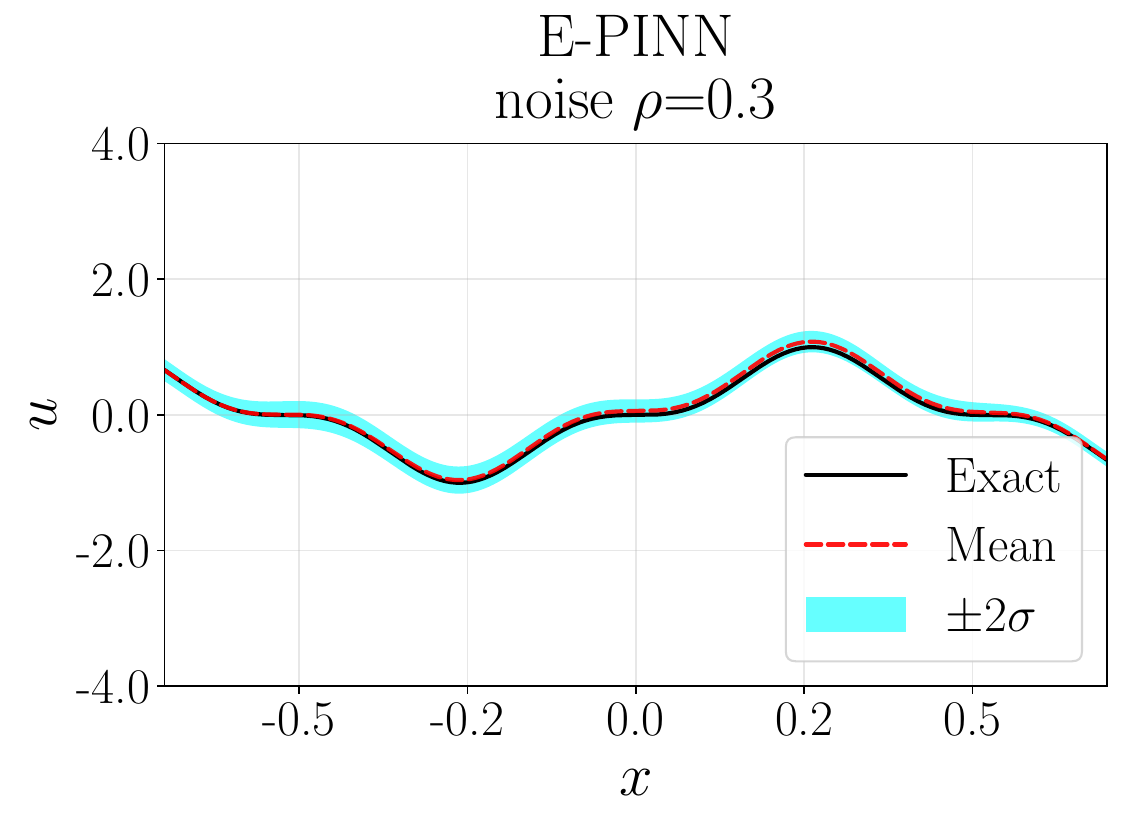}
        \caption{}
    \end{subfigure}
    \hfill
    \begin{subfigure}[b]{0.45\textwidth}
        \centering
        \incfig[width=\textwidth]{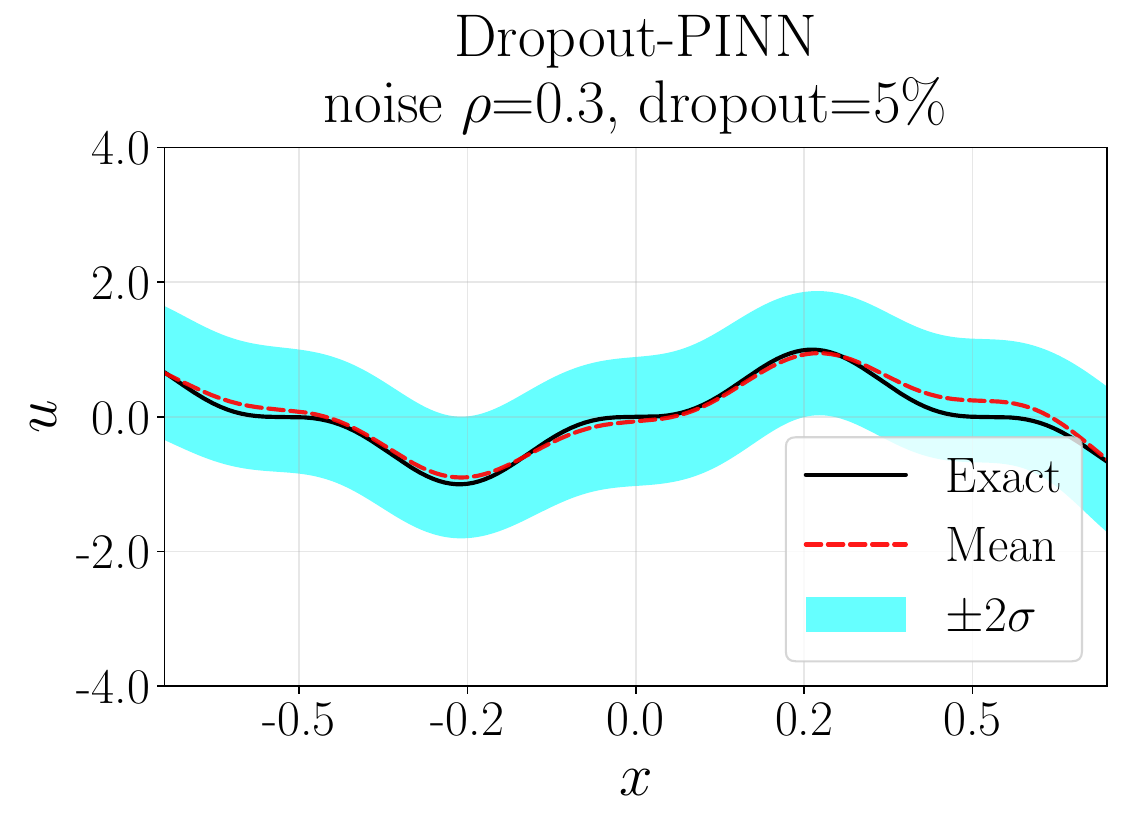}
        \caption{}
    \end{subfigure}
    \begin{subfigure}[b]{0.45\textwidth}
        \centering
        \incfig[width=\textwidth]{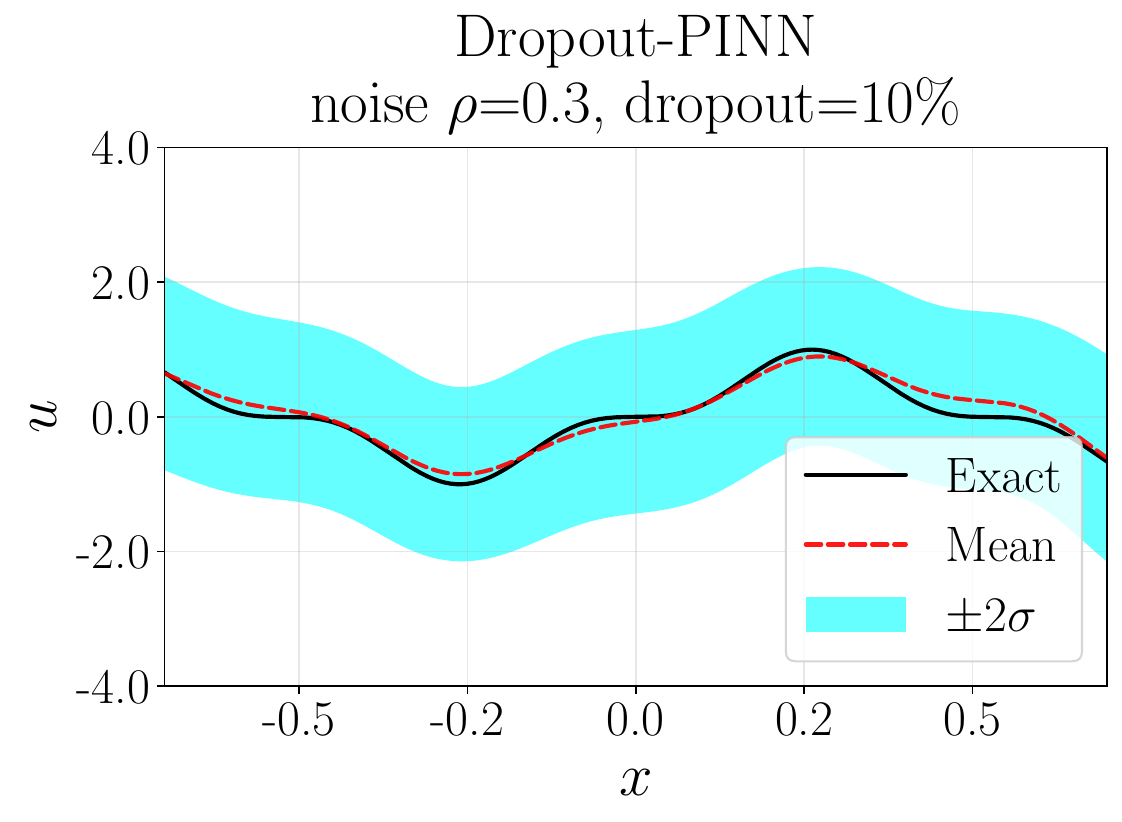}
        \caption{}
    \end{subfigure}
    \hfill
    \begin{subfigure}[b]{0.45\textwidth}
        \centering
        \incfig[width=\textwidth]{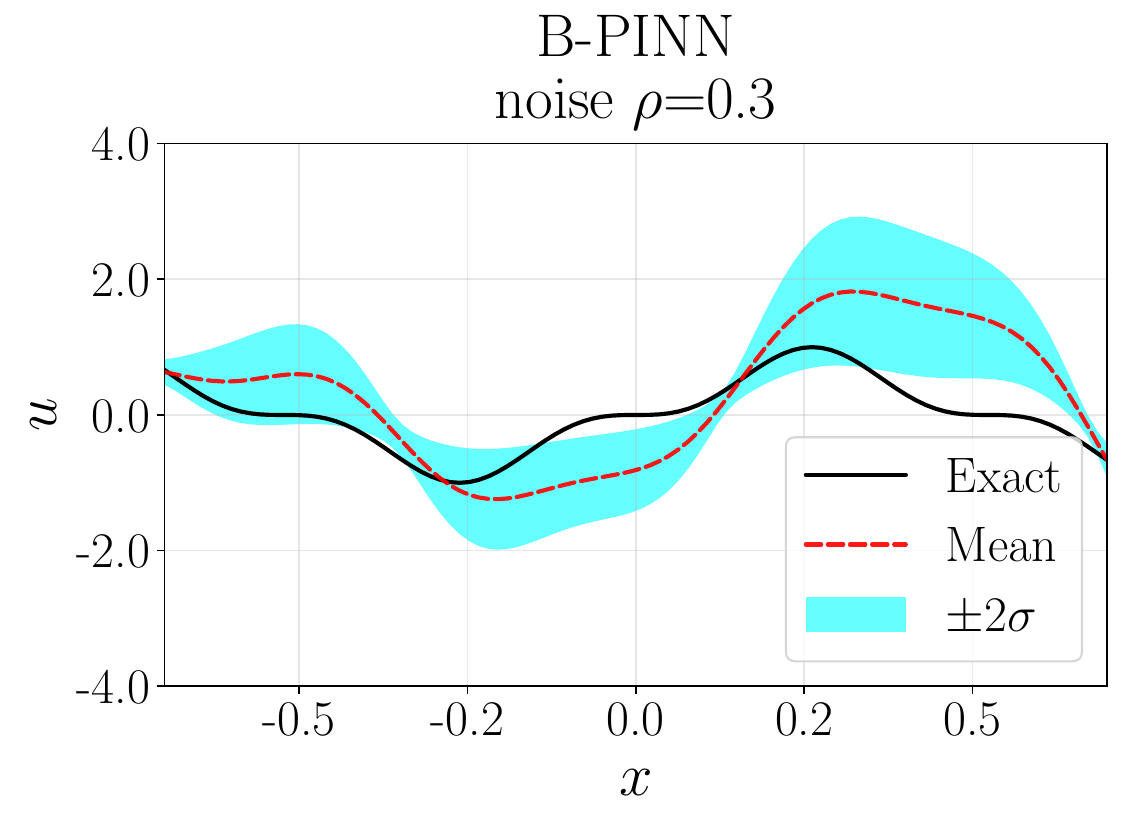}
        \caption{}
    \end{subfigure}
    
\caption{Mean and uncertainty predictions for $u$ in 1D nonlinear Poisson (Gaussian noise, $\sigma = 0.30\,\lVert u\rVert_{\infty}$) using (a) E-PINN (b) Dropout-PINN (5\%) (c) Dropout-PINN (10\%) (d) B-PINN.}
    \label{fig:1DNon-linPoissons_0.1}
\end{figure}

\begin{table}[H]
\centering
\caption{1D nonlinear Poisson: metrics vs.\ noise scale $\rho$ (standard deviation $\sigma=\rho\,\lVert u\rVert_{\infty}$). RMSE is computed with the predictive mean $\mu_u$. Time (s) is total wall-clock time; Time (epinet) (s) reports the E-PINN epinet-only training time (only populated for E-PINN rows). For $\rho=0$ (physics-only), B-PINN is not evaluated. (Sharpness/time/RMSE: lower is better; coverage: higher is better.)}
\label{tab:nonlin_poisson_all}
\sisetup{detect-weight=true,detect-family=true,mode=text}
\begin{tabular*}{\textwidth}{@{\extracolsep{\fill}}
S[table-format=1.2]
l
S[table-format=1.2]
S[table-format=1.2]
S[table-format=1.4]
S[table-format=4.2]
 c
}
\toprule
\multicolumn{1}{c}{$\rho$} & \multicolumn{1}{c}{Method} &
\multicolumn{1}{c}{Sharpness $\mu^w_{2\sigma}$} &
\multicolumn{1}{c}{Coverage (95\%)} &
\multicolumn{1}{c}{RMSE} &
\multicolumn{1}{c}{Time (s)} & \multicolumn{1}{c}{Time (epinet) (s)} \\
\midrule
0.00 & E-PINN       & \bfseries 0.20 & \bfseries 1.00 & \bfseries 0.0065 & 601.96 & 128.14 \\
0.00 & Dropout 5\%  & 2.05             & \bfseries 1.00 & 0.0256              & 391.66 & -- \\
0.00 & Dropout 10\% & 2.97             & \bfseries 1.00 & 0.0516              & \bfseries 387.82 & -- \\
\midrule
0.10 & E-PINN       & \bfseries 0.16 & \bfseries 1.00 & \bfseries 0.0123 & 611.54 & 141.42 \\
0.10 & B-PINN       & 0.61             & 1.00              & 0.0314              & 1719.08 & -- \\
0.10 & Dropout 5\%  & 2.07             & \bfseries 1.00 & 0.0416              & \bfseries 400.50 & -- \\
0.10 & Dropout 10\% & 2.96             & \bfseries 1.00 & 0.0644              & 557.71 & -- \\
\midrule
0.30 & E-PINN       & \bfseries 0.34 & \bfseries 1.00 & \bfseries 0.0494 & 552.38 & 132.48 \\
0.30 & B-PINN       & 1.26             & 0.51              & 0.7618              & 1763.16 & -- \\
0.30 & Dropout 5\%  & 1.90             & \bfseries 1.00 & 0.1110              & 500.60 & -- \\
0.30 & Dropout 10\% & 2.72             & \bfseries 1.00 & 0.1196              & \bfseries 369.77 & -- \\
\bottomrule
\end{tabular*}
\end{table}

Across these regimes (Table~\ref{tab:nonlin_poisson_all}), in the physics-only setting E-PINN yields calibrated, narrow bands ($\mu^w_{2\sigma}\approx0.20$) with the smallest error (RMSE $\approx6.5\times10^{-3}$). With interior $u$ data at $\rho=0.10$ and $\rho=0.30$, E-PINN remains sharp with full coverage and lower RMSE than the alternatives. Dropout-PINN produces substantially wider intervals (about $1.9$ to $3.0$) and larger errors across noise levels. In the B-PINN runs, coverage equals 1.00 at $\rho=0.10$ with moderate error (RMSE $\approx3.1\times10^{-2}$), while at $\rho=0.30$ coverage drops ($\approx0.51$) and RMSE increases ($\approx7.6\times10^{-1}$). This failure at $\rho=0.30$ reflects posterior pathologies under the fixed HMC configuration on a stiff, nonlinear residual: under-dispersion and poor exploration yield under-coverage and a biased mean. Smaller step sizes, better preconditioning (e.g., mass adaptation), or more samples could alleviate this at the cost of higher computational cost. The last column reports the E-PINN epinet-only training time, the incremental cost when a base PINN is already trained.

\subsubsection{1D flow through a porous medium}

We consider steady flow through a horizontal channel filled with a uniform porous medium. Let $x\in[0,1]$ denote the wall-normal coordinate and $u(x)$ the streamwise velocity. The dynamics are governed by the Darcy-Brinkman equation \cite{yu2013darcy, martys2001improved},

\begin{equation}
    -\frac{\nu_e}{\phi}\frac{\partial^2 u}{\partial x^2} + \frac{\nu u}{K} = f(x),\quad x \in [0, 1],
\end{equation}
where $(\nu_e/\phi)\,\partial^2_x u$ represents effective viscous diffusion in the porous matrix, $(\nu/K)\,u$ is the Darcy drag, and $f(x)$ is an external body force that absorbs the streamwise pressure gradient.

The exact solution is given by \cite{yang2021b,yu2013darcy}:
\begin{equation}
    u(x) = \frac{f K}{\nu} \left[ 1 - \frac{\cosh\left(r(x - \frac{H}{2})\right)}{\cosh\left(\frac{rH}{2}\right)} \right],
\end{equation}

and 
\begin{equation}
    r = \sqrt{\frac{\nu \phi}{\nu_e K}}.
\end{equation}

For this case, we take the effective viscosity as $\nu_e=10^{-3}$, fluid viscosity $\nu=10^{-3}$, porosity $\phi=0.4$, permeability $K=10^{-3}$, and channel height $H=1$. We impose no-slip Dirichlet boundary conditions, $u(0)=u(1)=0$; boundary values are treated as exact (noise-free) and are enforced via the boundary condition loss with weight $w_{\text{bc}}=10$. The base PINN is trained using 64 collocation points uniformly distributed in $x \in [0, 1]$ with exact boundary conditions. We evaluate two scenarios: (i) physics-only (PDE + exact BC; no interior data) and (ii) physics + noisy $u$ data at 32 interior sensors, where noise is injected on $u$ as a fraction of $\lVert u \rVert_{\infty}$ (fractions $0.10$ and $0.30$).

\begin{figure}[H]
    \centering
    \begin{subfigure}[b]{0.45\textwidth}
        \centering
        \incfig[width=\textwidth]{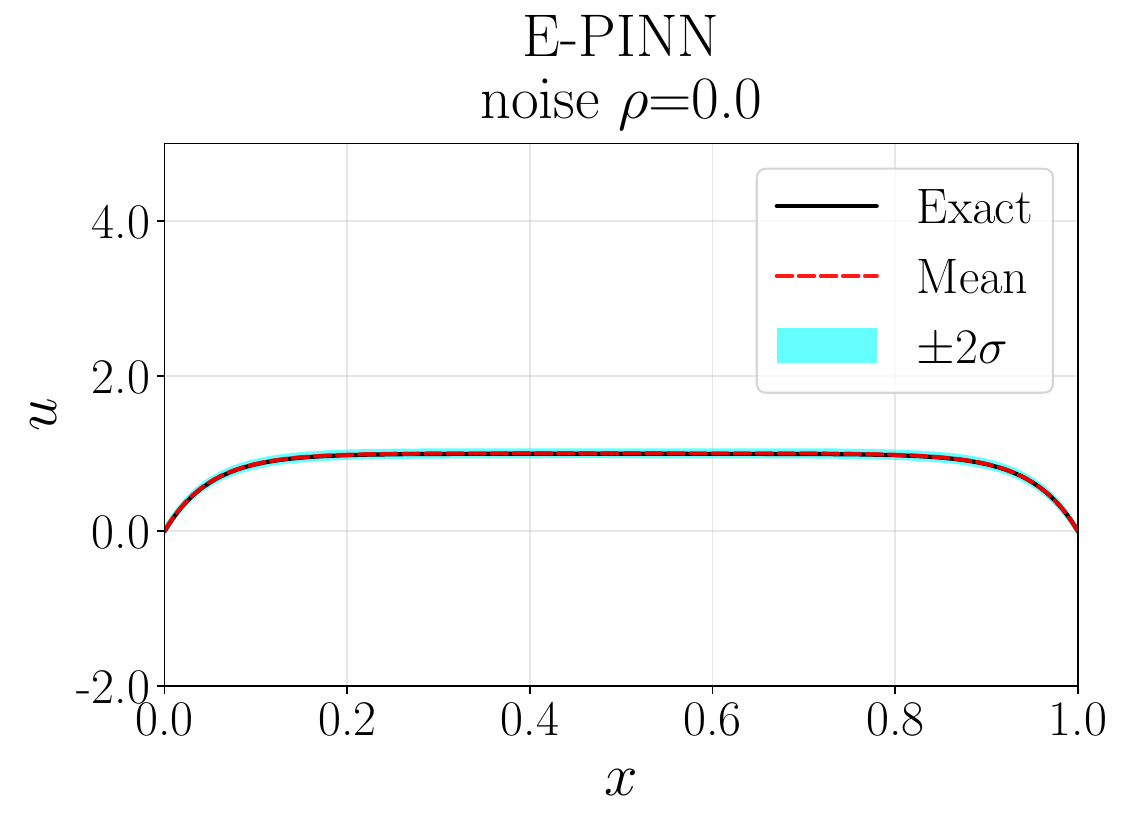}
        \caption{}
    \end{subfigure}
    \hfill
    \begin{subfigure}[b]{0.45\textwidth}
        \centering
        \incfig[width=\textwidth]{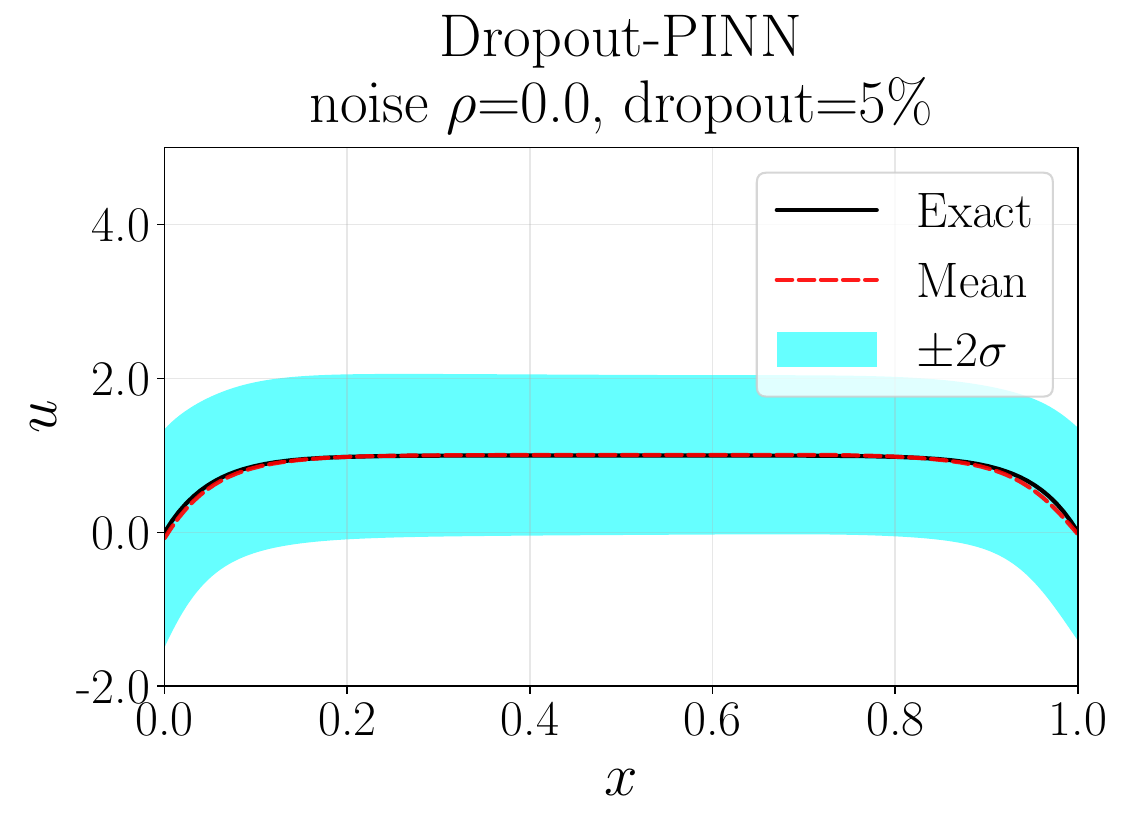}
        \caption{}
    \end{subfigure}
    
    \begin{subfigure}[b]{0.45\textwidth}
        \centering
        \incfig[width=\textwidth]{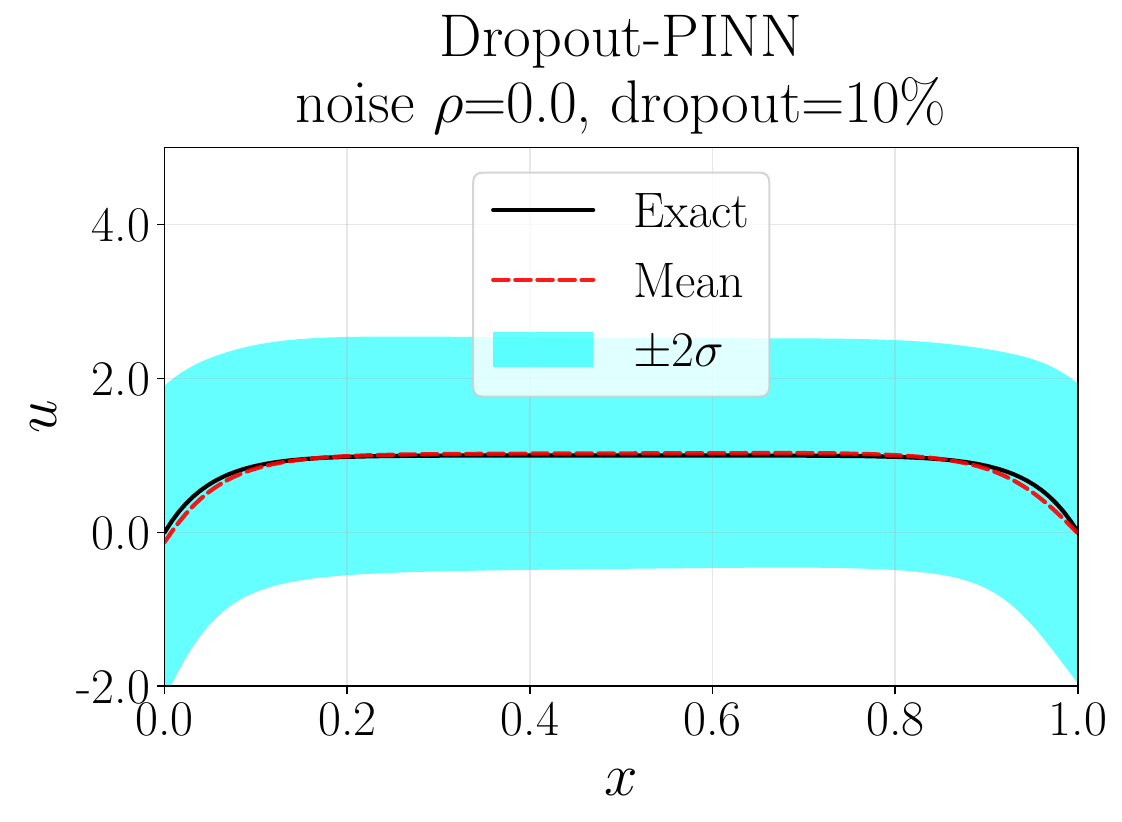}
        \caption{}
    \end{subfigure}

\caption{Mean and uncertainty predictions for $u$ in a 1D porous medium (physics-only, no interior data) using (a) E-PINN (b) Dropout-PINN (5\%) (c) Dropout-PINN (10\%). B-PINN is not evaluated in physics-only.}
    \label{fig:1DPorousMedium_0.0}
\end{figure}

\begin{figure}[H]
    \centering
    \begin{subfigure}[b]{0.45\textwidth}
        \centering
        \incfig[width=\textwidth]{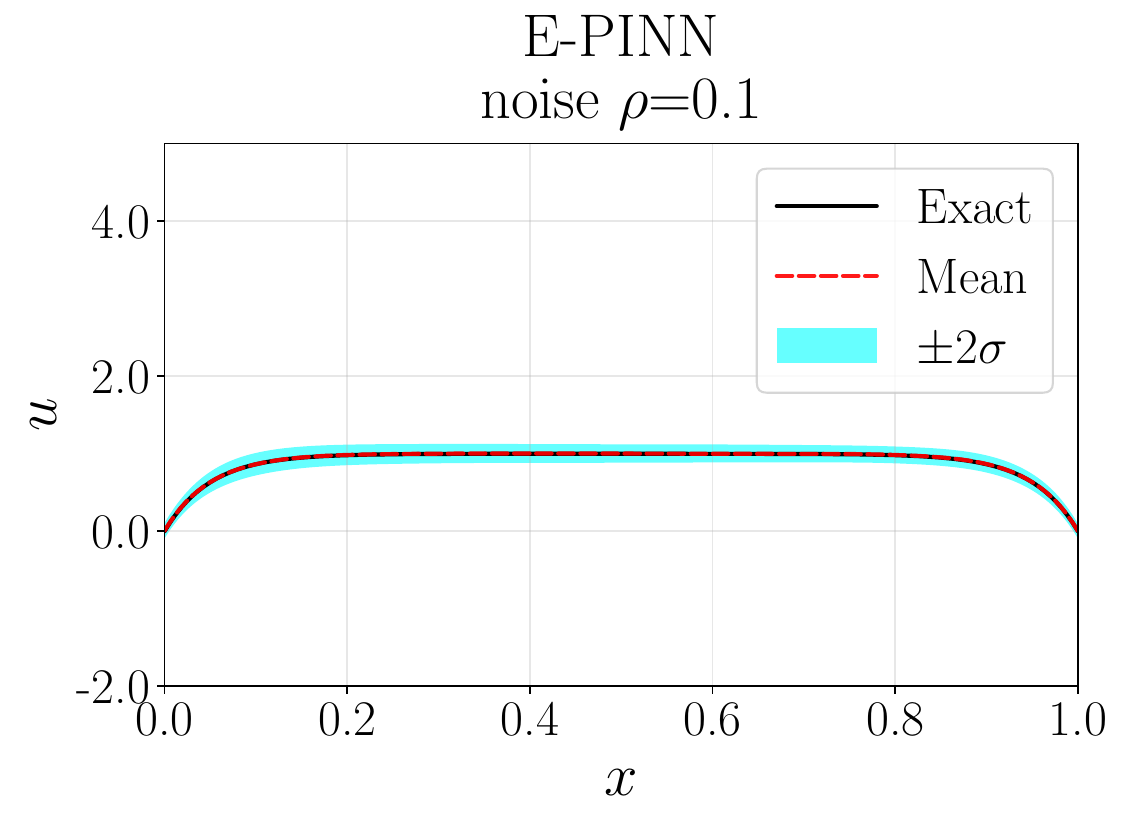}
        \caption{}
    \end{subfigure}
    \hfill
    \begin{subfigure}[b]{0.45\textwidth}
        \centering
        \incfig[width=\textwidth]{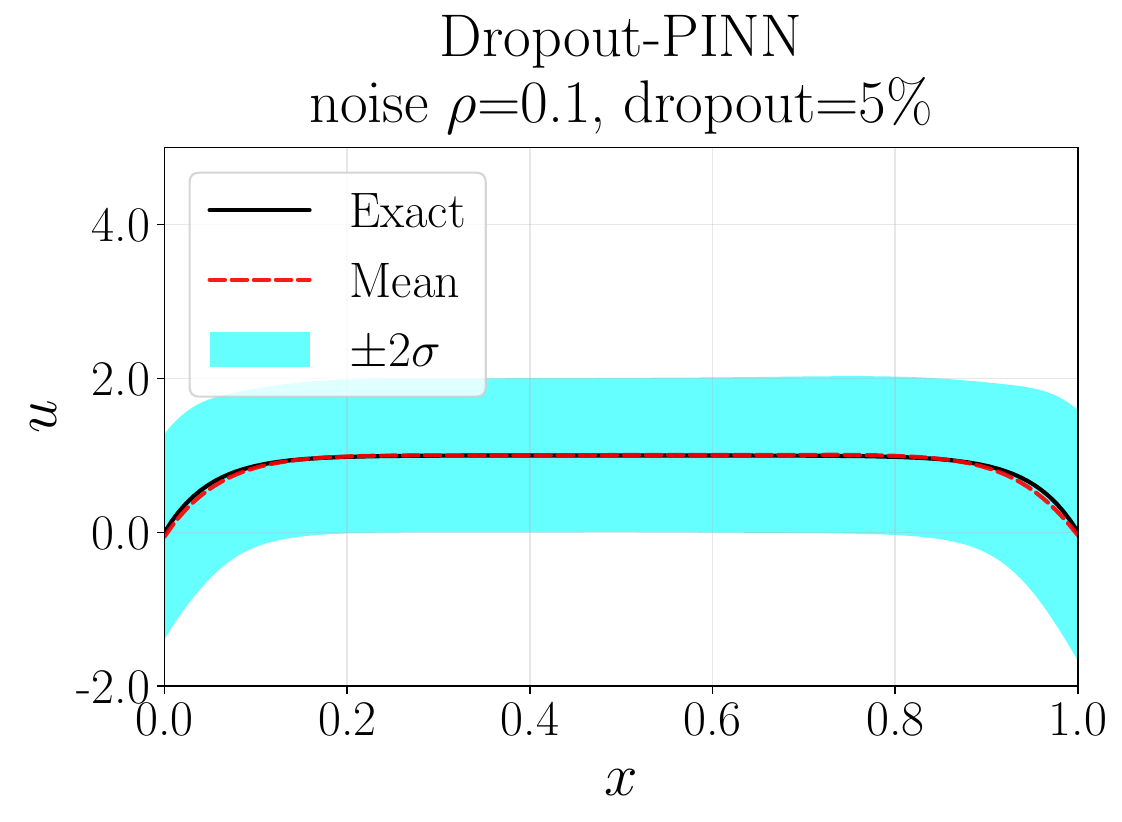}
        \caption{}
    \end{subfigure}
        
    \begin{subfigure}[b]{0.45\textwidth}
        \centering
        \incfig[width=\textwidth]{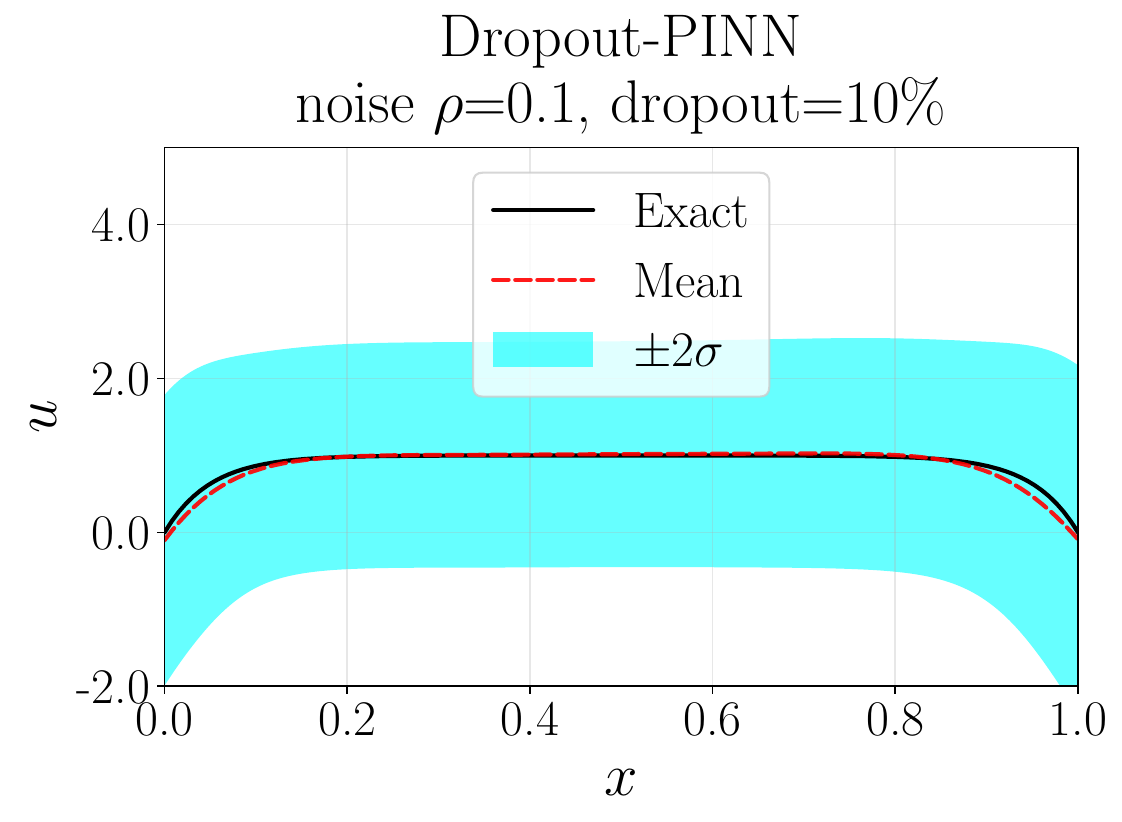}
        \caption{}
    \end{subfigure}
    \hfill
    \begin{subfigure}[b]{0.45\textwidth}
        \centering
        \incfig[width=\textwidth]{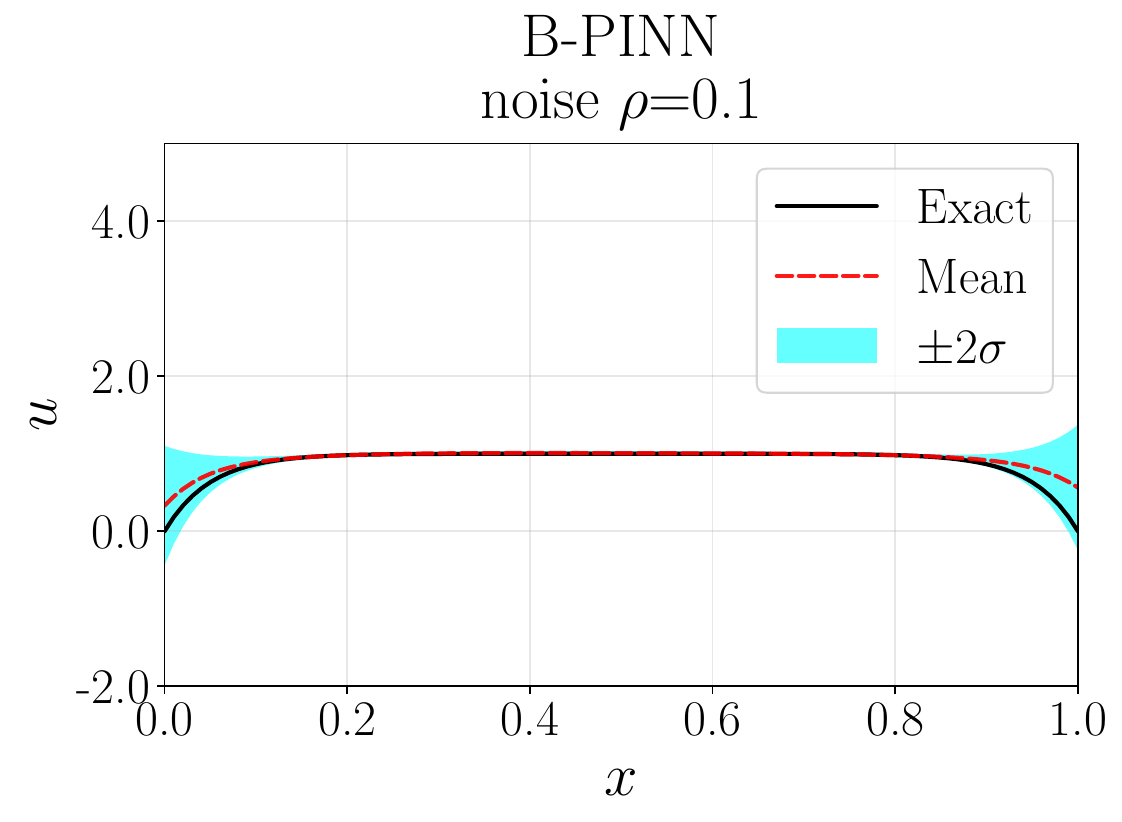}
        \caption{}
    \end{subfigure}
    
\caption{Mean and uncertainty predictions for $u$ in a 1D porous medium (Gaussian noise, $\sigma = 0.10\,\lVert u\rVert_{\infty}$) using (a) E-PINN (b) Dropout-PINN (5\%) (c) Dropout-PINN (10\%) (d) B-PINN.}
    \label{fig:1DPorousMedium_0.1}
\end{figure}

Figures~\ref{fig:1DPorousMedium_0.0} (physics-only) and \ref{fig:1DPorousMedium_0.1} ($\rho=0.10$) illustrate the mean and uncertainty for $u$. In the physics-only case, E-PINN and Dropout-PINN recover the profile, including steep boundary-layer gradients, with uncertainty bands that widen away from the walls; Dropout-PINN bands grow with the dropout rate. With noisy interior sensors ($\rho=0.10$), means degrade across methods near sharp gradients; E-PINN remains sharp, Dropout-PINN depends on the dropout rate, and B-PINN attains slightly narrower bands with higher RMSE.

Table~\ref{tab:porous_all} reports sharpness, coverage, RMSE, and time. In physics-only ($\rho=0$), E-PINN attains narrow bands ($\mu^w_{2\sigma}\approx0.132$) with full coverage and low RMSE ($\approx0.0055$); Dropout-PINN is much wider (about 2.16 to 3.13) with higher RMSE (about 0.027 to 0.048). With interior $u$ data ($\rho=0.10$ and $\rho=0.30$), E-PINN remains calibrated and sharp (about 0.239 and 0.182) with low RMSE (about 0.0034 and 0.013) at total times of about $4.5\times10^2$ to $4.8\times10^2$ s. B-PINN achieves slightly smaller sharpness (about 0.183 at both noise levels) but with markedly higher RMSE and higher total time. In these data-augmented runs, the B-PINN bands are narrow but centered on inaccurate means, as reflected by the high RMSE despite full coverage. Dropout-PINN produces the widest intervals with higher RMSE. The last column lists epinet-only training time for E-PINN (about 133 to 198 s), the incremental cost when reusing a trained base model.

Relative to \cite{yang2021b}, the B-PINN discrepancy arises from the noise model: their tests perturb the source and boundary measurements, while here noise is applied only to interior $u$ sensors with no added noise on the physics and boundary channels (in B-PINN likelihoods we use a small residual noise level of 0.01 for stability). Direct noise on $u$ leads to broader posteriors under the same PDE constraints, consistent with Table~\ref{tab:porous_all}.

\begin{table}[H]
\centering
\caption{1D porous medium: metrics vs.\ noise scale $\rho$ (standard deviation $\sigma=\rho\,\lVert u\rVert_{\infty}$). RMSE is computed with the predictive mean $\mu_u$. Time (s) is total wall-clock time; Time (epinet) (s) reports the E-PINN epinet-only training time (only populated for E-PINN rows). For $\rho=0$ (physics-only), B-PINN is not evaluated. (Sharpness/time/RMSE: lower is better; coverage: higher is better.)}
\label{tab:porous_all}
\sisetup{detect-weight=true,detect-family=true,mode=text}
\begin{tabular*}{\textwidth}{@{\extracolsep{\fill}}
S[table-format=1.2]
l
S[table-format=1.2]
S[table-format=1.2]
S[table-format=1.4]
S[table-format=4.2]
 c
}
\toprule
\multicolumn{1}{c}{$\rho$} & \multicolumn{1}{c}{Method} &
\multicolumn{1}{c}{Sharpness $\mu^w_{2\sigma}$} &
\multicolumn{1}{c}{Coverage (95\%)} &
\multicolumn{1}{c}{RMSE} &
\multicolumn{1}{c}{Time (s)} & \multicolumn{1}{c}{Time (epinet) (s)} \\
\midrule
0.00 & E-PINN       & \bfseries 0.13 & \bfseries 1.00 & \bfseries 0.0055 & 508.94 & 197.62 \\
0.00 & Dropout 5\%  & 2.16             & \bfseries 1.00 & 0.0275              & 324.76 & -- \\
0.00 & Dropout 10\% & 3.13             & \bfseries 1.00 & 0.0478              & \bfseries 322.63 & -- \\
\midrule
0.10 & E-PINN       & 0.24             & \bfseries 1.00 & \bfseries 0.0034 & 448.23 & 133.26 \\
0.10 & B-PINN       & \bfseries 0.18 & \bfseries 1.00 & 0.1087              & 1703.46 & -- \\
0.10 & Dropout 5\%  & 2.13             & \bfseries 1.00 & 0.0300              & \bfseries 361.57 & -- \\
0.10 & Dropout 10\% & 3.12             & \bfseries 1.00 & 0.0618              & 460.12 & -- \\
\midrule
0.30 & E-PINN       & 0.18             & \bfseries 1.00 & \bfseries 0.0132 & 478.58 & 136.90 \\
0.30 & B-PINN       & \bfseries 0.18 & \bfseries 1.00 & 0.1113              & 1786.83 & -- \\
0.30 & Dropout 5\%  & 2.09             & \bfseries 1.00 & 0.0378              & \bfseries 350.92 & -- \\
0.30 & Dropout 10\% & 3.05             & \bfseries 1.00 & 0.0663              &  358.40 & -- \\
\bottomrule
\end{tabular*}
\end{table}

\subsubsection{2D nonlinear Poisson equation}
We assess scalability beyond one dimension on a 2D nonlinear Poisson problem posed on $\Omega=[-1,1]^2$. We set the epinet prior scaling to $\alpha=0.05$. The governing equation is
\begin{equation}
\label{eq:2d_nl_poisson}
    \lambda\left( \frac{\partial^2 u}{\partial x^2} + \frac{\partial^2 u}{\partial y^2} \right) + u\,(u^2 - 1) = f(x,y), \qquad (x,y)\in\Omega, \quad \lambda = 0.01,
\end{equation}
with Dirichlet boundary conditions on $\partial\Omega$. For evaluation, we adopt the manufactured (exact) solution
\begin{equation}
    u(x,y) = \sin(\pi x)\,\sin(\pi y),
\end{equation}
and obtain $f(x,y)$ by substituting this expression into \eqref{eq:2d_nl_poisson}. In direct analogy with the 1D cases, we consider two regimes: (i) physics-only (PDE + exact boundary values; no interior data), and (ii) physics augmented with noisy interior measurements of $u$ at a sparse set of sensors, where the noise level is a fraction of $\lVert u \rVert_{\infty}$ (fractions $0.10$, $0.30$). We report sharpness (mean width of the $\pm 2\sigma$ band), empirical 95\% coverage, and RMSE on dense grids. For this case, we report B-PINN in the data-augmented setting.

\begin{figure}[H]
    \centering
    \begin{subfigure}[b]{0.32\textwidth}\centering
        \incfig[width=\textwidth]{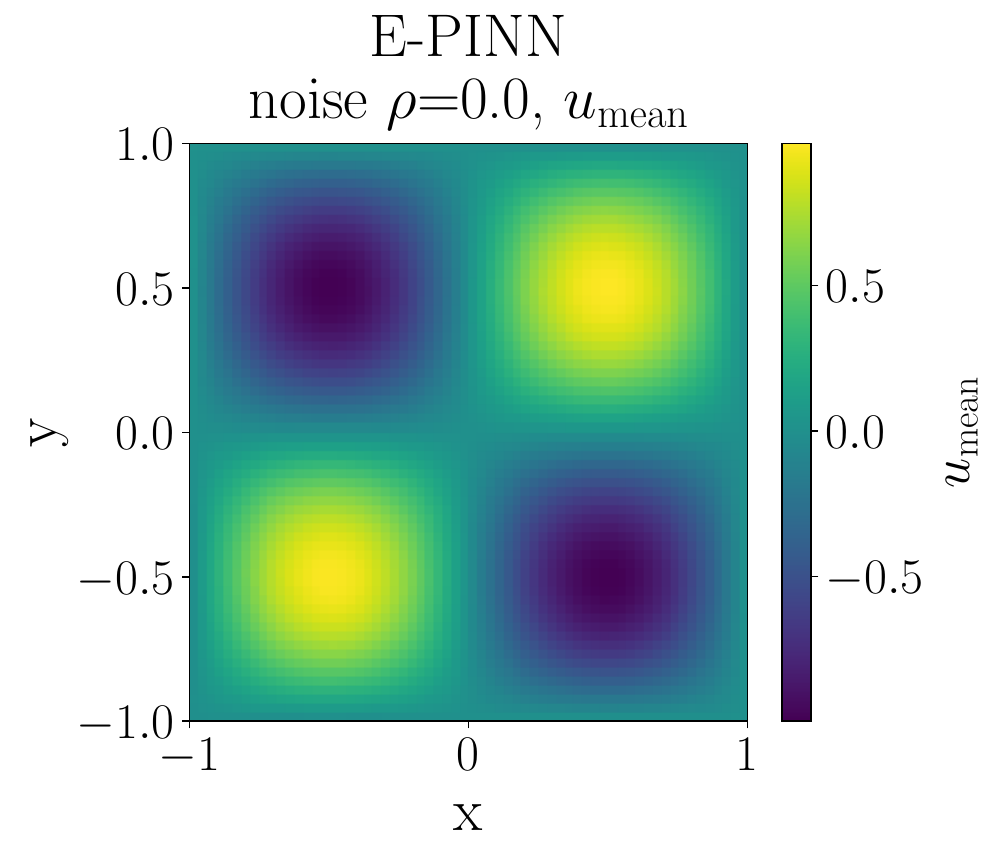}
    \end{subfigure}\hfill
    \begin{subfigure}[b]{0.32\textwidth}\centering
        \incfig[width=\textwidth]{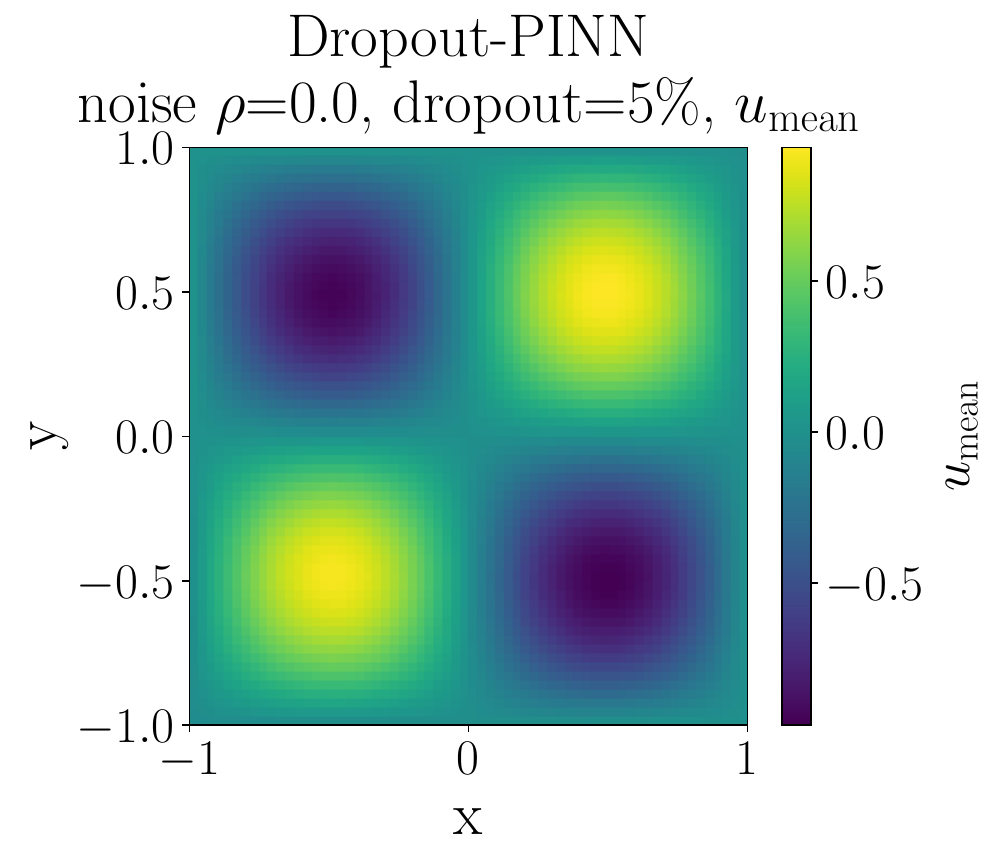}
    \end{subfigure}\hfill
    \begin{subfigure}[b]{0.32\textwidth}\centering
        \incfig[width=\textwidth]{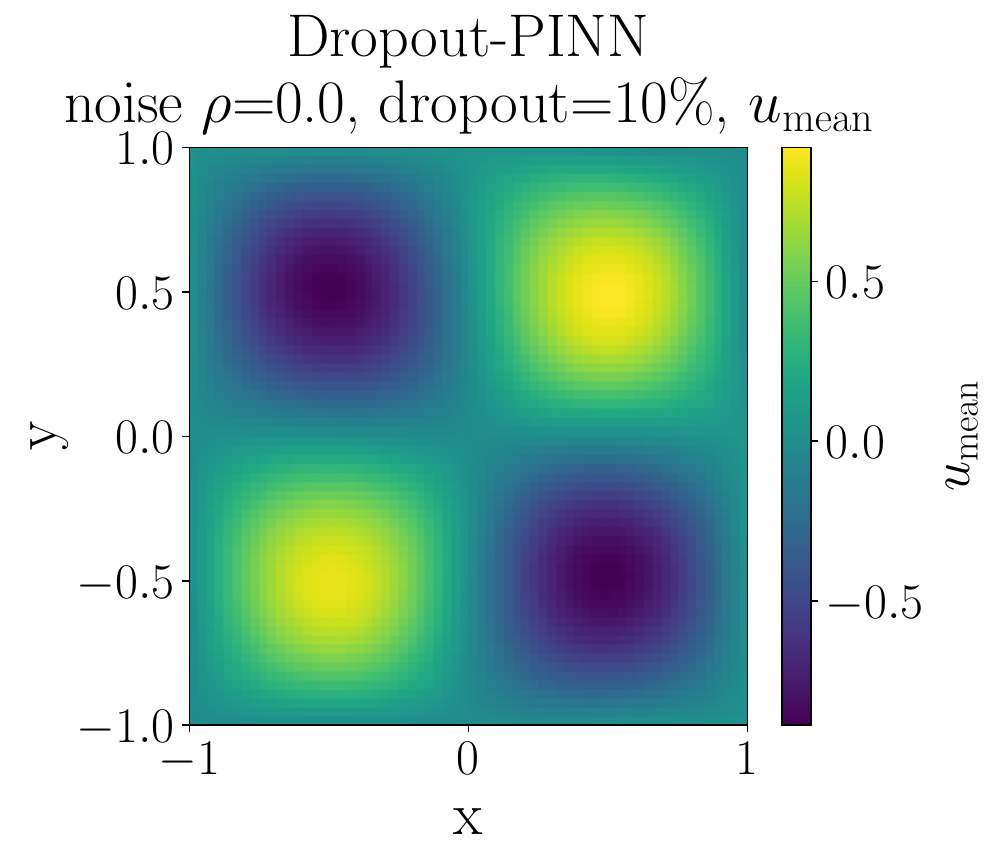}
    \end{subfigure}
    \par\vspace{0.4em}
    \begin{subfigure}[b]{0.32\textwidth}\centering
        \incfig[width=\textwidth]{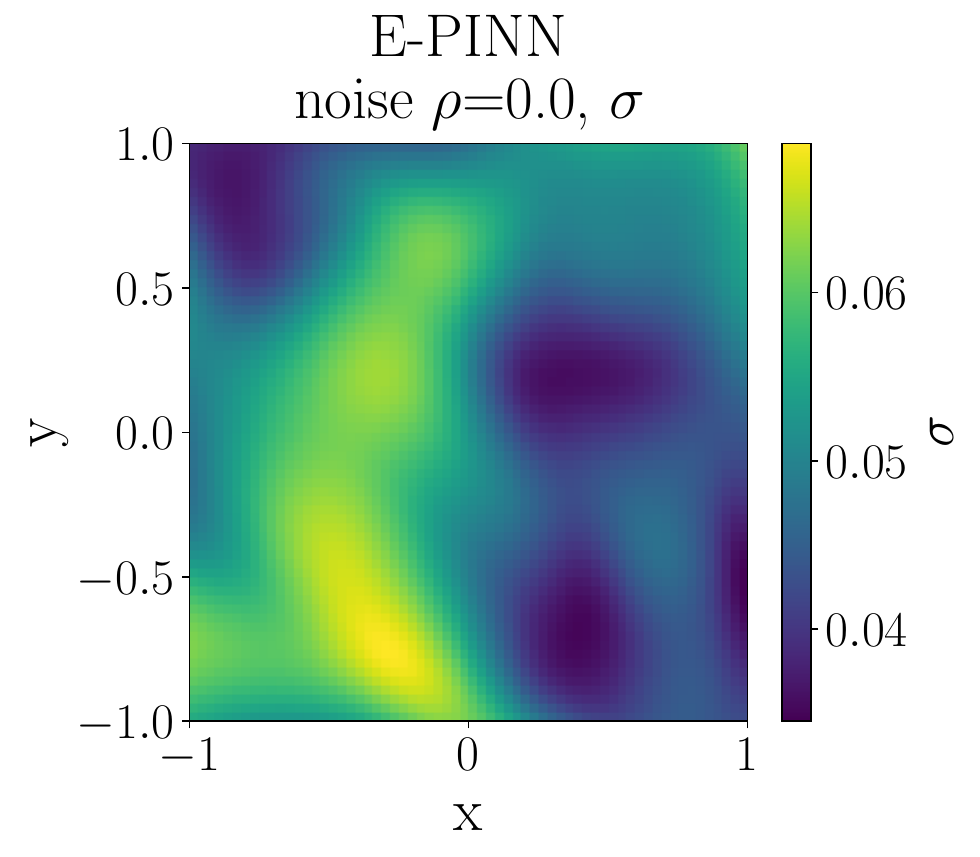}
    \end{subfigure}\hfill
    \begin{subfigure}[b]{0.32\textwidth}\centering
        \incfig[width=\textwidth]{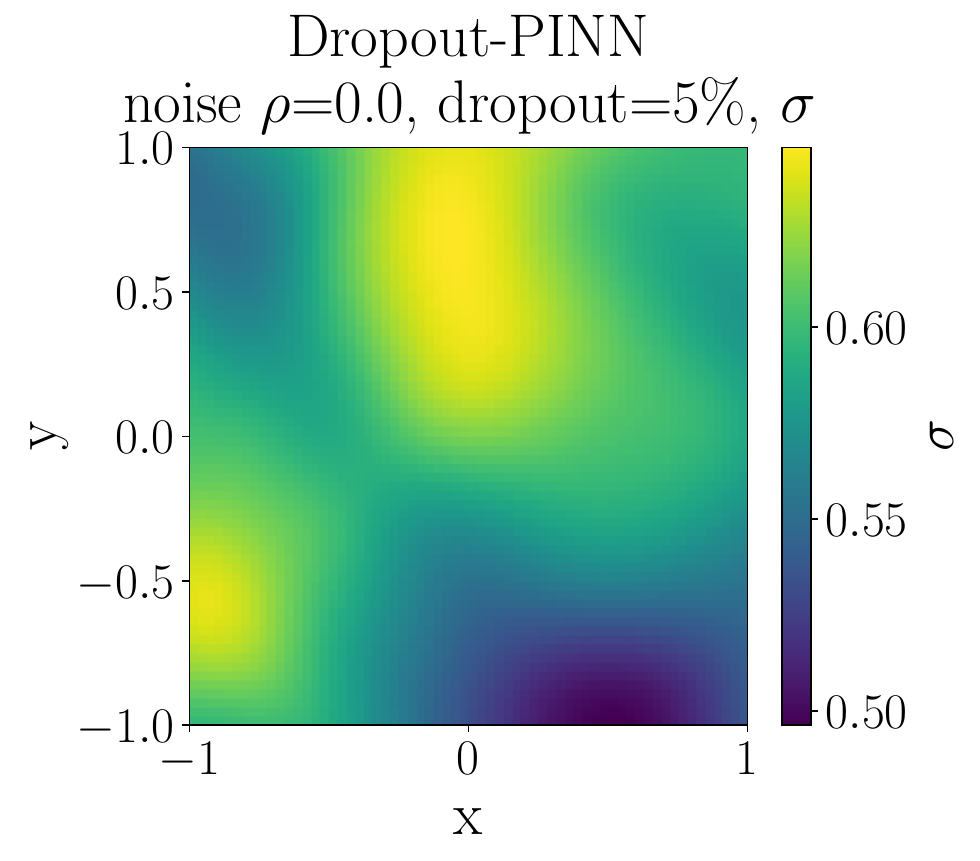}
    \end{subfigure}\hfill
    \begin{subfigure}[b]{0.32\textwidth}\centering
        \incfig[width=\textwidth]{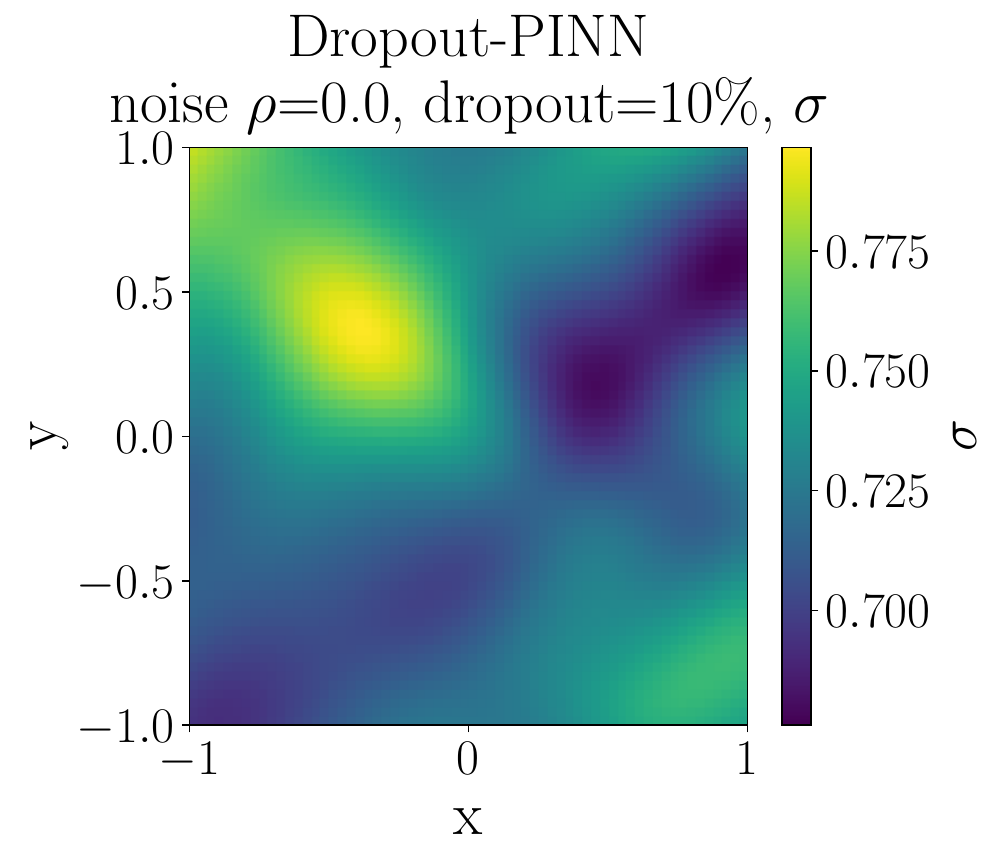}
    \end{subfigure}
    \par\vspace{0.4em}
    \begin{subfigure}[b]{0.32\textwidth}\centering
        \incfig[width=\textwidth]{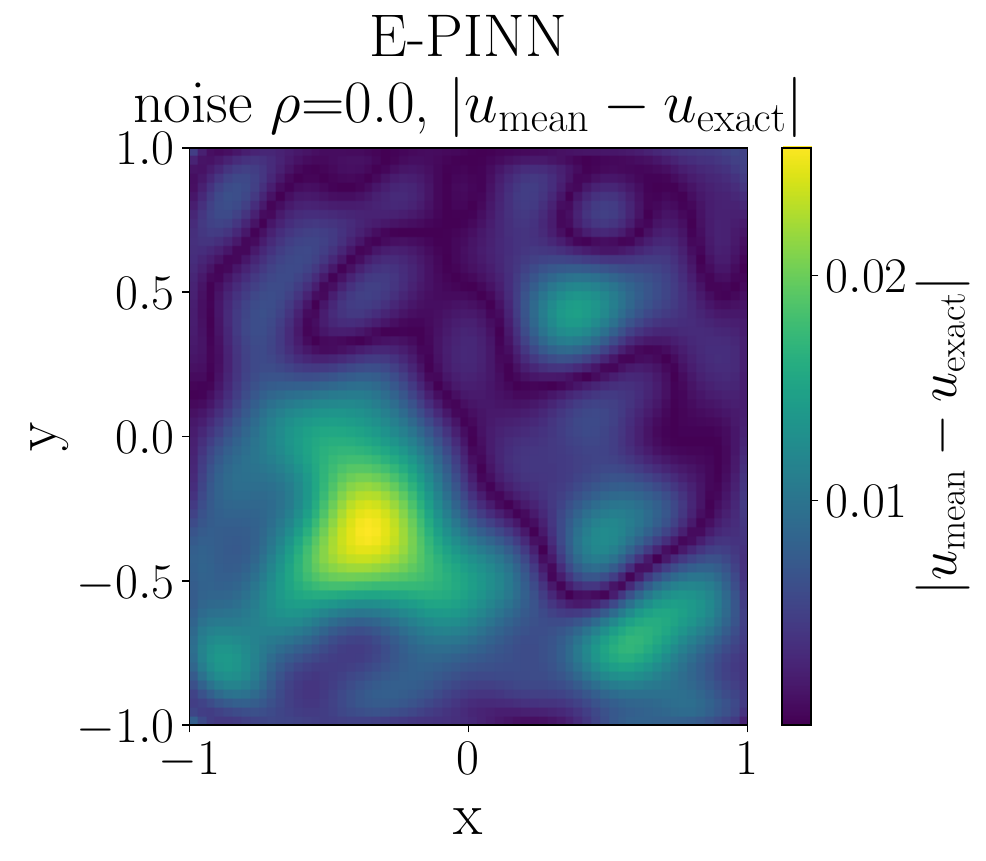}
    \end{subfigure}\hfill
    \begin{subfigure}[b]{0.32\textwidth}\centering
        \incfig[width=\textwidth]{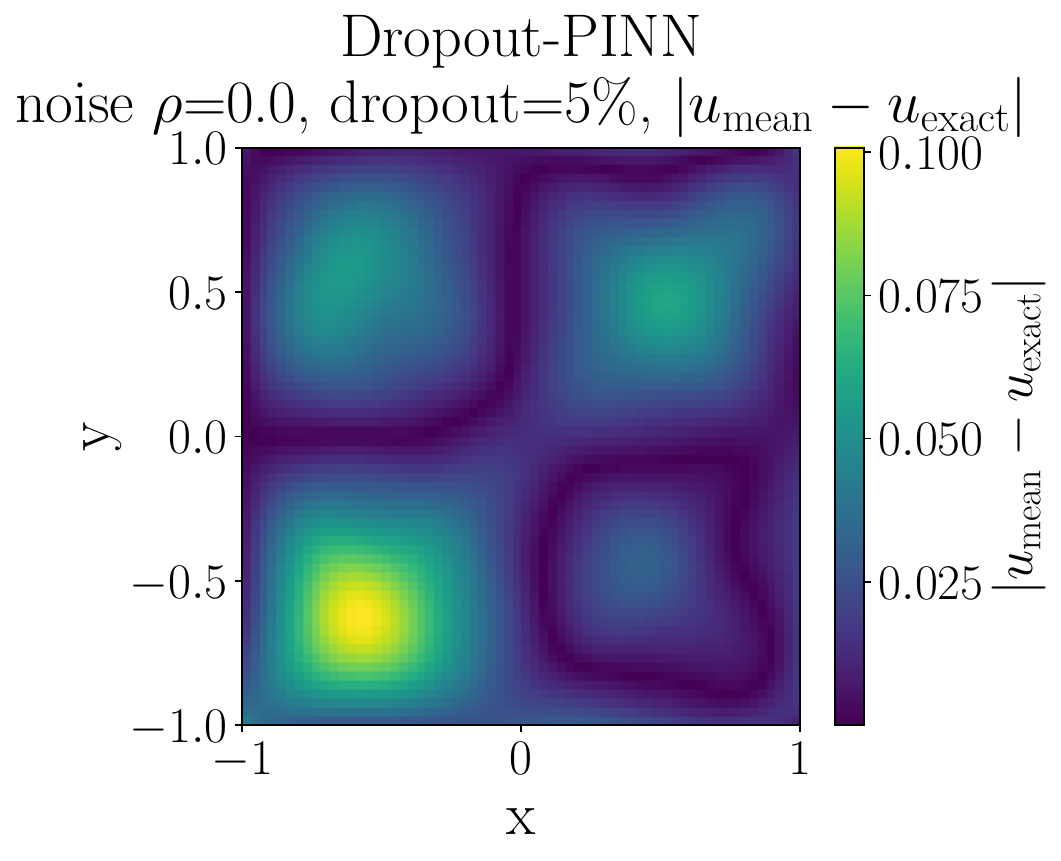}
    \end{subfigure}\hfill
    \begin{subfigure}[b]{0.32\textwidth}\centering
        \incfig[width=\textwidth]{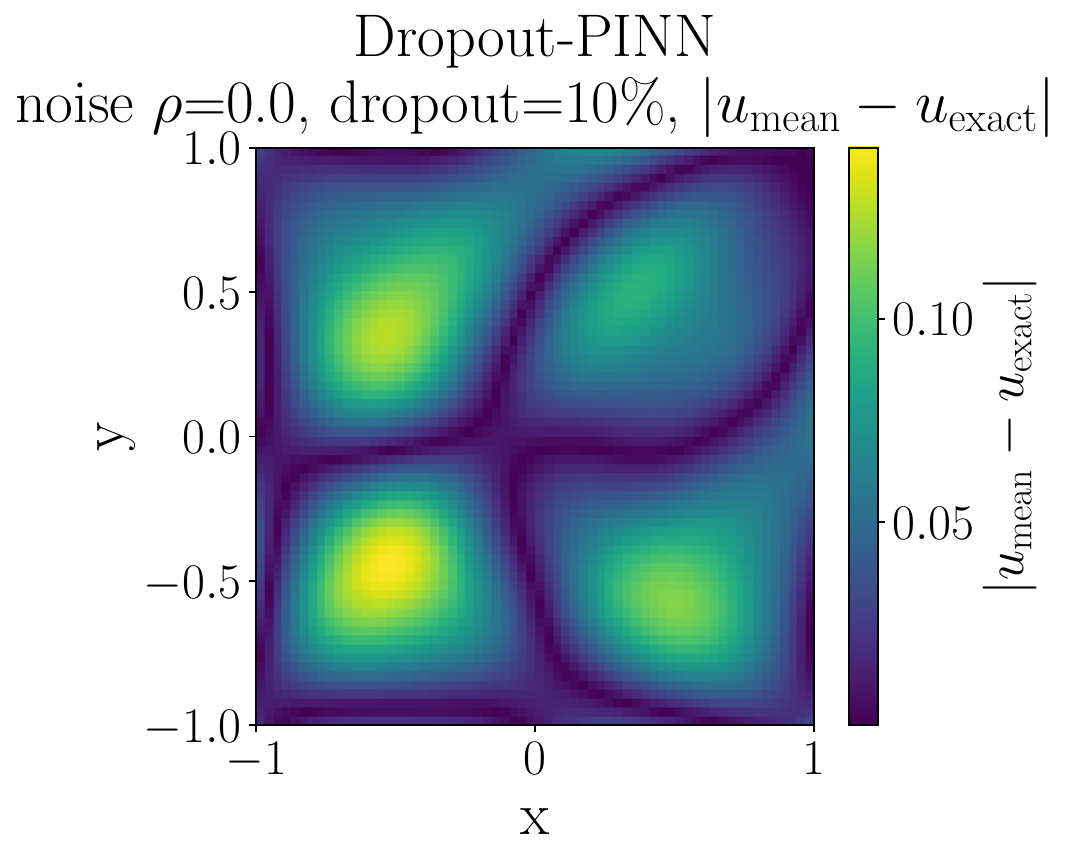}
    \end{subfigure}
    \caption{2D nonlinear Poisson (physics-only): top, mean $u$; middle, epistemic standard deviation; bottom, absolute error. Columns: E-PINN (left), Dropout-PINN (5\%) (center), Dropout-PINN (10\%) (right).}
    \label{fig:2D_nlP_grid_0.00}
\end{figure}

\begin{figure}[H]
    \centering
    \begin{subfigure}[b]{0.32\textwidth}\centering
        \incfig[width=\textwidth]{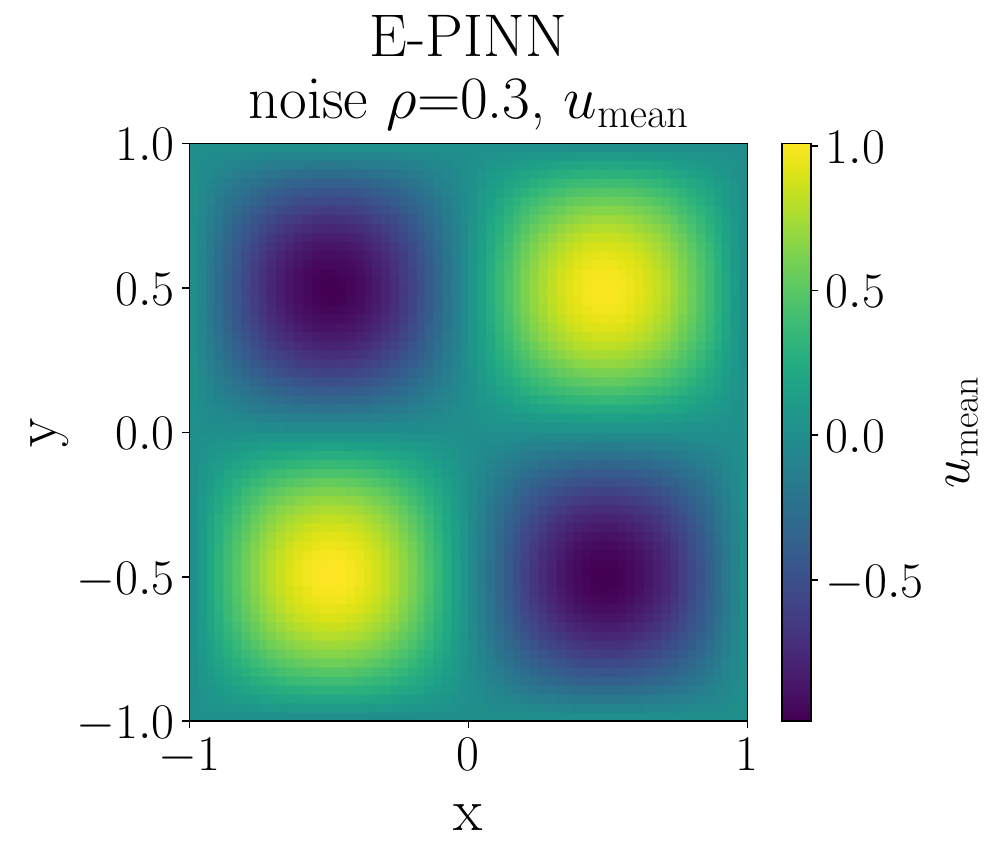}
    \end{subfigure}\hfill
    \begin{subfigure}[b]{0.32\textwidth}\centering
        \incfig[width=\textwidth]{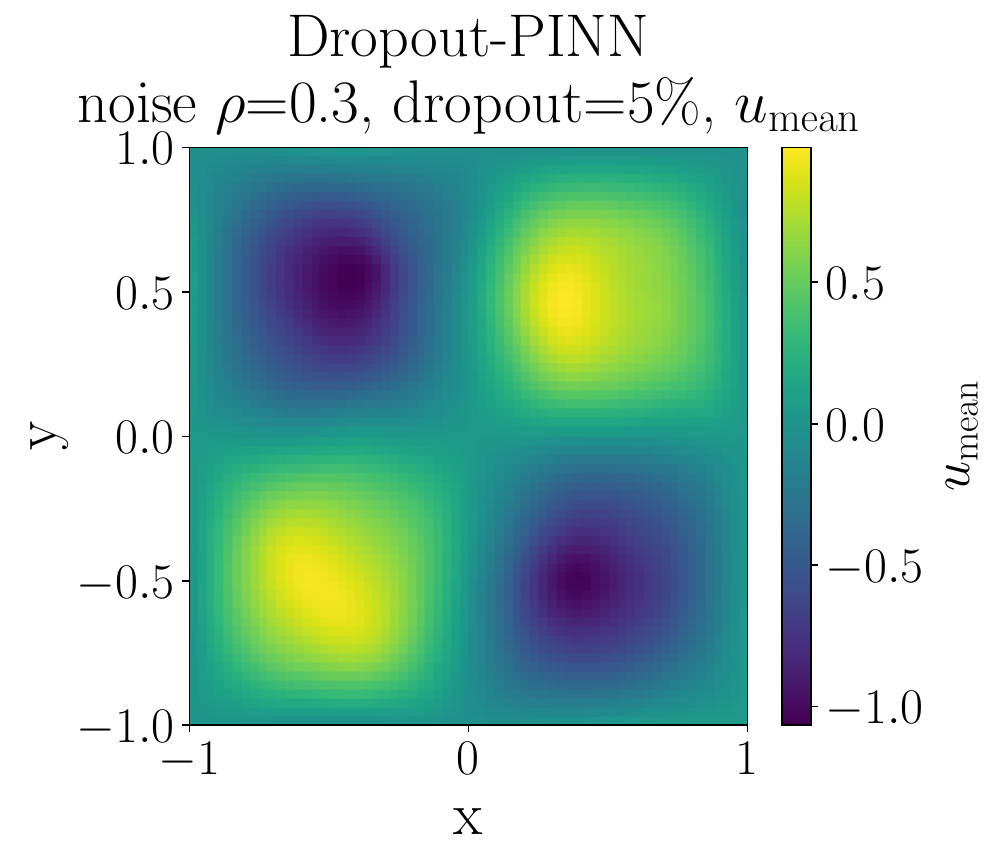}
    \end{subfigure}\hfill
    \begin{subfigure}[b]{0.32\textwidth}\centering
        \incfig[width=\textwidth]{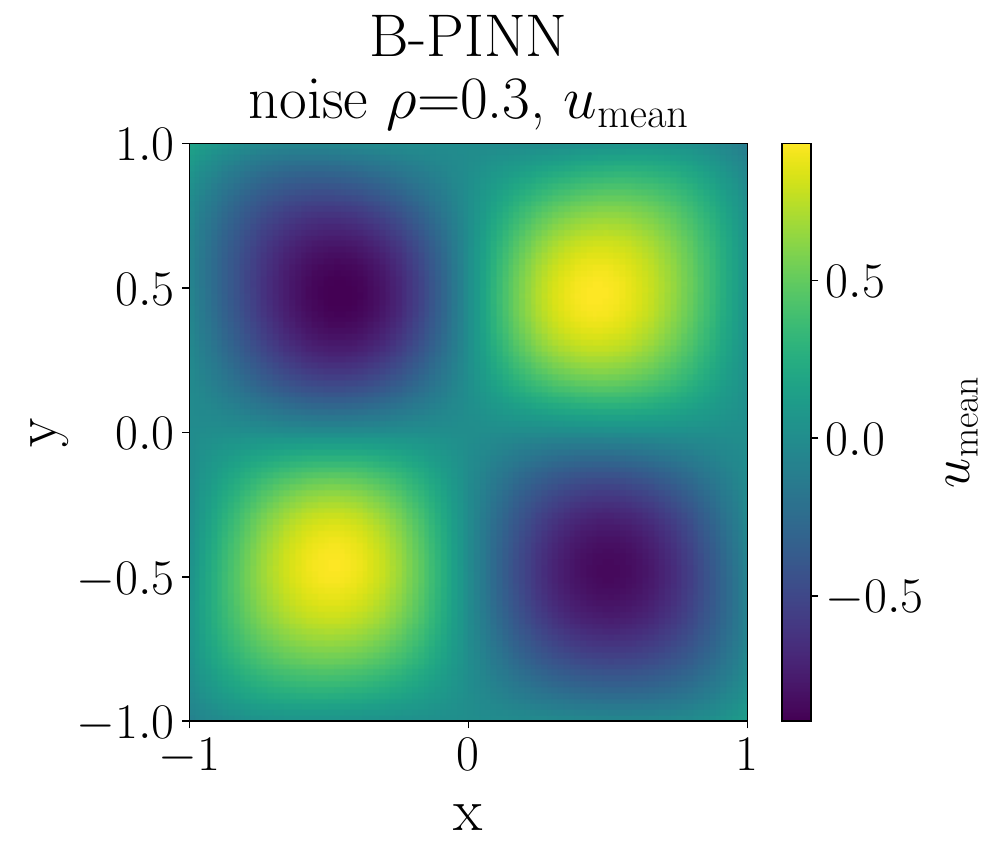}
    \end{subfigure}
    \par\vspace{0.4em}
    \begin{subfigure}[b]{0.32\textwidth}\centering
        \incfig[width=\textwidth]{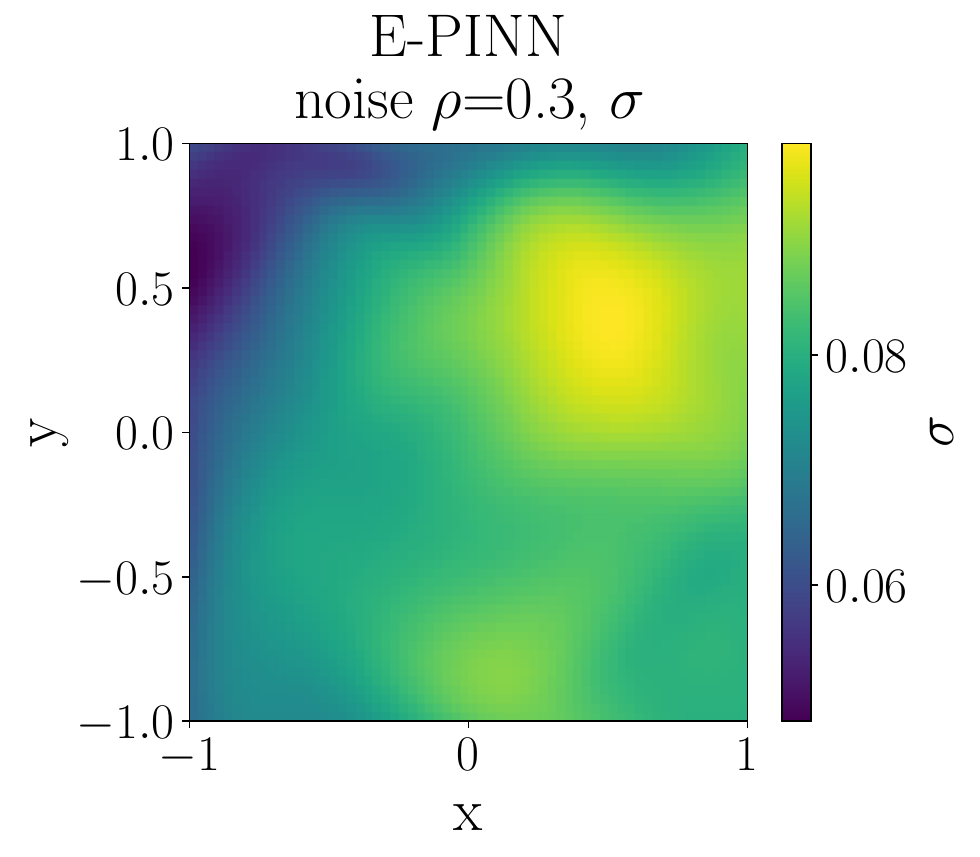}
    \end{subfigure}\hfill
    \begin{subfigure}[b]{0.32\textwidth}\centering
        \incfig[width=\textwidth]{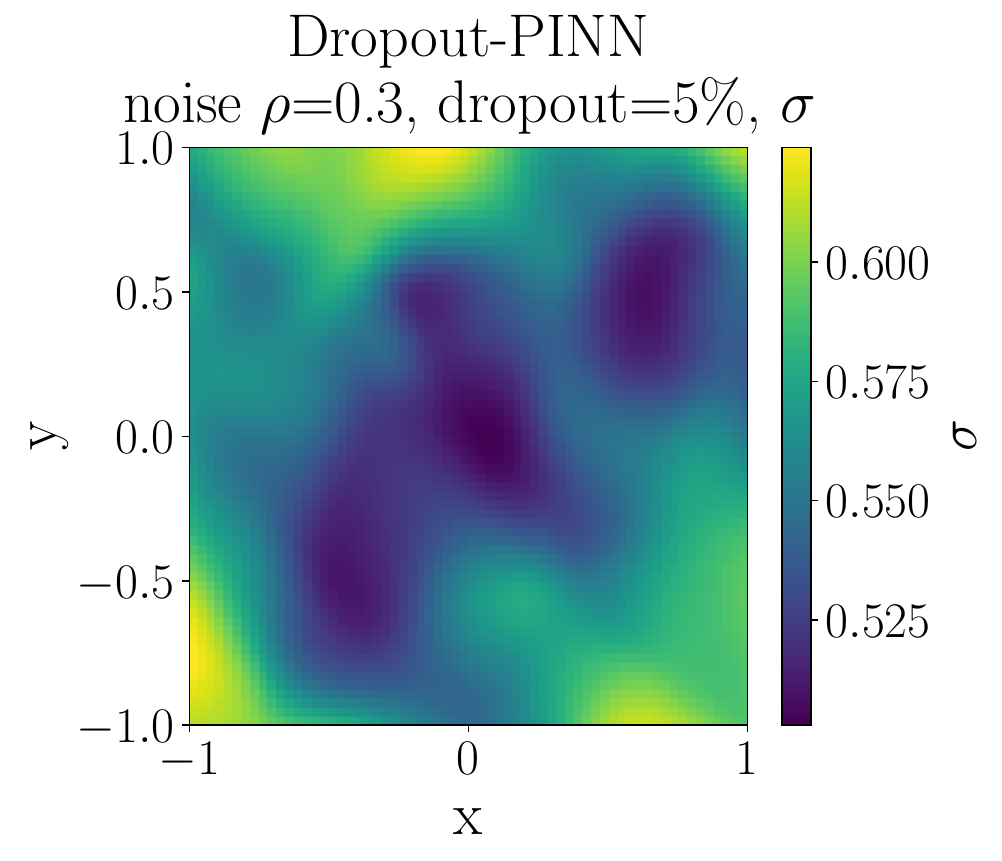}
    \end{subfigure}\hfill
    \begin{subfigure}[b]{0.32\textwidth}\centering
        \incfig[width=\textwidth]{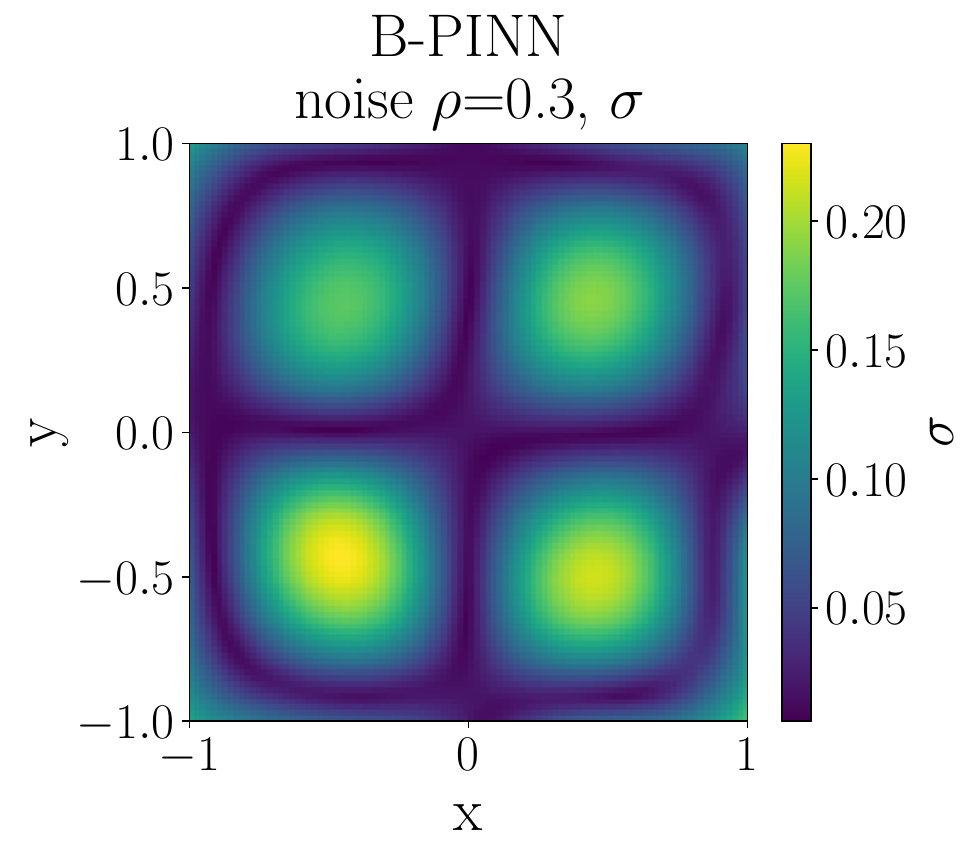}
    \end{subfigure}
    \par\vspace{0.4em}
    \begin{subfigure}[b]{0.32\textwidth}\centering
        \incfig[width=\textwidth]{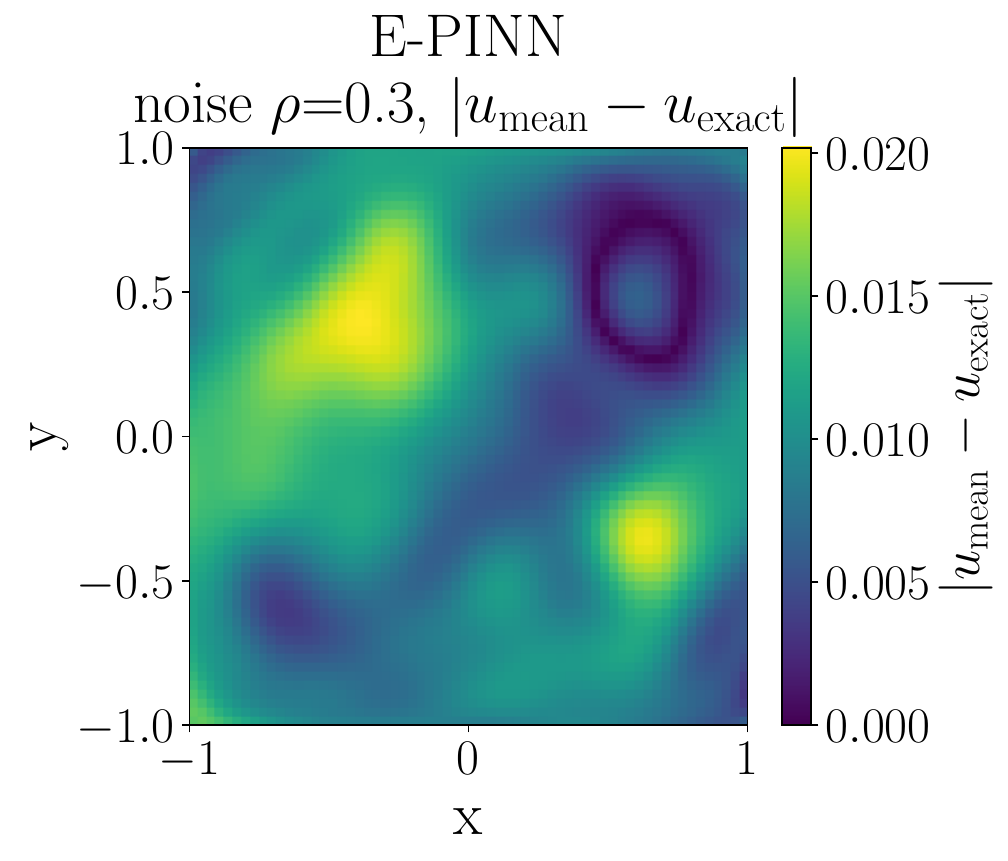}
    \end{subfigure}\hfill
    \begin{subfigure}[b]{0.32\textwidth}\centering
        \incfig[width=\textwidth]{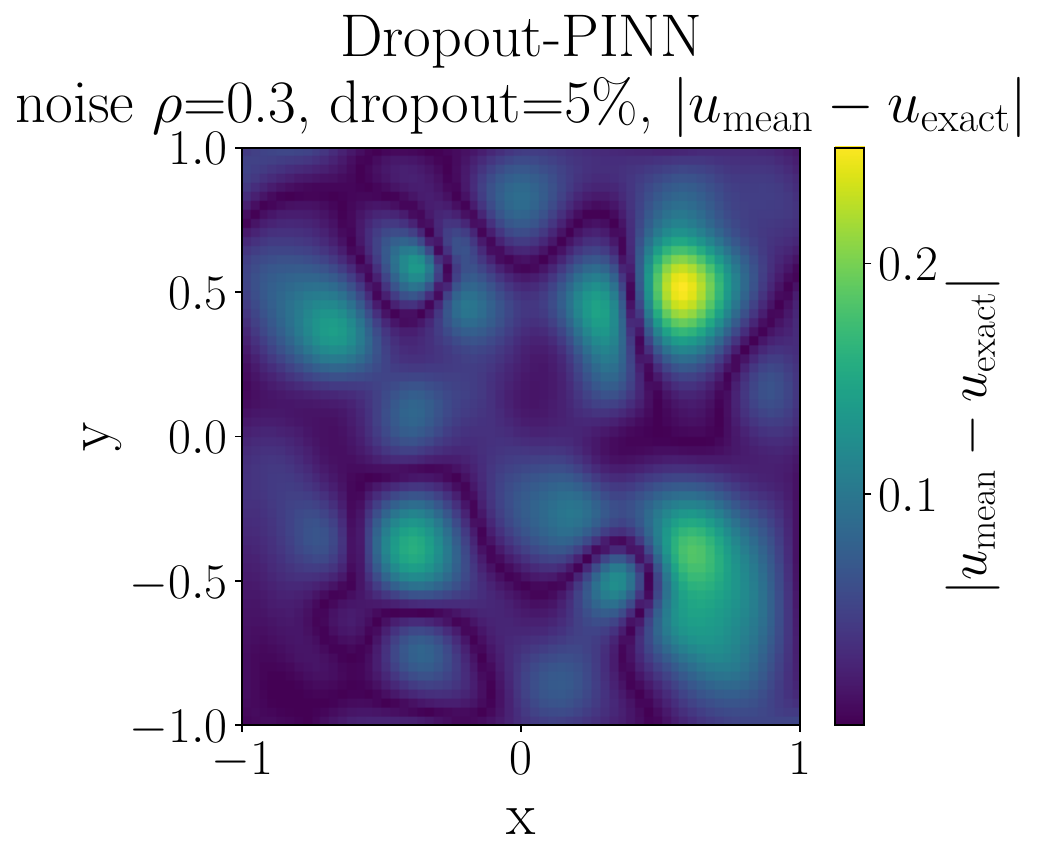}
    \end{subfigure}\hfill
    \begin{subfigure}[b]{0.32\textwidth}\centering
        \incfig[width=\textwidth]{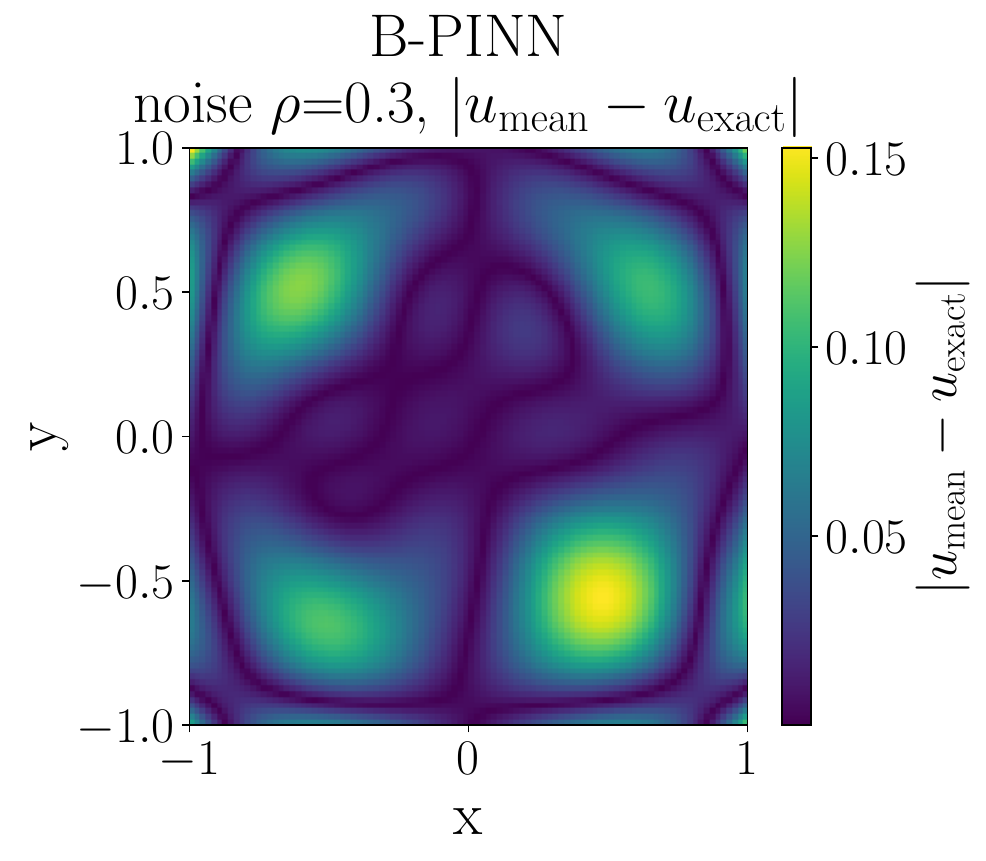}
    \end{subfigure}
    \caption{2D nonlinear Poisson ($\rho=0.30$): top, mean $u$; middle, epistemic standard deviation; bottom, absolute error. Columns: E-PINN (left), Dropout-PINN (center), B-PINN (right).}
    \label{fig:2D_nlP_grid_0.30}
\end{figure}

Figure~\ref{fig:2D_nlP_grid_0.00} summarizes the physics-only case (mean, standard deviation, and error); Figure~\ref{fig:2D_nlP_grid_0.30} shows the corresponding comparison at $\rho=0.30$. Additional plots for $\rho=0.10$ are provided in Appendix~\ref{app:2d_nonlin_appendix}.

\begin{table}[H]
\centering
\caption{2D nonlinear Poisson: metrics vs.\ noise scale $\rho$ (standard deviation $\sigma=\rho\,\lVert u\rVert_{\infty}$). RMSE is computed with the predictive mean $\mu_u$. Time (s) is total wall-clock time; Time (epinet) (s) reports the E-PINN epinet-only training time (only populated for E-PINN rows). For $\rho=0$ (physics-only), B-PINN is not evaluated. (Sharpness/time/RMSE: lower is better; coverage: higher is better.)}
\label{tab:2d_nonlin_poisson_all}
\sisetup{detect-weight=true,detect-family=true,mode=text}
\begin{tabular*}{\textwidth}{@{\extracolsep{\fill}}
S[table-format=1.2]
l
S[table-format=1.2]
S[table-format=1.2]
S[table-format=1.4]
S[table-format=4.2]
 c
}
\toprule
\multicolumn{1}{c}{$\rho$} & \multicolumn{1}{c}{Method} &
\multicolumn{1}{c}{Sharpness $\mu^w_{2\sigma}$} &
\multicolumn{1}{c}{Coverage (95\%)} &
\multicolumn{1}{c}{RMSE} &
\multicolumn{1}{c}{Time (s)} & \multicolumn{1}{c}{Time (epinet) (s)} \\
\midrule
0.00 & E-PINN        & \bfseries 0.20 & \bfseries 1.00 & \bfseries 0.0081 & 1698.99 & 478.19 \\
0.00 & Dropout 5\%   & 2.36             & \bfseries 1.00 & 0.0321              & \bfseries 1180.79 & -- \\
0.00 & Dropout 10\%  & 2.91             & \bfseries 1.00 & 0.0589              & 1202.74 & -- \\
\midrule
0.10 & E-PINN        &  0.33 & \bfseries 1.00 & \bfseries 0.0082 & 1699.26 & 493.76 \\
0.10 & B-PINN        & \bfseries 0.26             & 0.99              & 0.0428              & 2623.84 & -- \\
0.10 & Dropout 5\%   & 2.08             & \bfseries 1.00 & 0.0425              & \bfseries 1298.83 & -- \\
0.10 & Dropout 10\%  & 2.62             & \bfseries 1.00 & 0.0697              & 1333.20 & -- \\
\midrule
0.30 & E-PINN        & 0.32 & \bfseries 1.00 & \bfseries 0.0104 & 1764.36 & 510.09 \\
0.30 & B-PINN        & \bfseries 0.31             & 0.99              & 0.0493              & 2555.59 & -- \\
0.30 & Dropout 5\%   & 2.22             & \bfseries 1.00 & 0.0637              & \bfseries 1212.97 & -- \\
0.30 & Dropout 10\%  & 3.19             & \bfseries 1.00 & 0.0767              & 2480.38 & -- \\
\bottomrule
\end{tabular*}
\end{table}

Across these regimes (Table~\ref{tab:2d_nonlin_poisson_all}), E-PINN attains calibrated bands with narrow sharpness in physics-only ($\mu^w_{2\sigma}\approx0.20$) and remains sharp with interior $u$ data ($\approx0.33$ at $\rho=0.10$, $\approx0.32$ at $\rho=0.30$), yielding the smallest RMSEs ($\approx8\times10^{-3}$ to $\approx1\times10^{-2}$). Dropout-PINN produces substantially wider intervals ($\approx2.1$ to $\approx3.2$) with larger errors across noise levels. B-PINN coverage is $0.99$ with sharpness comparable to E-PINN but higher RMSE ($\approx4.3\times10^{-2}$ to $\approx4.9\times10^{-2}$) and higher total runtime ($\approx2.6\times10^{3}$ s on CPU). The last column reports the E-PINN epinet-only training time ($\approx4.8\times10^{2}$ to $\approx5.1\times10^{2}$ s), the incremental cost when a base PINN is already trained.

\subsection{Inverse problems}
\subsubsection{1D heat equation}
We demonstrate inverse modeling by estimating the thermal diffusivity $\kappa$ in the 1D heat equation with initial and boundary conditions:
\begin{subequations}\label{eq:heat_system}
\begin{align}
    \frac{\partial u}{\partial t} &= \kappa\, \frac{\partial^2 u}{\partial x^2}, && x\in[0,1],\; t\in[0,1], \label{eq:heat_system_pde}\\
    u(x,0) &= \sin(\pi x), \label{eq:heat_system_ic}\\
    u(0,t) &= 0,\quad u(1,t) = 0. \label{eq:heat_system_bc}
\end{align}
\end{subequations}
Training uses noise-free PDE, boundary, and initial-condition targets together with noisy interior measurements of $u(x,t)$ at space-time sensors, where the noise level is a fraction of $\lVert u \rVert_{\infty}$ (fractions $0.10$, $0.30$). The data are generated with true thermal diffusivity $\kappa=0.1$. We predict a spatiotemporal field $\hat\kappa(x,t)$ jointly with $u$ and penalize its variance so that the learned $\hat\kappa$ is effectively constant; we summarize $\kappa$ by spatiotemporal averaging.
\begin{align}
\mathcal{L}_{\kappa} &= \operatorname{Var}_{(x,t)\in\mathcal{C}}[\hat\kappa(x,t)],\\
\mathcal{L}_{\text{pde}} &= \frac{1}{N_f}\sum_{i=1}^{N_f} \Big|\frac{\partial u_\theta}{\partial t}(x_i^f,t_i^f) - \hat\kappa(x_i^f,t_i^f)\, \frac{\partial^2 u_\theta}{\partial x^2}(x_i^f,t_i^f)\Big|^2.
\end{align}
The heat inverse objective augments the generic loss with $\mathcal{L}_{\kappa}$; $\hat\kappa$ is shared across methods. Under this design, E-PINN yields a distribution over $\kappa$ via Monte Carlo sampling over the epistemic index $z$ (with epinet prior scaling $\alpha=0.05$); Dropout-PINN yields a $\kappa$ distribution via Monte Carlo dropout; and B-PINN yields a $\kappa$ posterior via Hamiltonian Monte Carlo sampling of the network weights.

\begin{figure}[H]
    \centering
    \begin{subfigure}[b]{0.4\textwidth}
        \centering
        \incfig[width=\textwidth]{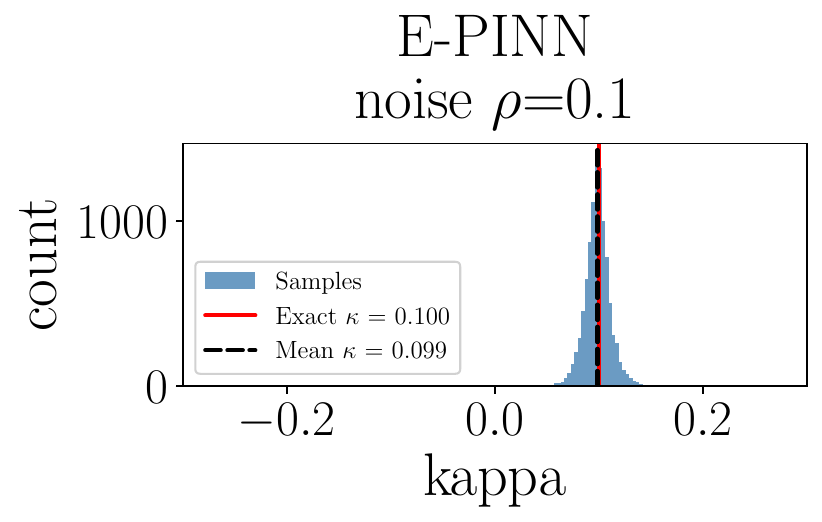}
        \caption{E-PINN}
    \end{subfigure}\hfill
    \begin{subfigure}[b]{0.4\textwidth}
        \centering
        \incfig[width=\textwidth]{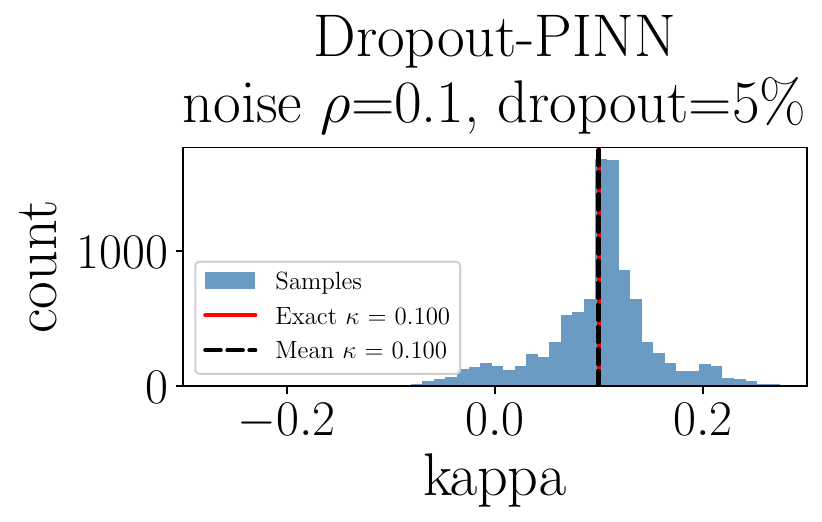}
        \caption{Dropout-PINN (5\%)}
    \end{subfigure}
    \par\vspace{0.4em}
    \begin{subfigure}[b]{0.4\textwidth}
        \centering
        \incfig[width=\textwidth]{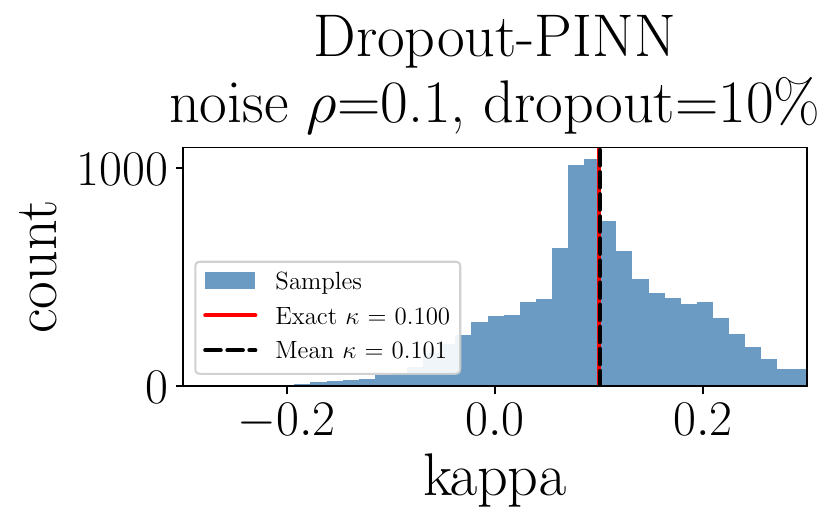}
        \caption{Dropout-PINN (10\%)}
    \end{subfigure}\hfill
    \begin{subfigure}[b]{0.4\textwidth}
        \centering
        \incfig[width=\textwidth]{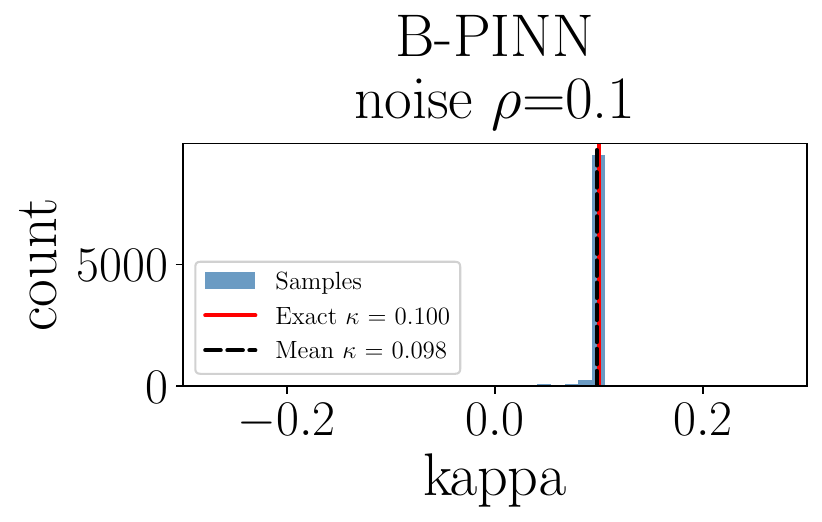}
        \caption{B-PINN}
    \end{subfigure}
    \caption{1D heat equation (inverse): histograms of $\kappa$ for E-PINN, Dropout-PINN (5\% and 10\%), and B-PINN.}
    \label{fig:heat_kappa_hist}
\end{figure}

We summarize $\kappa$ estimates and wall-clock timings across noise levels in Table~\ref{tab:heat_kappa_time}. For visual reference, Figure~\ref{fig:heat_kappa_hist} shows representative $\kappa$ histograms for the three methods (data-augmented regime). Because this is a temporal problem, the table reports moments of the inferred scalar parameter $\kappa$ from Monte Carlo/HMC samples, while the line plots elsewhere show instantaneous $u(x,t)$ means and $\pm 2\sigma$ bands at fixed times. For E-PINN, the "epinet time" column reports the incremental cost when a base PINN is already trained. At $\rho=0.30$, the B-PINN posterior mean underestimates $\kappa$ (about 0.078 versus the true 0.1), while E-PINN and Dropout-PINN remain centered near 0.1 with larger dispersion for dropout. In the B-PINN implementation used here, the likelihood includes boundary/initial-condition and PDE-residual channels with strong but finite precisions together with interior $u$-data; the underestimation reflects the information balance under these fixed precisions.
\begin{table}[h]
\centering
\caption{Inverse heat: $\kappa$ (mean $\pm$ std) and timing across Gaussian noise scales $\rho$ (standard deviation $\sigma = \rho\,\lVert u\rVert_{\infty}$). Time (epinet) (s) is the E-PINN epinet-only training time (only populated for E-PINN rows).}
\label{tab:heat_kappa_time}
\sisetup{detect-weight=true,detect-family=true,mode=text}
\begin{tabular*}{\textwidth}{@{\extracolsep{\fill}} S[table-format=1.2] l c S[table-format=4.2] c }
\toprule
\multicolumn{1}{c}{$\rho$} & \multicolumn{1}{c}{Method} & \multicolumn{1}{c}{$\kappa$ (mean $\pm$ std)} & \multicolumn{1}{c}{Time (s)} & \multicolumn{1}{c}{Time (epinet) (s)} \\
\midrule\midrule
0.10 & E-PINN        & $0.099\,\pm\,0.012$ & 4258.50 & 1068.63 \\
0.10 & B-PINN        & $0.098\,\pm\,0.009$ & 5553.45 & -- \\
0.10 & Dropout 5\%  & $0.100\,\pm\,0.058$ & 1075.56 & -- \\
0.10 & Dropout 10\% & $0.101\,\pm\,0.093$ & 1073.78 & -- \\
\midrule
0.30 & E-PINN        & $0.101\,\pm\,0.015$ & 4238.53 & 1026.48 \\
0.30 & B-PINN        & $0.078\,\pm\,0.019$ & 2583.11 & -- \\
0.30 & Dropout 5\%  & $0.101\,\pm\,0.056$ & 1219.62 & -- \\
0.30 & Dropout 10\% & $0.094\,\pm\,0.119$ & 1570.04 & -- \\
\bottomrule
\end{tabular*}
\end{table}

\begin{figure}[H]
    \centering
    \begin{subfigure}[b]{0.32\textwidth}
        \centering
        \incfig[width=\textwidth]{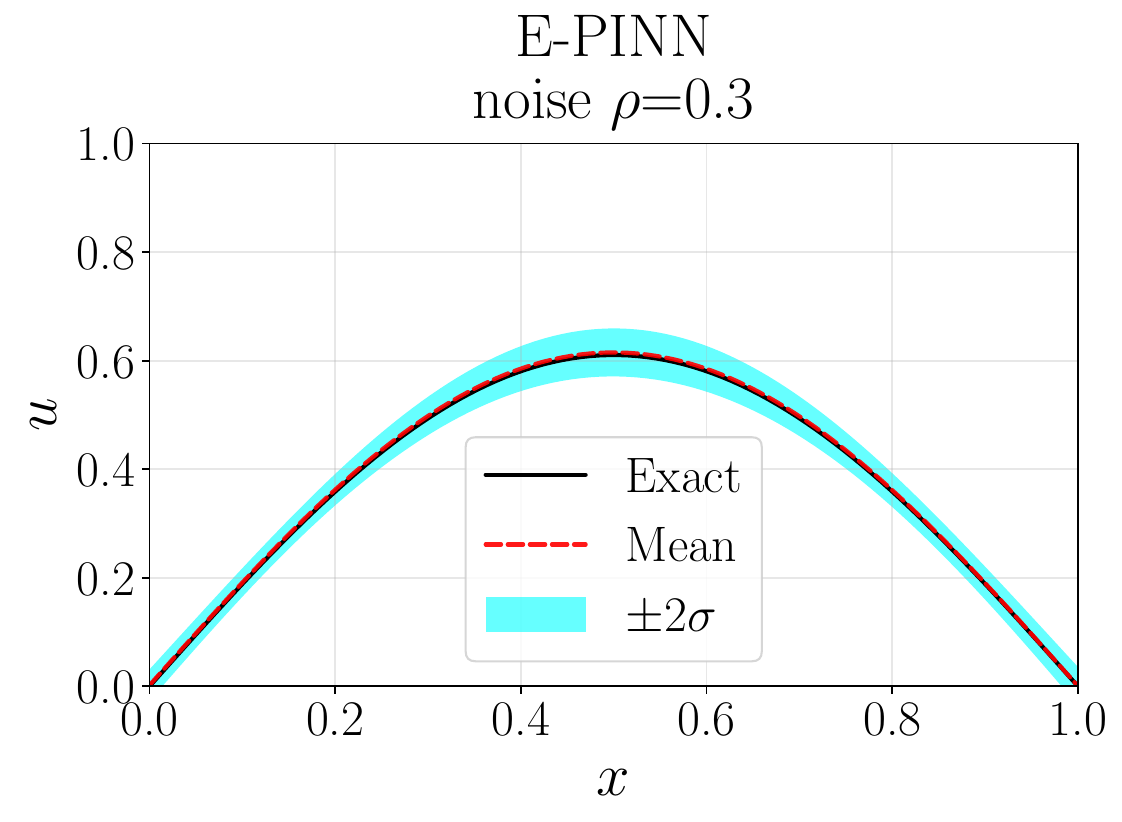}
        \caption{E-PINN}
    \end{subfigure}\hfill
    \begin{subfigure}[b]{0.32\textwidth}
        \centering
        \incfig[width=\textwidth]{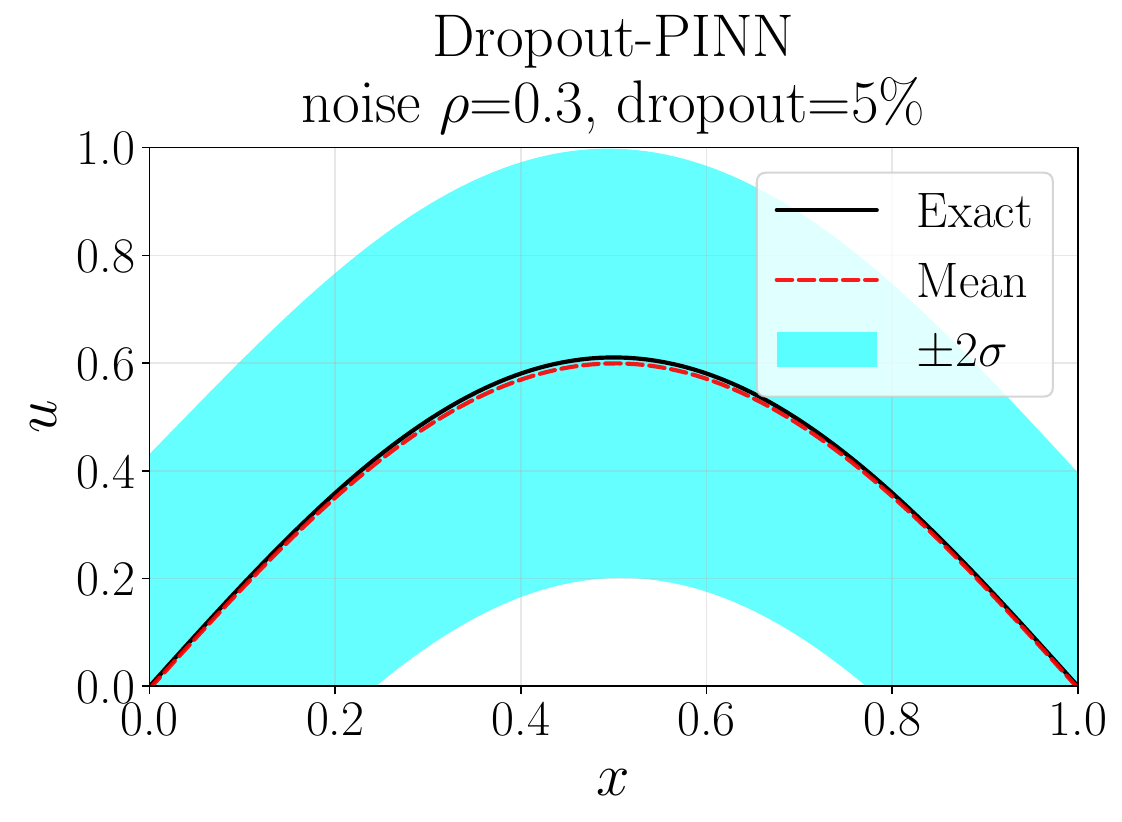}
        \caption{Dropout-PINN (5\%)}
    \end{subfigure}\hfill
    \begin{subfigure}[b]{0.32\textwidth}
        \centering
        \incfig[width=\textwidth]{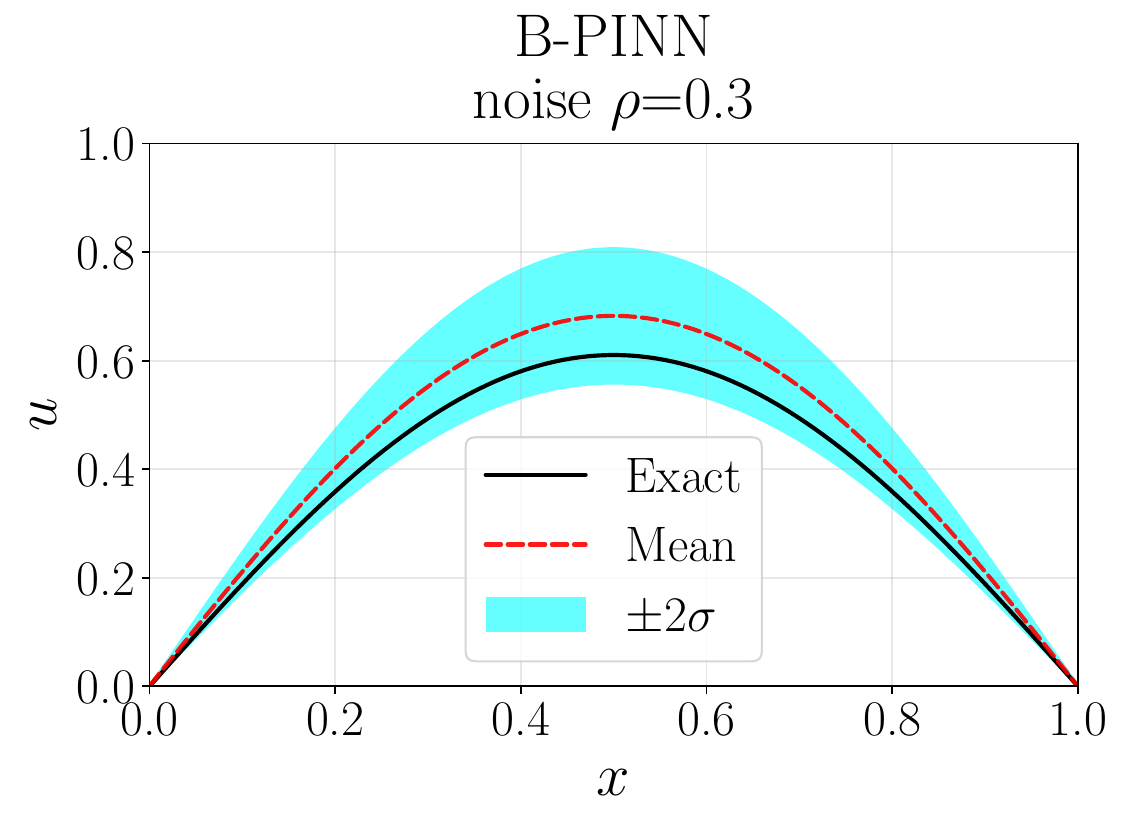}
        \caption{B-PINN}
    \end{subfigure}
    \caption{1D heat equation (inverse; $\rho=0.30$, $t=0.50$): mean and $\pm 2\sigma$ epistemic bands for $u(x,t)$ (instantaneous slice). Methods: (a) E-PINN, (b) Dropout-PINN (5\%), (c) B-PINN.}
    \label{fig:heat_u_t050}
\end{figure}

\begin{figure}[H]
    \centering
    \begin{subfigure}[b]{0.32\textwidth}
        \centering
        \incfig[width=\textwidth]{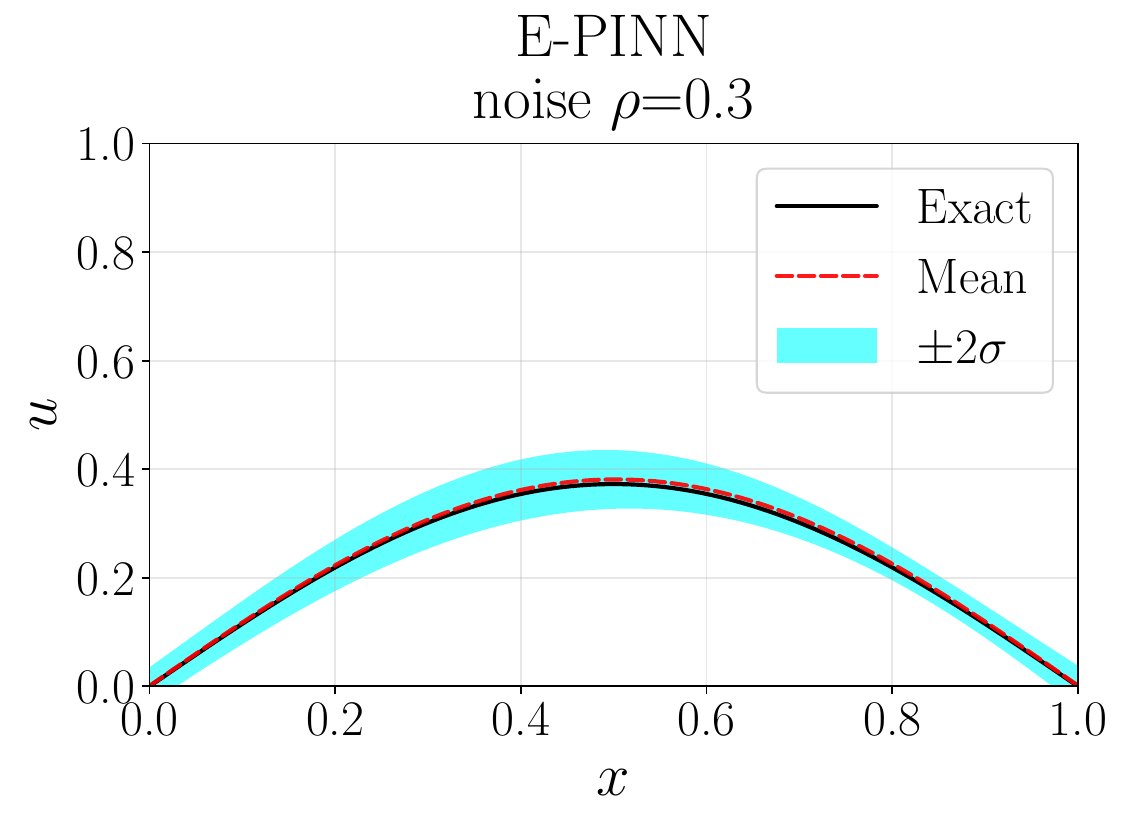}
        \caption{E-PINN}
    \end{subfigure}\hfill
    \begin{subfigure}[b]{0.32\textwidth}
        \centering
        \incfig[width=\textwidth]{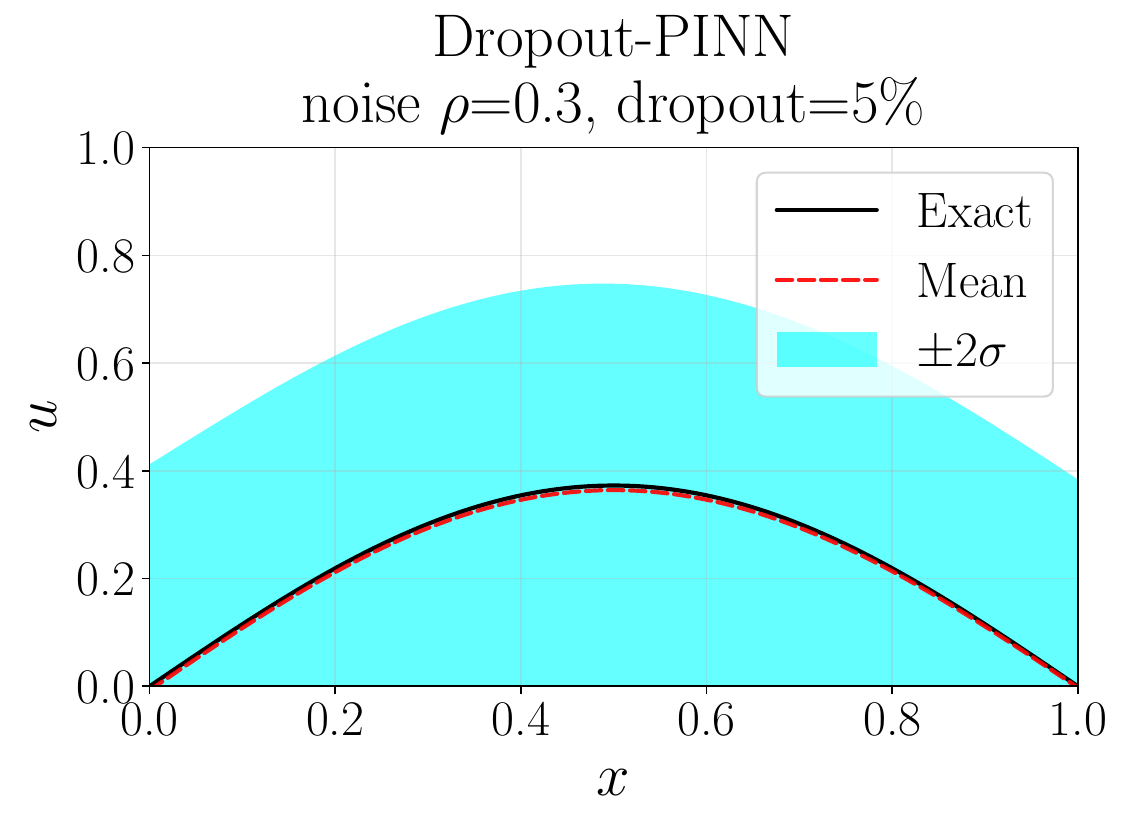}
        \caption{Dropout-PINN (5\%)}
    \end{subfigure}\hfill
    \begin{subfigure}[b]{0.32\textwidth}
        \centering
        \incfig[width=\textwidth]{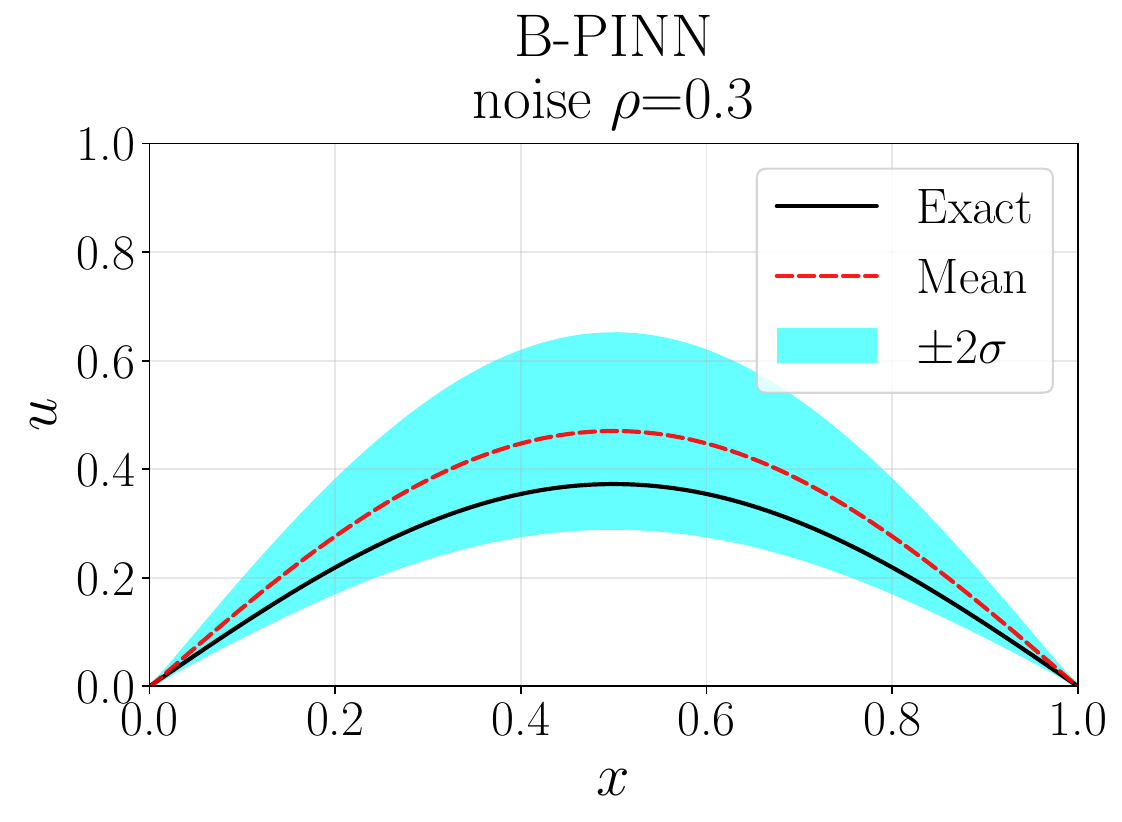}
        \caption{B-PINN}
    \end{subfigure}
    \caption{1D heat equation (inverse; $\rho=0.30$, $t=1.00$): mean and $\pm 2\sigma$ epistemic bands for $u(x,t)$ (instantaneous slice). Methods: (a) E-PINN, (b) Dropout-PINN (5\%), (c) B-PINN.}
    \label{fig:heat_u_t100}
\end{figure}

\subsection{Ablation study}\label{sec:ablation}

This section studies how E-PINN hyperparameters affect mean predictions and epistemic uncertainty. All ablations use the viscous Burgers equation on $x\in[-1,1]$, $t\in[0,1]$,

\begin{equation}
    \frac{\partial u}{\partial t} + u\,\frac{\partial u}{\partial x} - \nu\,\frac{\partial^2 u}{\partial x^2} = f(x,t),
\end{equation}

with viscosity $\nu = 0.01/\pi$. We adopt the exact solution

\begin{equation}
    u(x,t) = \sin(kx + \omega t),
\end{equation}

with $k=\pi$ and $\omega=2$, and set $f(x,t)$ by substituting this $u$ into the PDE. We vary four factors: the number of collocation points $N_{\text{colloc}}$, the number of epinet parameters (hidden units), the epinet training epochs, and the base network training epochs. In all ablations the base network is pre-trained and then held fixed while the epinet is trained. Unless stated otherwise, the architecture is fixed to three hidden layers of 32 units for both the base network and the epinet $((32,32,32))$, the epistemic index has dimension $d_z=8$, the prior contribution is scaled by $\alpha=0.05$, the base network is trained for $10^5$ epochs, and the epinet is trained for $10^4$ epochs. The only quantity that changes within each ablation is the variable of interest (e.g., $N_{\text{colloc}}$, $H$, or the number of epochs).

\subsubsection{Effect of number of collocation points}

\begin{figure}[H]
    \centering
    \incfig[width=0.45\textwidth]{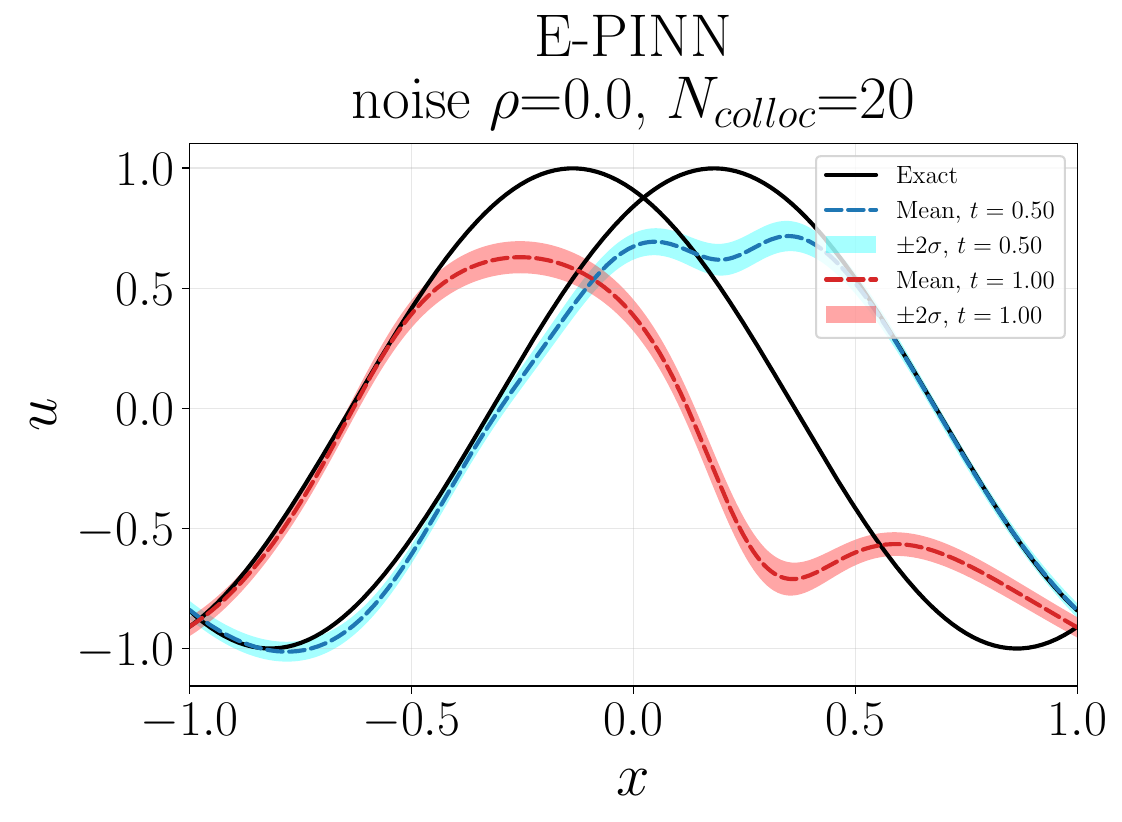} 
    \incfig[width=0.45\textwidth]{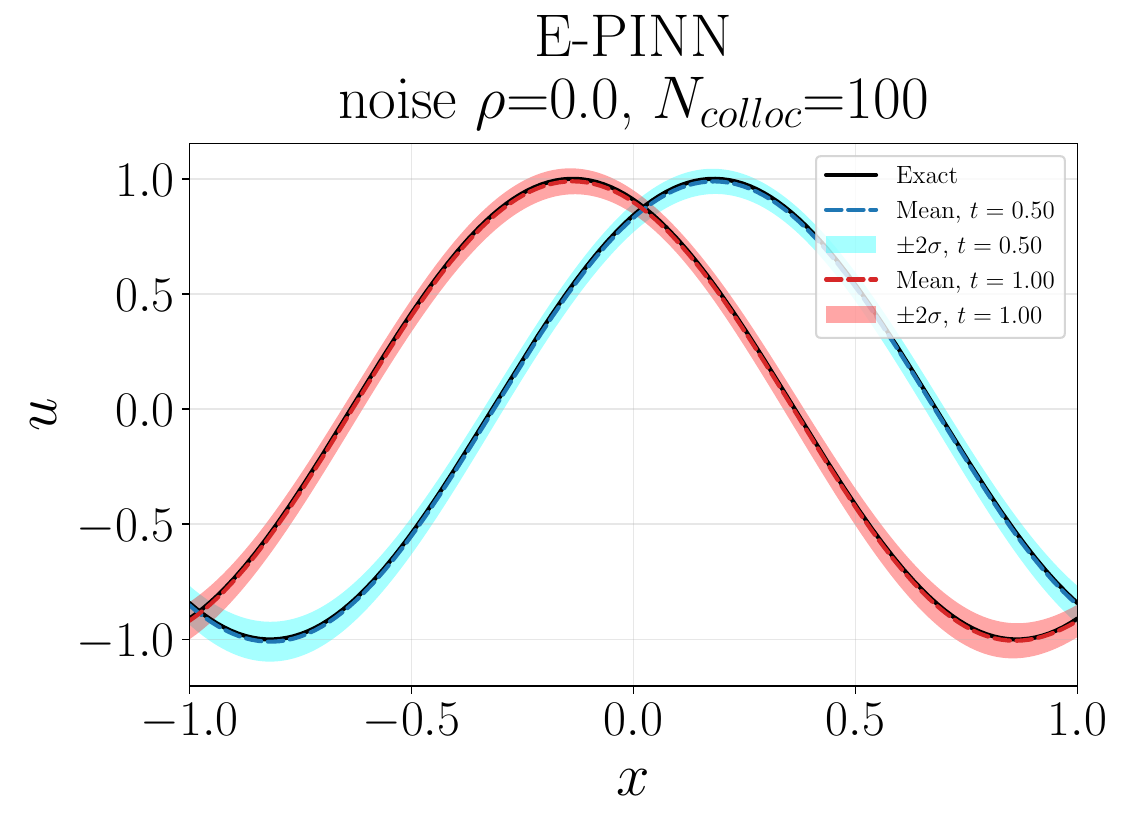} 
    \incfig[width=0.45\textwidth]{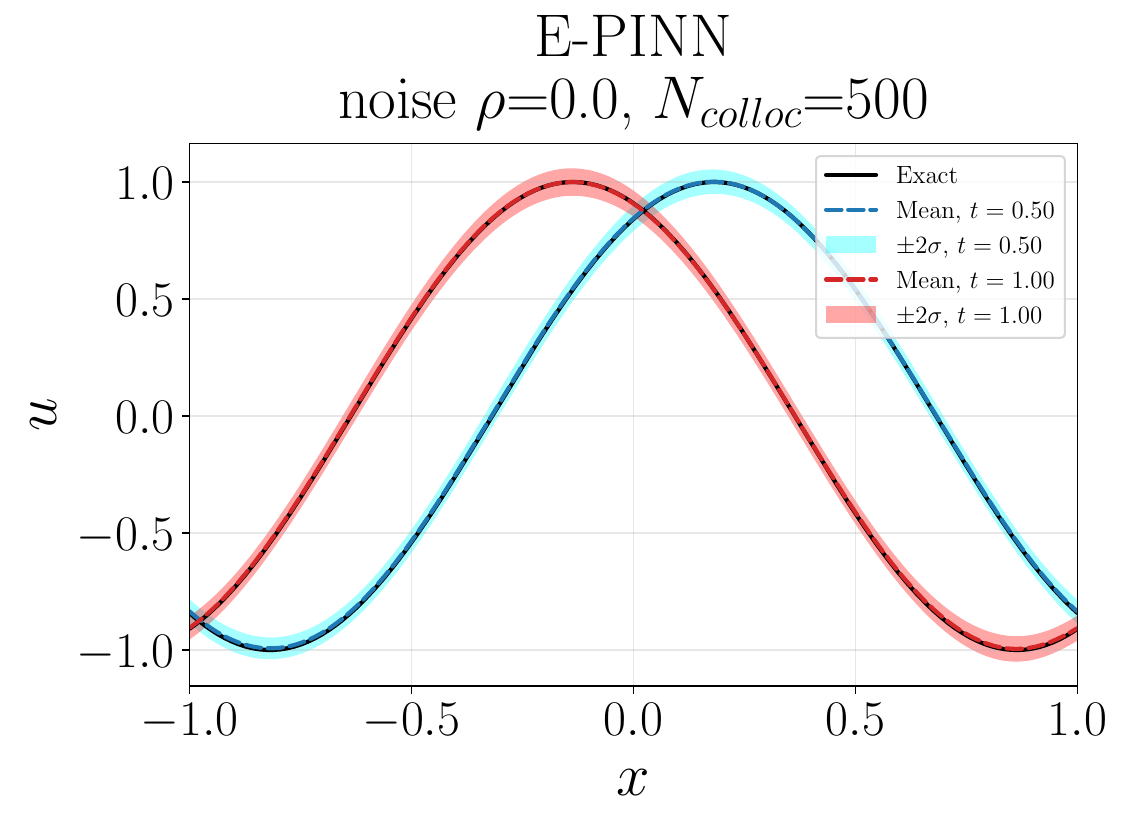} 
    \incfig[width=0.45\textwidth]{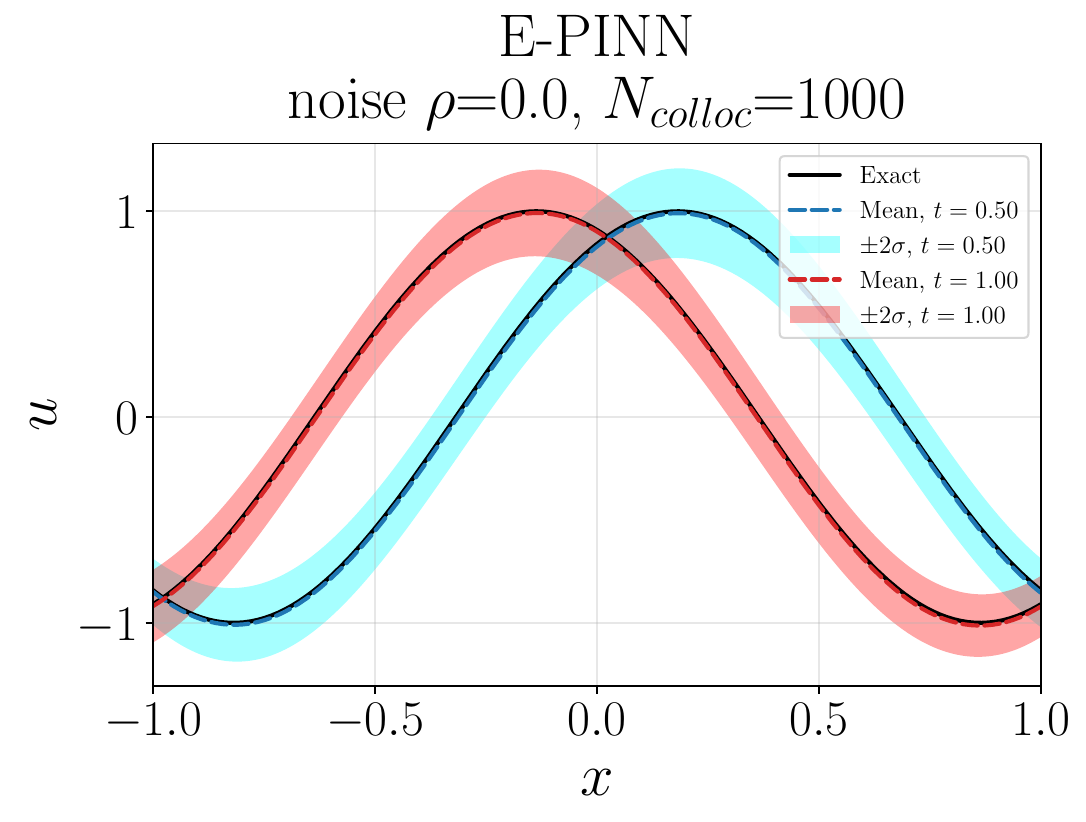} 

    \caption{Effect of collocation points. Shown for $N_{\text{colloc}}\in\{20,100,500,1000\}$. Each panel shows the predictive mean and $\pm2\sigma$ epistemic bands for $u$; architecture and training schedule are held fixed across runs.}
    \label{fig: Ablation_numSamps}
\end{figure}

This section studies the effect of the number of collocation points ($N_{\text{colloc}}$) used for training the base network and the epinet. All runs are physics only, and architectures and training schedules are held fixed across settings. Figure~\ref{fig: Ablation_numSamps} shows the four cases used in the study. Table~\ref{tab:ablation_burgers_ncolloc} reports sharpness (mean width of the $\pm 2\sigma$ band), empirical $95\%$ coverage, RMSE, and wall-clock time.

\begin{table}[H]
\centering
\caption{Burgers ablation (collocation): E-PINN metrics versus $N_{\text{colloc}}$ in physics only. RMSE is computed with the predictive mean $\mu_u$. Time (s) is total wall clock time; Time (epinet) (s) is the epinet only training time. (Sharpness/time/RMSE: lower is better; coverage: higher is better.)}
\label{tab:ablation_burgers_ncolloc}
\sisetup{detect-weight=true,detect-family=true,mode=text}
\begin{tabular*}{\textwidth}{@{\extracolsep{\fill}}
S[table-format=4.0]
S[table-format=1.2]
S[table-format=1.2]
S[table-format=1.4]
S[table-format=4.2]
S[table-format=4.2]
}
\toprule
\multicolumn{1}{c}{$N_{\text{colloc}}$} &
\multicolumn{1}{c}{Sharpness $\mu^w_{2\sigma}$} &
\multicolumn{1}{c}{Coverage (95\%)} &
\multicolumn{1}{c}{RMSE} &
\multicolumn{1}{c}{Time (s)} & \multicolumn{1}{c}{Time (epinet) (s)} \\
\midrule
20   & 0.09 & 0.35 & 0.2208 & 955.08 & 160.75 \\
100  & 0.14 & 1.00 & 0.0101 & 919.99 & 167.39 \\
500  & 0.10 & 1.00 & 0.0047 & 1440.84 & 252.71 \\
1000 & 0.39 & 1.00 & 0.0118 & 1240.20 & 281.78 \\
\bottomrule
\end{tabular*}
\end{table}

In this configuration (Table~\ref{tab:ablation_burgers_ncolloc}), $N_{\text{colloc}}=20$ underfits the physics, yielding low coverage and a large error. Moving to $N_{\text{colloc}}=100$ raises coverage to 1.00 and reduces RMSE by over an order of magnitude ($\approx 22\times$), with a slight decrease in total time. At $N_{\text{colloc}}=500$, sharpness narrows further and RMSE attains its minimum with higher total time. At $N_{\text{colloc}}=1000$, the mean error increases relative to 500 and the bands widen, while time decreases relative to 500 but remains in the same range, indicating a non-monotone dependence of sharpness and error on $N_{\text{colloc}}$ in this setup. With a fixed training budget, increasing collocation points raises residual constraint per epoch; for this PDE and optimizer schedule, $N_{\text{colloc}}\in[100,500]$ provides a stable calibration regime where the base features remain informative for the epinet and the epinet concentrates uncertainty where the solution is most variable. At $N_{\text{colloc}}=1000$, the base optimization is comparatively less complete under the same budget, and the epinet expresses this as wider bands while preserving nominal coverage.

\subsubsection{Effect of number of epinet network parameters}
This ablation varies only the epinet hidden width, $H$, with the number of epinet layers fixed at two. The base network architecture and all training schedules remain fixed across settings. We set $N_{\text{colloc}}=1000$; the base network is trained for $10^{5}$ epochs and the epinet for $10^{4}$ epochs.

\begin{figure}[]
    \centering
    \incfig[width=0.45\textwidth]{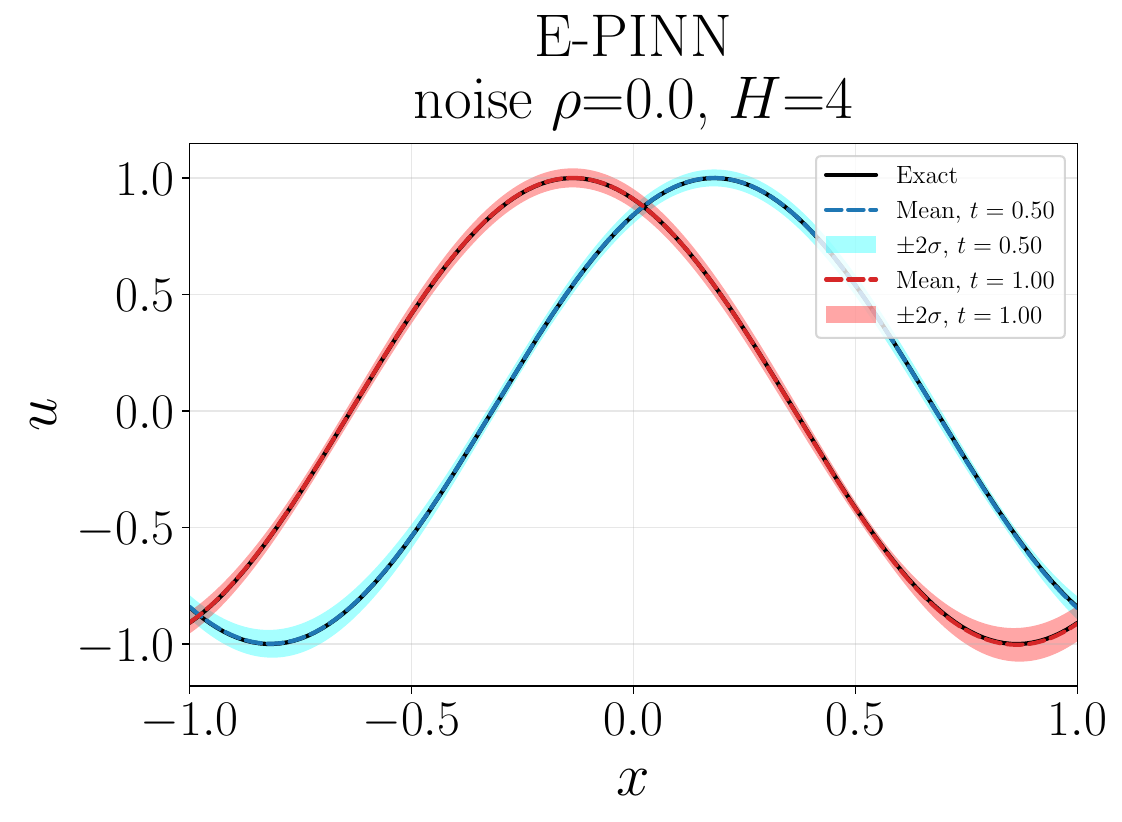}
    \incfig[width=0.45\textwidth]{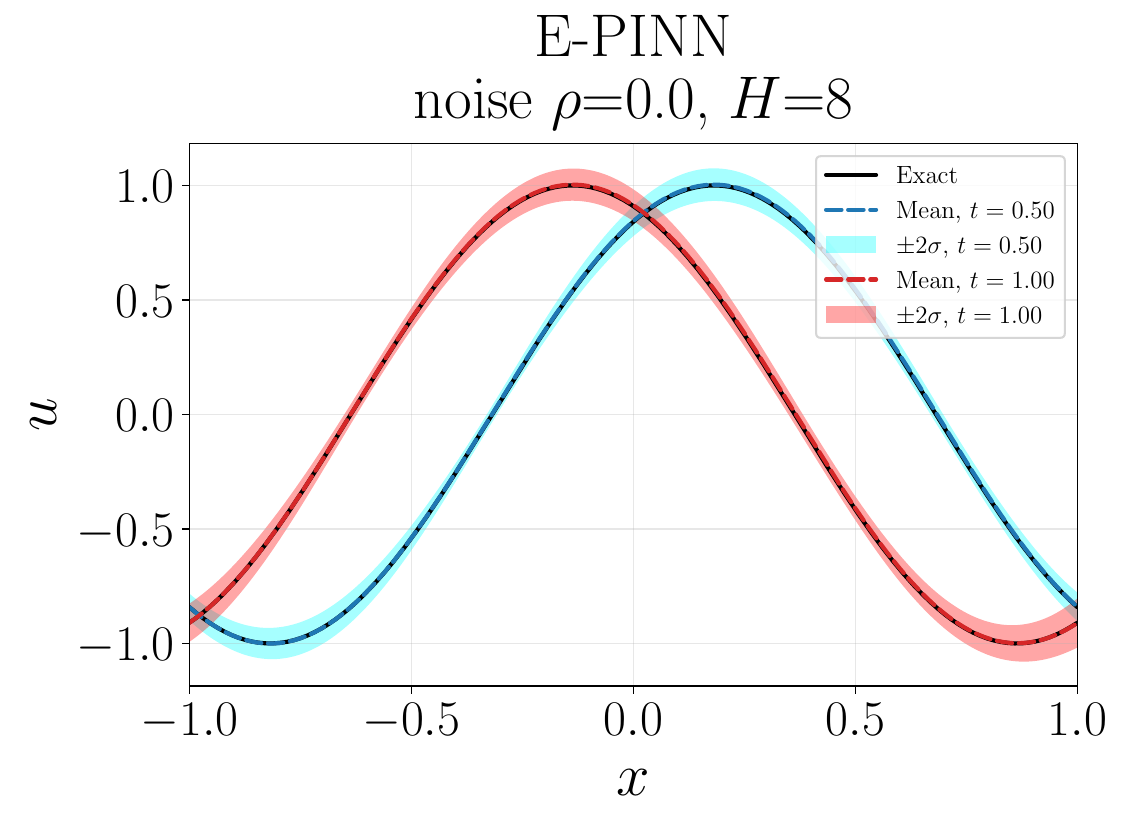}
    \incfig[width=0.45\textwidth]{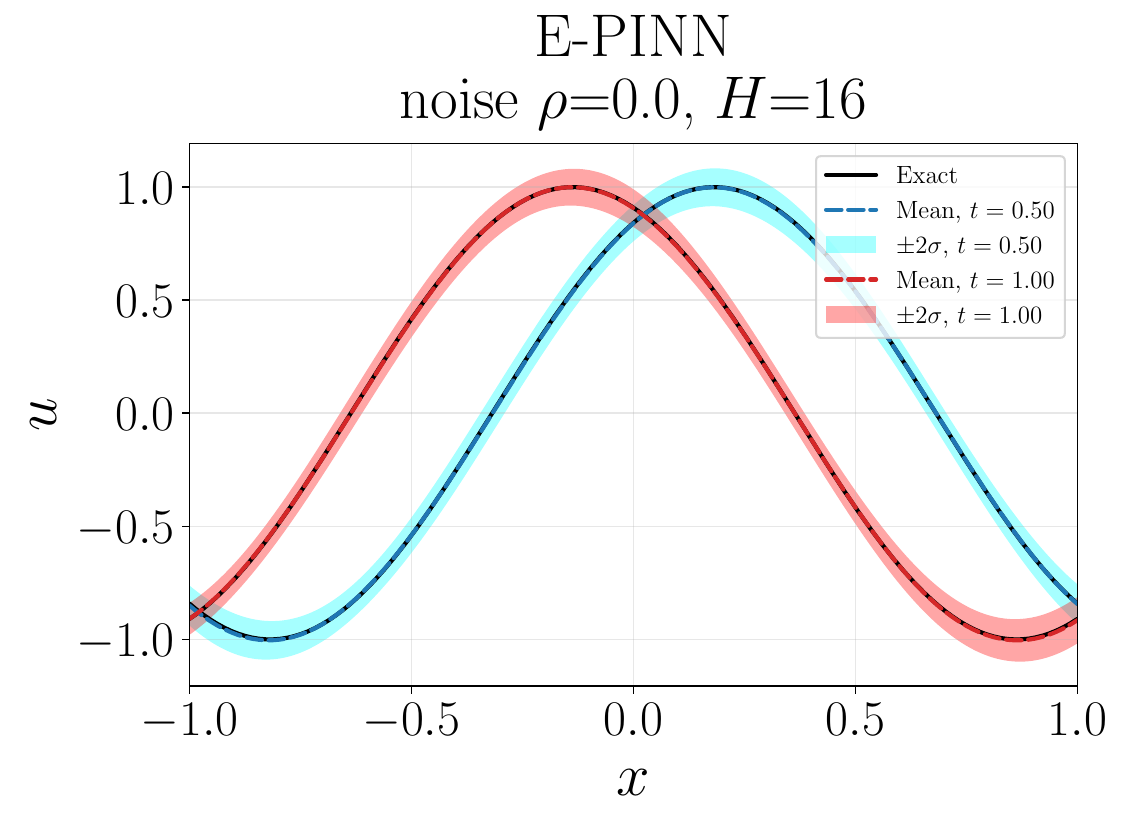}
    \incfig[width=0.45\textwidth]{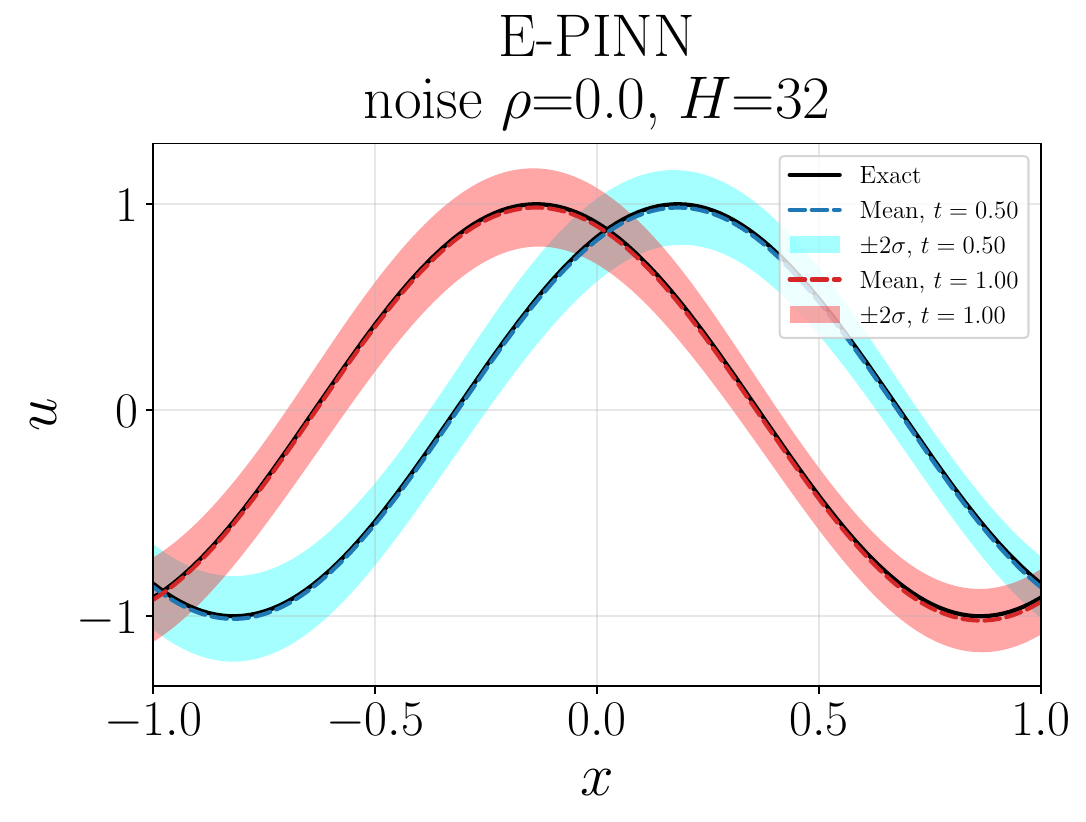}
    \incfig[width=0.45\textwidth]{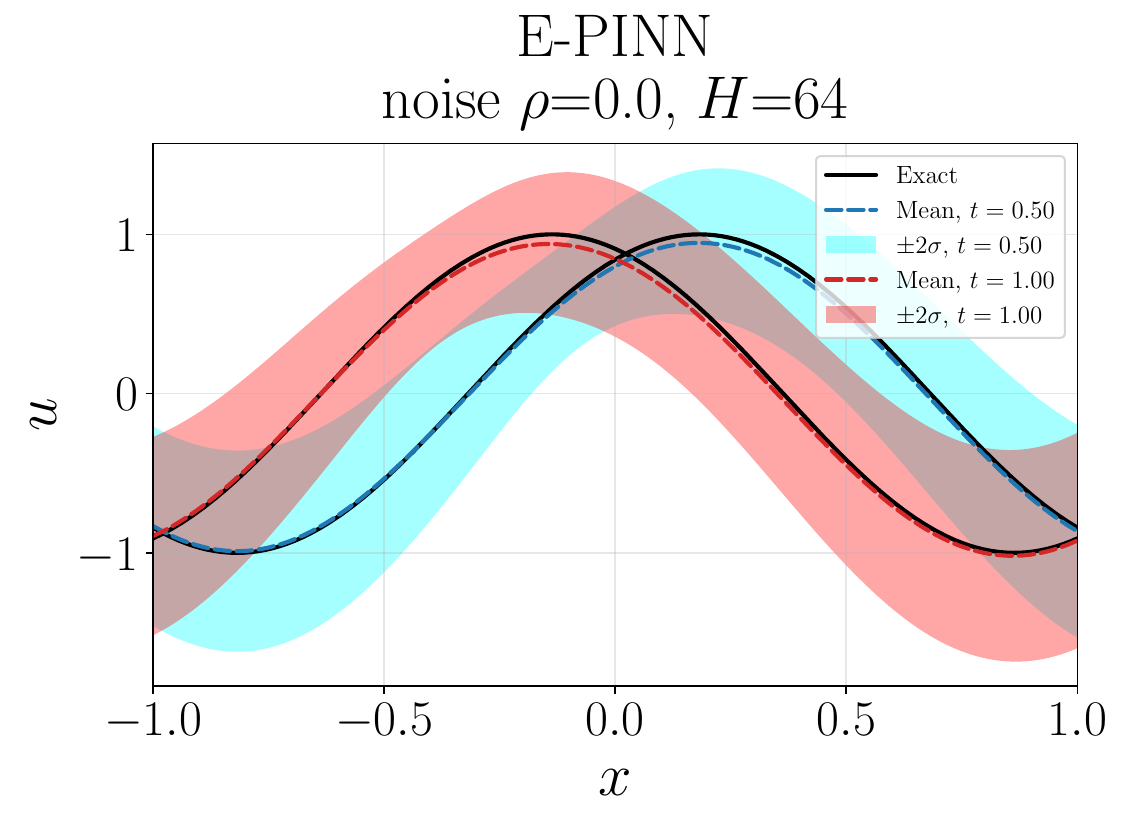}

    \caption{Effect of epinet width $H$. Shown for $H\in\{4,8,16,32,64\}$ with $N_{\text{colloc}}=1000$. Each panel shows the predictive mean and $\pm2\sigma$ epistemic bands for $u$; all other settings are fixed across runs.}
    \label{fig: Ablation_numParams}
\end{figure}

Figure~\ref{fig: Ablation_numParams} and Table~\ref{tab:Table_numParams} show that all settings achieve nominal coverage. For $H\in\{4,8,16\}$, bands are narrow ($\mu^w_{2\sigma}\approx0.09$ to $0.16$) and RMSE remains near $10^{-3}$. Increasing $H$ to $32$ and $64$ enlarges the bands and raises RMSE without improving coverage, with only moderate changes in time. In this setting the epinet acts as a low-variance corrector when its capacity is commensurate with the structure present in the base features; overly wide epinets increase the variance of the learned correction without improving fit to the noiseless target, which appears as wider bands and larger RMSE at similar coverage. For this problem, $H\in[8,16]$ provides a practical trade-off across sharpness, accuracy, and time.

\begin{table}[H]
\centering
\caption{Epinet width ablation: E-PINN metrics versus $H$. RMSE is computed with the predictive mean $\mu_u$. Time (s) is total wall clock time; Time (epinet) (s) is the epinet only training time. (Sharpness/time/RMSE: lower is better; coverage: higher is better.)}
\label{tab:Table_numParams}
\sisetup{detect-weight=true,detect-family=true,mode=text}
\begin{tabular*}{\textwidth}{@{\extracolsep{\fill}}
S[table-format=2.0]
S[table-format=1.2]
S[table-format=1.2]
S[table-format=1.4]
S[table-format=4.2]
S[table-format=4.2]
}
\toprule
\multicolumn{1}{c}{$H$} &
\multicolumn{1}{c}{Sharpness $\mu^w_{2\sigma}$} &
\multicolumn{1}{c}{Coverage (95\%)} &
\multicolumn{1}{c}{RMSE} &
\multicolumn{1}{c}{Time (s)} & \multicolumn{1}{c}{Time (epinet) (s)} \\
\midrule
4  & 0.09 & 1.00 & 0.0016 & 1172.99 & 237.50 \\
8  & 0.12 & 1.00 & 0.0034 & 1210.00 & 248.76 \\
16 & 0.16 & 1.00 & 0.0033 & 1186.97 & 246.22 \\
32 & 0.37 & 1.00 & 0.0146 & 1194.05 & 254.31 \\
64 & 1.09 & 1.00 & 0.0308 & 1320.37 & 299.26 \\
\bottomrule
\end{tabular*}
\end{table}

\subsubsection{Effect of the number of training epochs in the epinet}
\begin{figure}[H]
    \centering
    \incfig[width=0.45\textwidth]{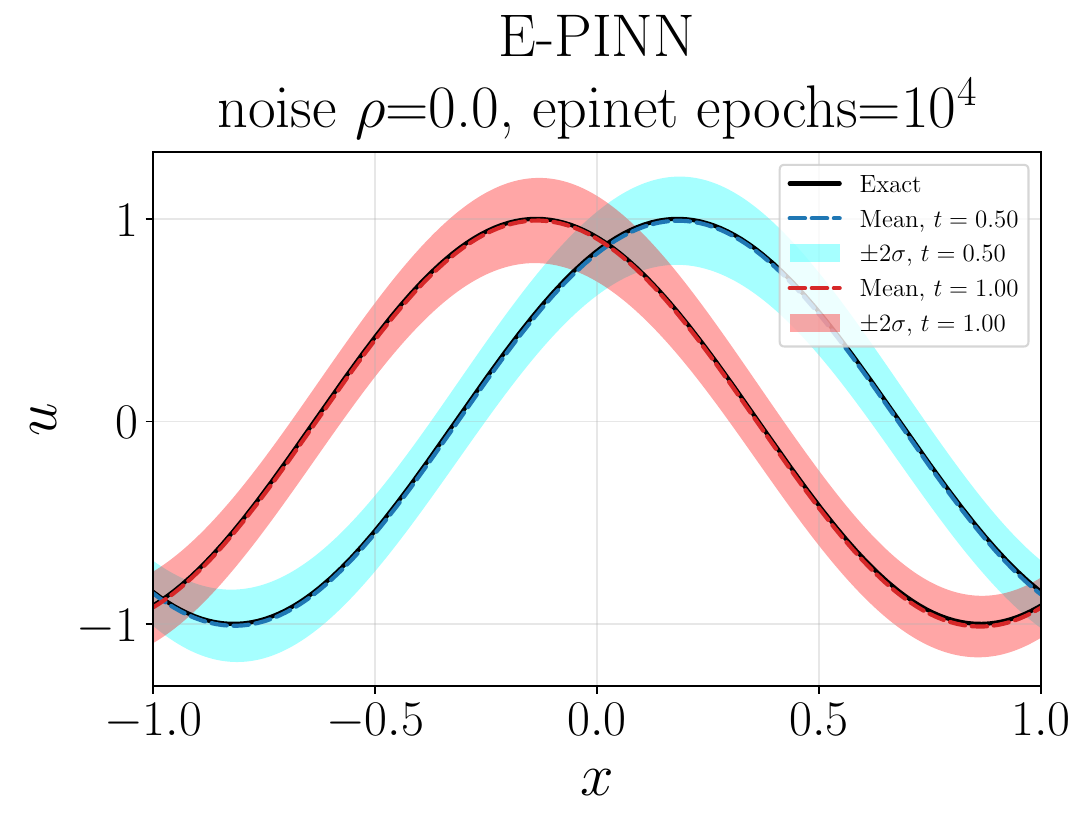}
    \incfig[width=0.45\textwidth]{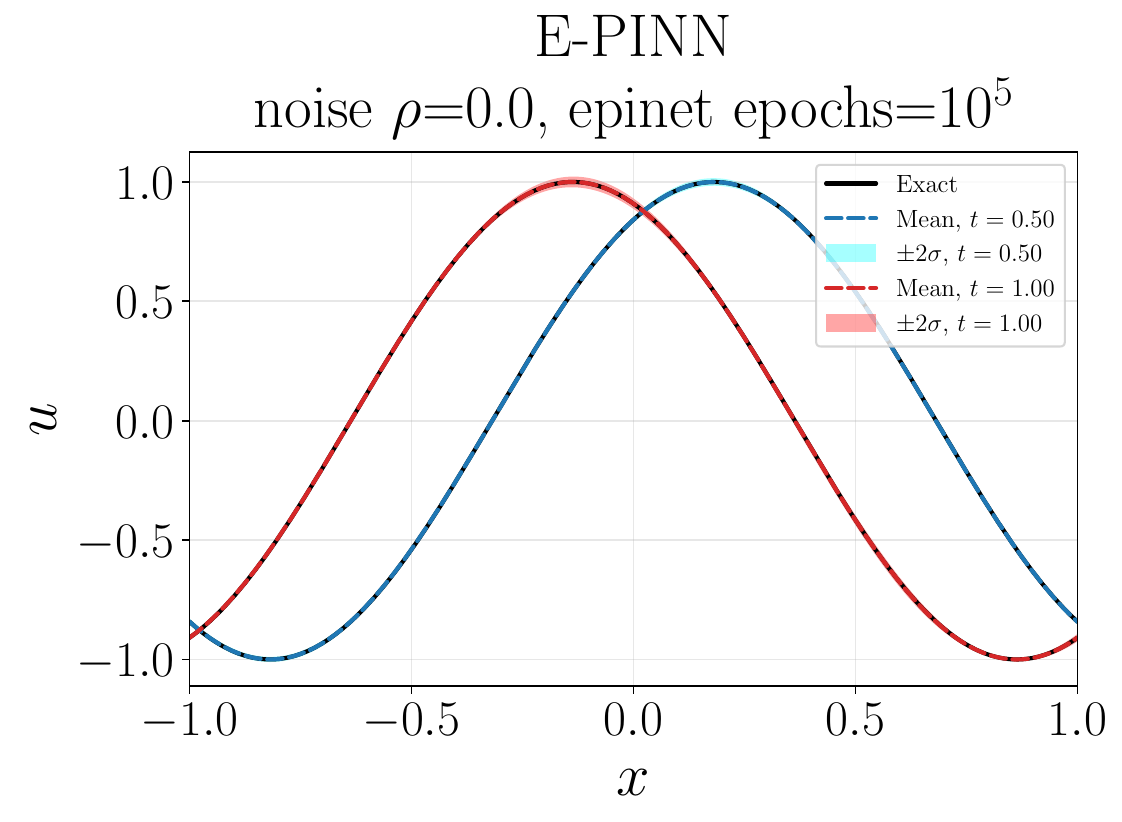}
    \incfig[width=0.45\textwidth]{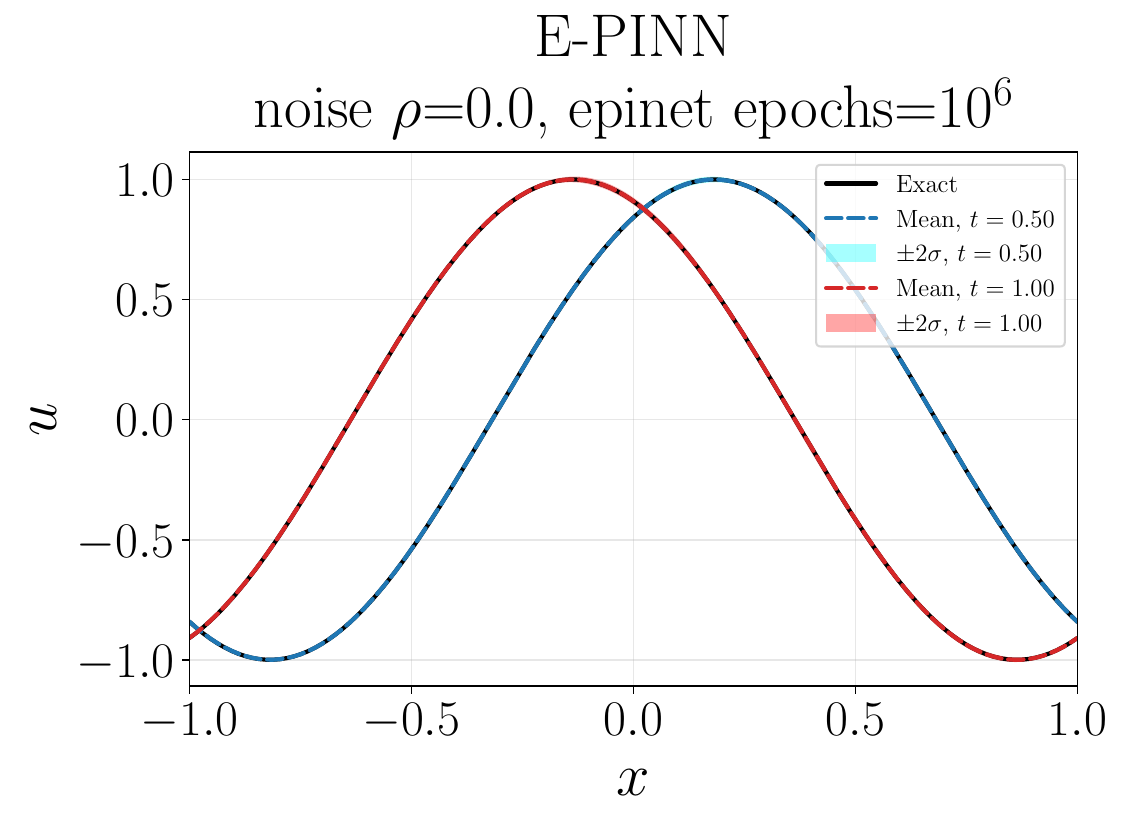}
\caption{Effect of epinet training epochs. Shown for epinet epochs $\{10^4,10^5,10^6\}$. Each panel shows the predictive mean and $\pm2\sigma$ epistemic bands for $u$; all other settings are fixed across runs.}
    \label{fig:Ablation_Epochs}
\end{figure}

This ablation investigates the effect of the epinet training budget under a standardized configuration in which the architecture remains at the defaults $(32,32,32)$ and $N_{\text{colloc}}=1000$ is used for stability. We vary only the number of epinet training epochs while keeping the base network fixed at its default budget. Physics and boundary conditions are exact, and no interior $u$ sensors are used. The epinet epochs are $\{10^{4},10^{5},10^{6}\}$. Figure~\ref{fig:Ablation_Epochs} shows the three cases used in the study.

\begin{table}[H]
\centering
\caption{Epinet epoch ablation: E-PINN metrics versus epinet epochs. RMSE is computed with the predictive mean $\mu_u$. (Sharpness/time/RMSE: lower is better; coverage: higher is better.)}
\label{tab:Table_Epochs}
\sisetup{detect-weight=true,detect-family=true,mode=text}
\begin{tabular*}{\textwidth}{@{\extracolsep{\fill}}
l
S[table-format=1.2]
S[table-format=1.2]
S[table-format=1.4]
S[table-format=4.2]
S[table-format=4.2]
}
\toprule
\multicolumn{1}{c}{epinet epochs} &
\multicolumn{1}{c}{Sharpness $\mu^w_{2\sigma}$} &
\multicolumn{1}{c}{Coverage (95\%)} &
\multicolumn{1}{c}{RMSE} &
\multicolumn{1}{c}{Time (s)} & \multicolumn{1}{c}{Time (epinet) (s)} \\
\midrule
$10^{4}$ & 0.39 & 1.00 & 0.0118 & 1248.34 & 275.17 \\
$10^{5}$ & 0.03 & 1.00 & 0.0004 & 4833.58 & 3599.50 \\
$10^{6}$ & 0.01 & 1.00 & 0.0003 & 28408.96 & 27419.18 \\
\bottomrule
\end{tabular*}
\end{table}

From $10^{4}$ to $10^{5}$ epinet epochs (Table~\ref{tab:Table_Epochs}), sharpness decreases markedly and RMSE drops by an order of magnitude while coverage remains conservative (\(\approx 1.00\)). Total time increases because of the longer epinet schedule, indicating that the epinet at $10^{4}$ epochs is under-optimized and benefits from continued training. At $10^{6}$ epochs, bands tighten further and RMSE decreases marginally while coverage remains at 1.00. The incremental gains beyond $10^{5}$ come at substantially higher cost, suggesting diminishing returns past $10^{5}$ under this configuration.

\subsubsection{Effect of the number of training epochs in the base network}
\begin{figure}[H]
    \centering
    \incfig[width=0.45\textwidth]{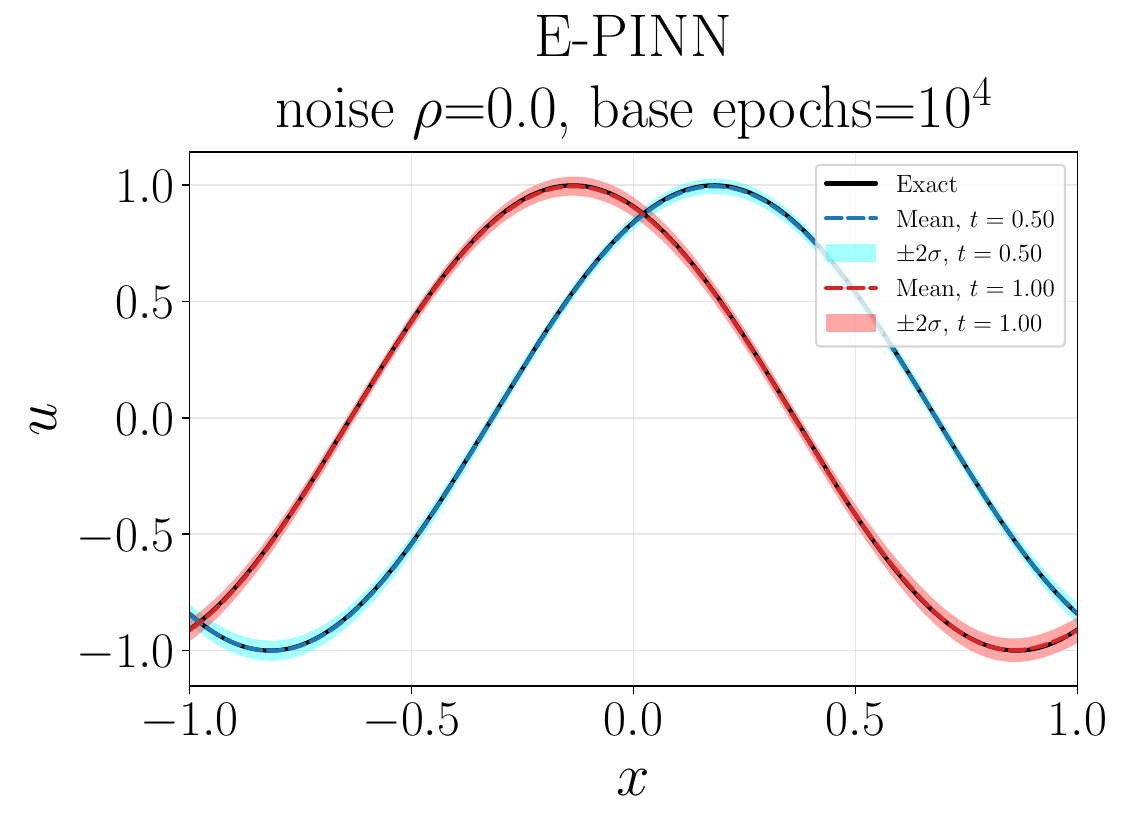}
    \incfig[width=0.45\textwidth]{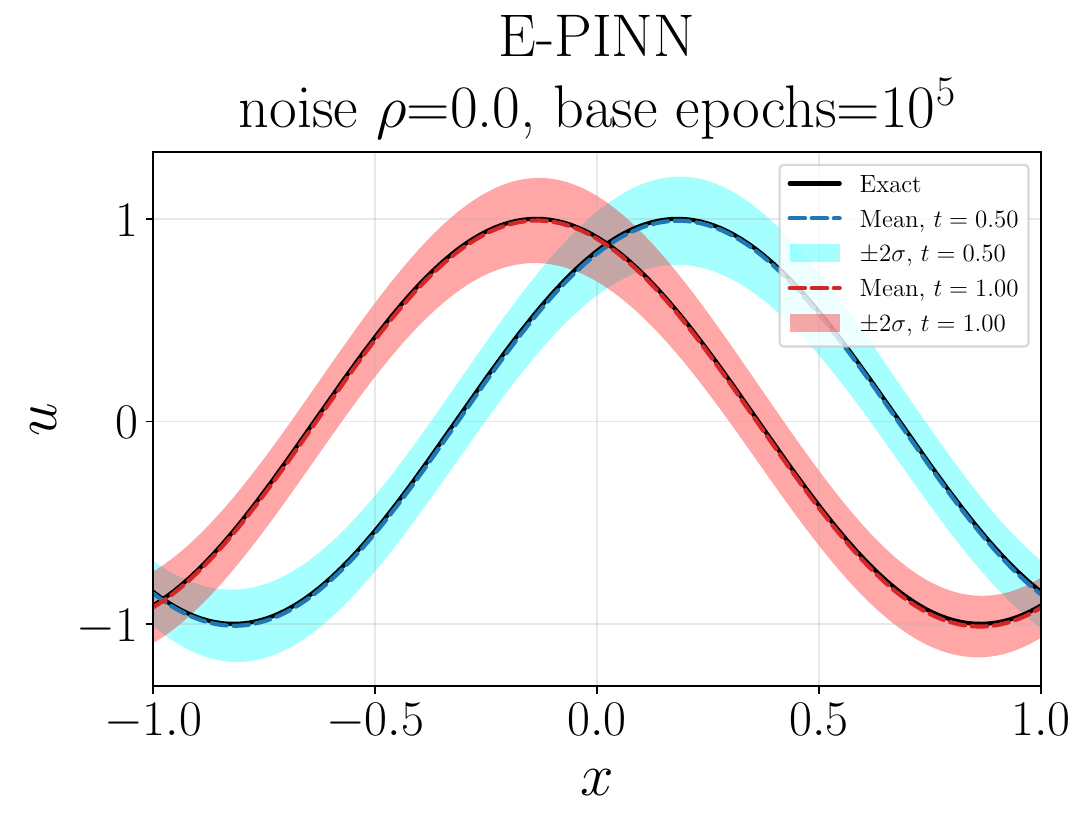}
    \incfig[width=0.45\textwidth]{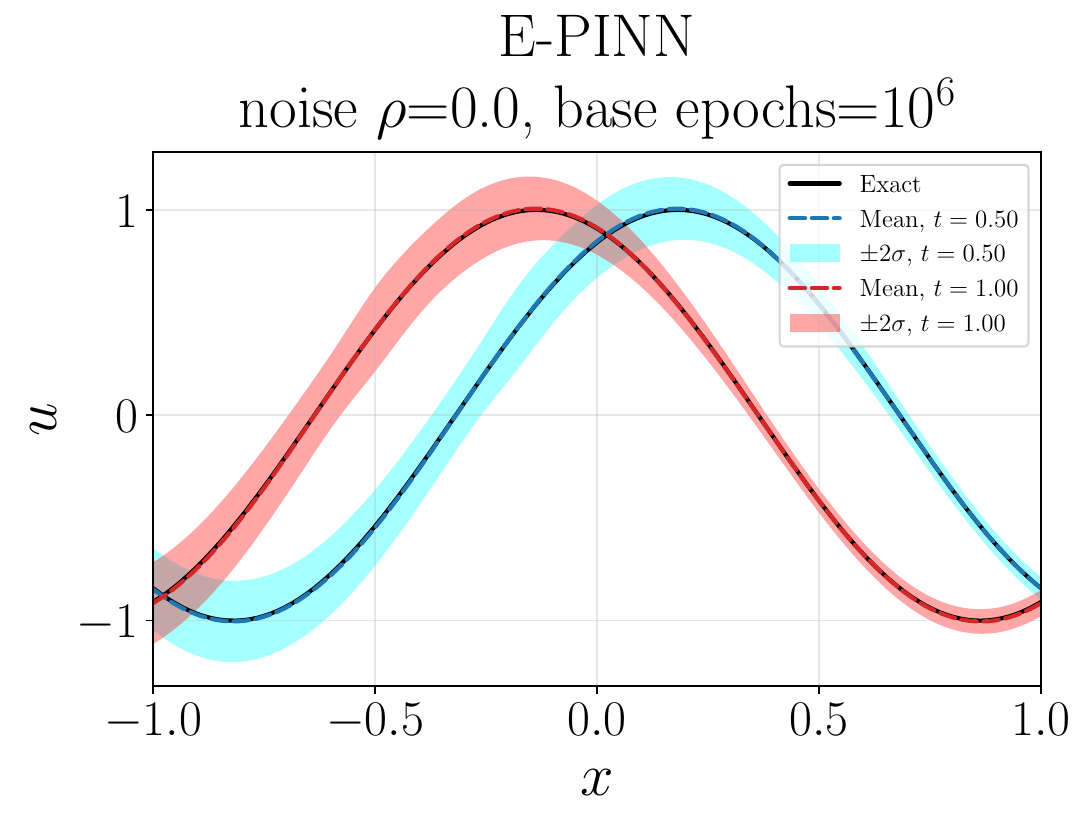}
\caption{Effect of base training epochs. Shown for base epochs $\{10^4,10^5,10^6\}$. Each panel shows the predictive mean and $\pm2\sigma$ epistemic bands for $u$; epinet epochs are held fixed.}
    \label{fig:Ablation_Epochs_Base}
\end{figure}

This ablation studies the effect of the base network training budget. The configuration and data channels match the previous subsection, with the default architecture $(32,32,32)$ and $N_{\text{colloc}}=1000$. We vary only the base epochs $\{10^{4},10^{5},10^{6}\}$; the epinet training remains fixed to $10^{4}$ epochs. Figure~\ref{fig:Ablation_Epochs_Base} shows the three cases used in the study.

For consistency across runs, wall-clock throughput for B-PINN uses $M=11{,}000$ total samples with $1{,}000$ burn-in steps, matching the default configuration.

\begin{table}[H]
\centering
\caption{Base epoch ablation: E-PINN metrics versus base epochs. RMSE is computed with the predictive mean $\mu_u$. (Sharpness/time/RMSE: lower is better; coverage: higher is better.)}
\label{tab:Table_Epochs_Base}
\sisetup{detect-weight=true,detect-family=true,mode=text}
\begin{tabular*}{\textwidth}{@{\extracolsep{\fill}}
l
S[table-format=1.2]
S[table-format=1.2]
S[table-format=1.4]
S[table-format=4.2]
S[table-format=4.2]
}
\toprule
\multicolumn{1}{c}{Base epochs} &
\multicolumn{1}{c}{Sharpness $\mu^w_{2\sigma}$} &
\multicolumn{1}{c}{Coverage (95\%)} &
\multicolumn{1}{c}{RMSE} &
\multicolumn{1}{c}{Time (s)} & \multicolumn{1}{c}{Time (epinet) (s)} \\
\midrule
$10^{4}$ & 0.07 & 1.00 & 0.0027 & 718.14 & 285.06 \\
$10^{5}$ & 0.39 & 1.00 & 0.0118 & 1772.63 & 317.59 \\
$10^{6}$ & 0.29 & 1.00 & 0.0039 & 9541.34 & 313.91 \\
\bottomrule
\end{tabular*}
\end{table}

With the epinet fixed at $10^{4}$ epochs (Table~\ref{tab:Table_Epochs_Base}), $10^{4}$ base epochs produce the tightest bands and the lowest RMSE. Increasing the base budget to $10^{5}$ inflates sharpness and error; at $10^{6}$ the RMSE recovers while bands remain wider than at $10^{4}$. Total time grows with the base budget and dominates overall throughput at $10^{6}$. These trends suggest that, in physics-only training of the base, coupling the base schedule to the epinet schedule matters: longer base optimization can reduce the utility of the base features to the epinet under a fixed epinet budget, which the epinet expresses as wider corrections with larger variance. In this regime, relatively shorter base training paired with a fixed epinet budget yields better overall accuracy and sharper calibrated intervals.

\subsubsection{Effect of interior $u$ noise}
\begin{figure}[H]
    \centering
    \incfig[width=0.45\textwidth]{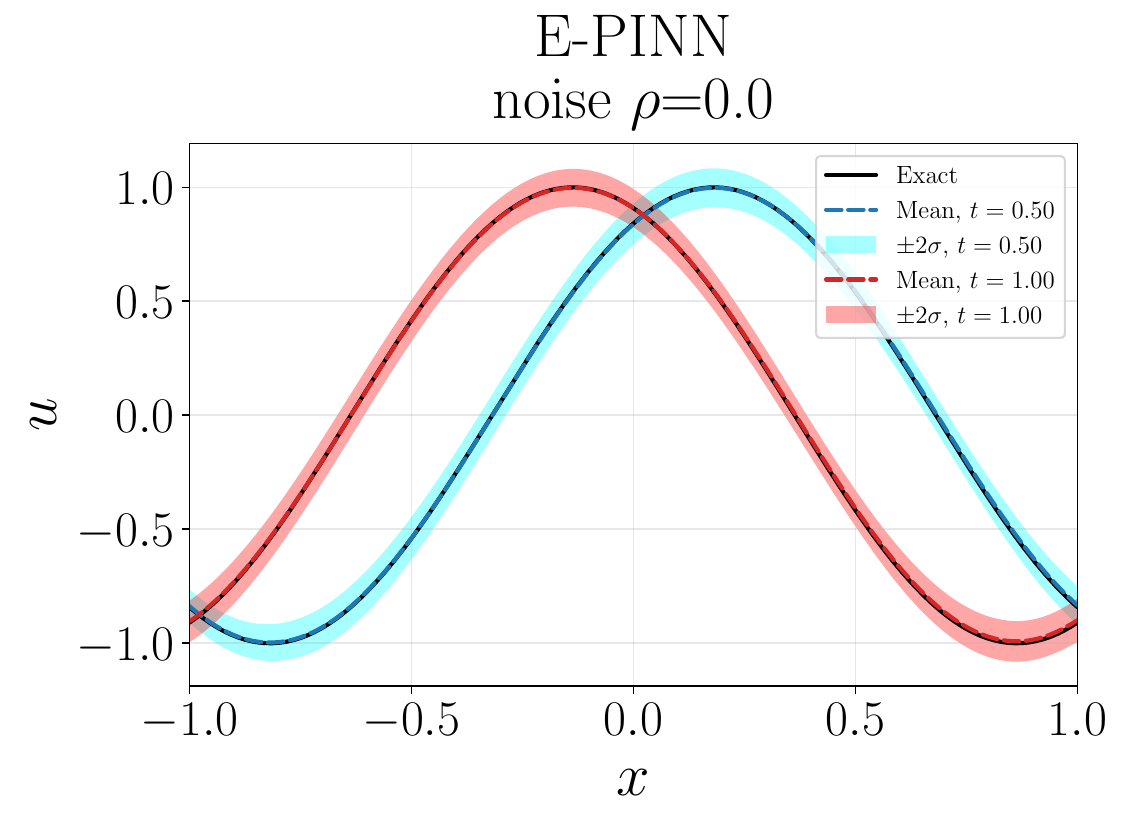}
    \incfig[width=0.45\textwidth]{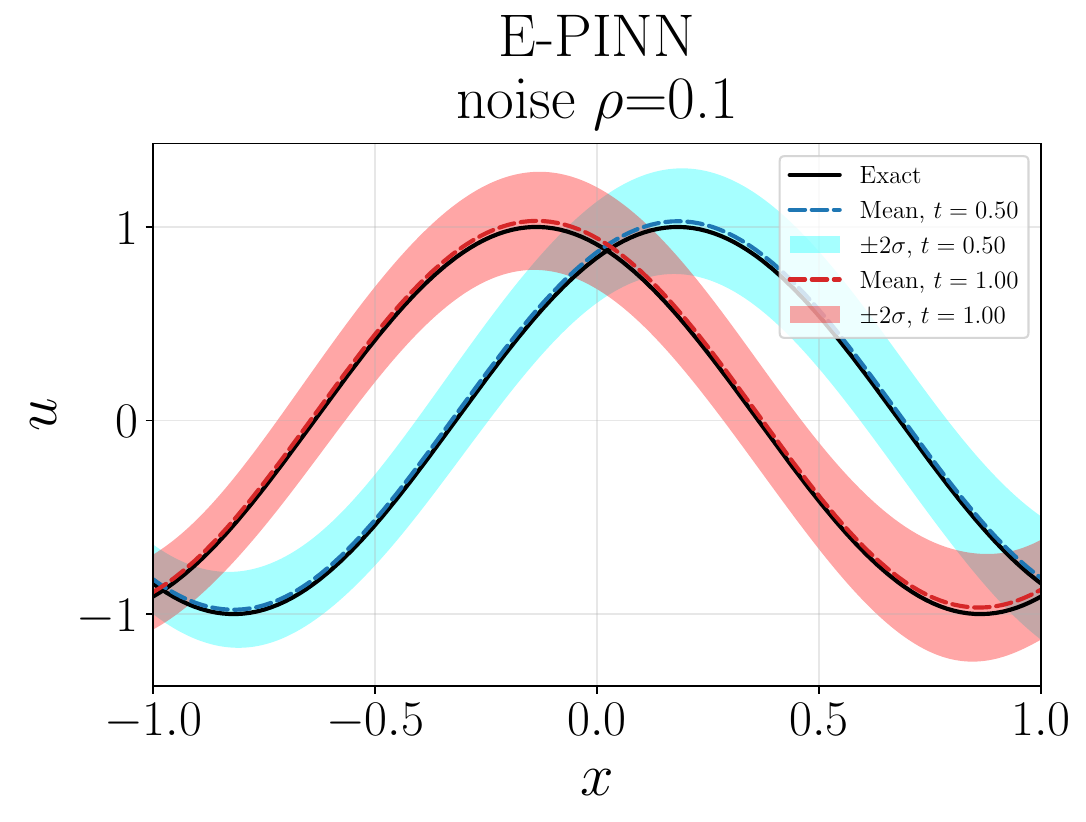}
    \incfig[width=0.45\textwidth]{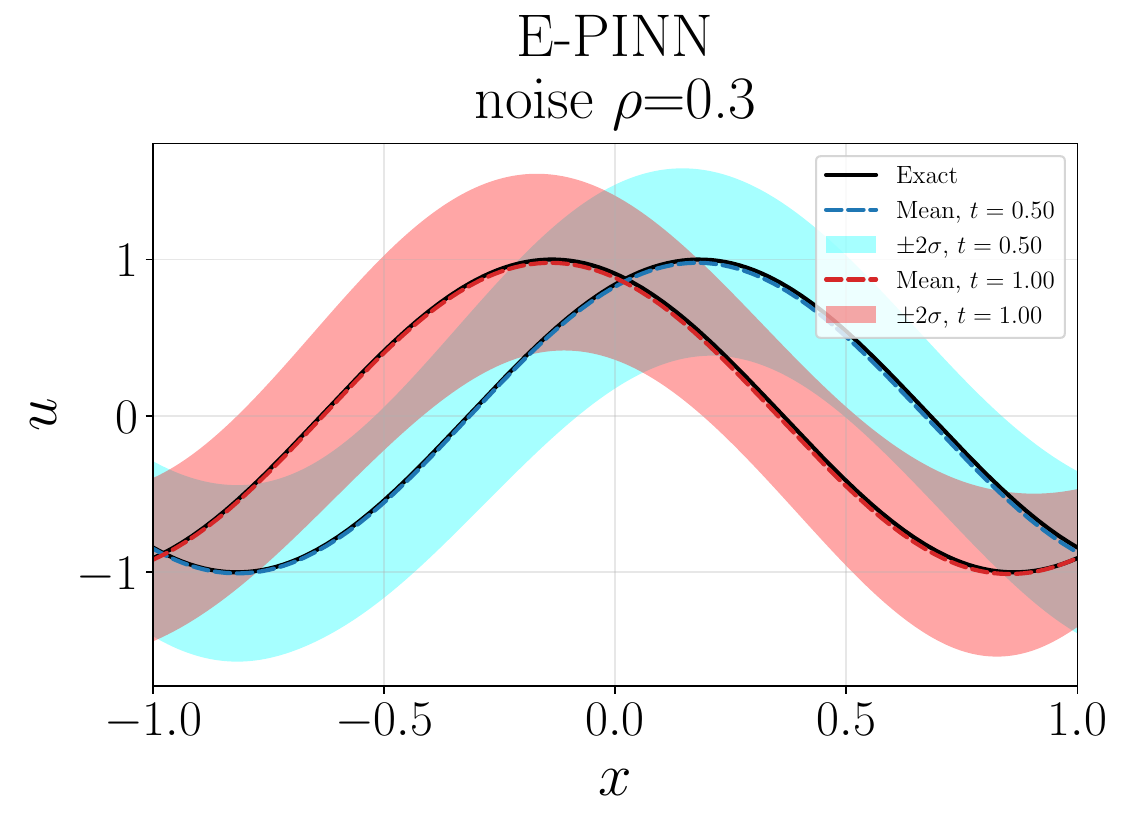}
    \caption{Effect of interior $u$ noise. Shown for $\rho\in\{0.00,0.10,0.30\}$ with $\sigma=\rho\,\lVert u\rVert_{\infty}$. The case $\rho=0.00$ uses physics only; $\rho>0$ adds noisy interior $u$ data. Each panel shows the predictive mean and $\pm2\sigma$ epistemic bands for $u$; other settings are fixed.}
    \label{fig:Ablation_Noise}
\end{figure}

This section studies the effect of interior measurement noise in the data augmented setting. We use three levels: $\rho\in\{0.00,0.10,0.30\}$, where the standard deviation is $\sigma=\rho\,\lVert u\rVert_{\infty}$. The case $\rho=0.00$ uses physics only, and $\rho>0$ uses physics plus noisy interior $u$ data. All other hyperparameters are kept at their defaults. Figure~\ref{fig:Ablation_Noise} shows the three cases used in the study. Table~\ref{tab:ablation_burgers_noise} reports the measured metrics.

\begin{table}[H]
\centering
\caption{Noise ablation: E-PINN metrics versus $\rho$ for Burgers. RMSE is computed with the predictive mean $\mu_u$. (Sharpness/time/RMSE: lower is better; coverage: higher is better.)}
\label{tab:ablation_burgers_noise}
\sisetup{detect-weight=true,detect-family=true,mode=text}
\begin{tabular*}{\textwidth}{@{\extracolsep{\fill}}
S[table-format=1.2]
S[table-format=1.2]
S[table-format=1.2]
S[table-format=1.4]
S[table-format=4.2]
S[table-format=4.2]
}
\toprule
\multicolumn{1}{c}{$\rho$} &
\multicolumn{1}{c}{Sharpness $\mu^w_{2\sigma}$} &
\multicolumn{1}{c}{Coverage (95\%)} &
\multicolumn{1}{c}{RMSE} &
\multicolumn{1}{c}{Time (s)} & \multicolumn{1}{c}{Time (epinet) (s)} \\
\midrule
0.00 & 0.17 & 1.00 & 0.0046 & 4718.98 & 1303.11 \\
0.10 & 0.52 & 1.00 & 0.0276 & 4625.84 & 1294.11 \\
0.30 & 1.16 & 1.00 & 0.0220 & 5431.61 & 1509.17 \\
\bottomrule
\end{tabular*}
\end{table}

Sharpness increases with noise while coverage remains conservative (\(\approx 1.00\)) in all cases (Table~\ref{tab:ablation_burgers_noise}). RMSE increases from $\rho=0.00$ to $\rho=0.10$ and is lower at $\rho=0.30$ than at $\rho=0.10$. Under the fixed loss weights, noisier interior measurements reduce the effective information content of the data channel relative to the PDE channel. At higher noise, the optimizer leans more on the physics residual, which regularizes the mean and can yield a lower RMSE on the noiseless target, while the epinet widens the bands to preserve empirical coverage. The epinet training time is similar across noise levels (same size and epochs) and increases slightly at higher noise; total time rises at $\rho=0.30$ due to longer base optimization.

\section{Conclusions}

This work introduces E-PINNs, a framework that extends physics-informed neural networks by efficiently representing epistemic uncertainty through a compact, companion epinet. The epinet is trained on stop-gradient features from a pre-trained PINN and uses a low-dimensional epistemic index for Monte Carlo estimation. In practice, this attaches uncertainty to an existing surrogate with modest additional cost.

Extensive evaluations across forward PDE problems and an inverse problem indicate that E-PINNs produce conservative uncertainty bands with accurate means while maintaining a favorable balance between uncertainty quality and computational cost. Relative to Bayesian PINNs based on Hamiltonian Monte Carlo, E-PINNs generally achieve comparable or better accuracy at lower wall-clock times on CPU; in one inverse case B-PINN ran faster but underestimated $\kappa$ (posterior mean $\approx 0.078$ vs true $0.10$). Compared with Dropout-PINNs, E-PINNs deliver sharper intervals with better empirical coverage under the settings tested.

Ablation studies provide guidance for practical use. Under a fixed training budget, moderate numbers of collocation points suffice to calibrate the epinet features; pushing them much higher without adjusting the schedule can widen intervals. Modest epinet capacity is effective, while substantially larger widths increase variance and error without improving coverage. Extending epinet training reduces error and tightens bands up to about $10^{5}$ epochs, after which improvements are marginal relative to cost. With higher interior $u$ noise, intervals widen while empirical coverage remains near nominal levels.

In summary, E-PINNs offer a computationally efficient approach to epistemic uncertainty for PINNs. The method attaches to existing surrogates with a small training overhead and performs well across the problems studied. Future work will extend E-PINNs to larger operator-learning settings and evaluate performance on experimental datasets.

\section{Code and data availability}
All code, trained models, and data required to replicate the examples presented in this paper will be released upon publication.

\section{Acknowledgements}
This project was completed with support from the U.S. Department of Energy, Advanced Scientific Computing Research program, under the "Resolution-invariant deep learning for accelerated propagation of epistemic and aleatory uncertainty in multi-scale energy storage systems, and beyond" project (Project No. 81824) and under the "Uncertainty Quantification for Multifidelity Operator Learning (MOLUcQ)" project (Project No. 81739). The computational work was performed using PNNL Institutional Computing at Pacific Northwest National Laboratory. Pacific Northwest National Laboratory (PNNL) is a multi-program national laboratory operated for the U.S. Department of Energy (DOE) by Battelle Memorial Institute under Contract No. DE-AC05-76RL01830.

\clearpage
\appendix
\section*{Appendix}
\label{app:appendix}

\subsection*{Additional 2D nonlinear Poisson figures}
\label{app:2d_nonlin_appendix}
\begin{figure}[H]
    \centering
    \begin{subfigure}[b]{0.32\textwidth}\centering
        \incfig[width=\textwidth]{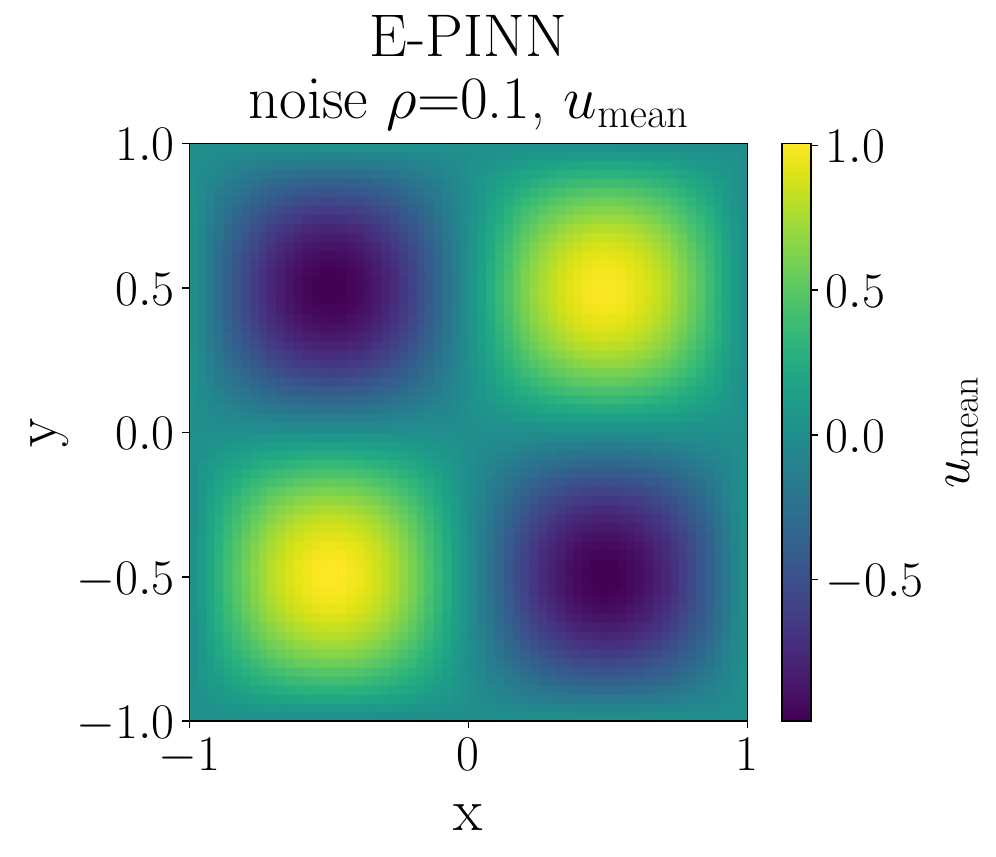}
    \end{subfigure}\hfill
    \begin{subfigure}[b]{0.32\textwidth}\centering
        \incfig[width=\textwidth]{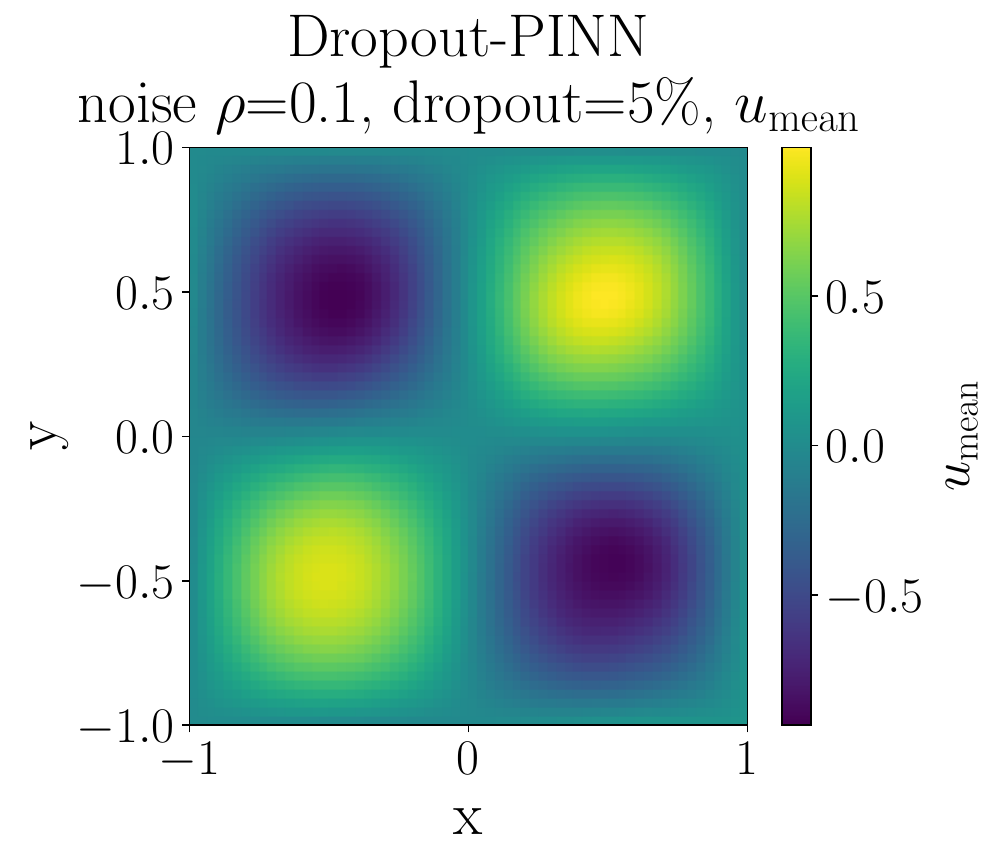}
    \end{subfigure}\hfill
    \begin{subfigure}[b]{0.32\textwidth}\centering
        \incfig[width=\textwidth]{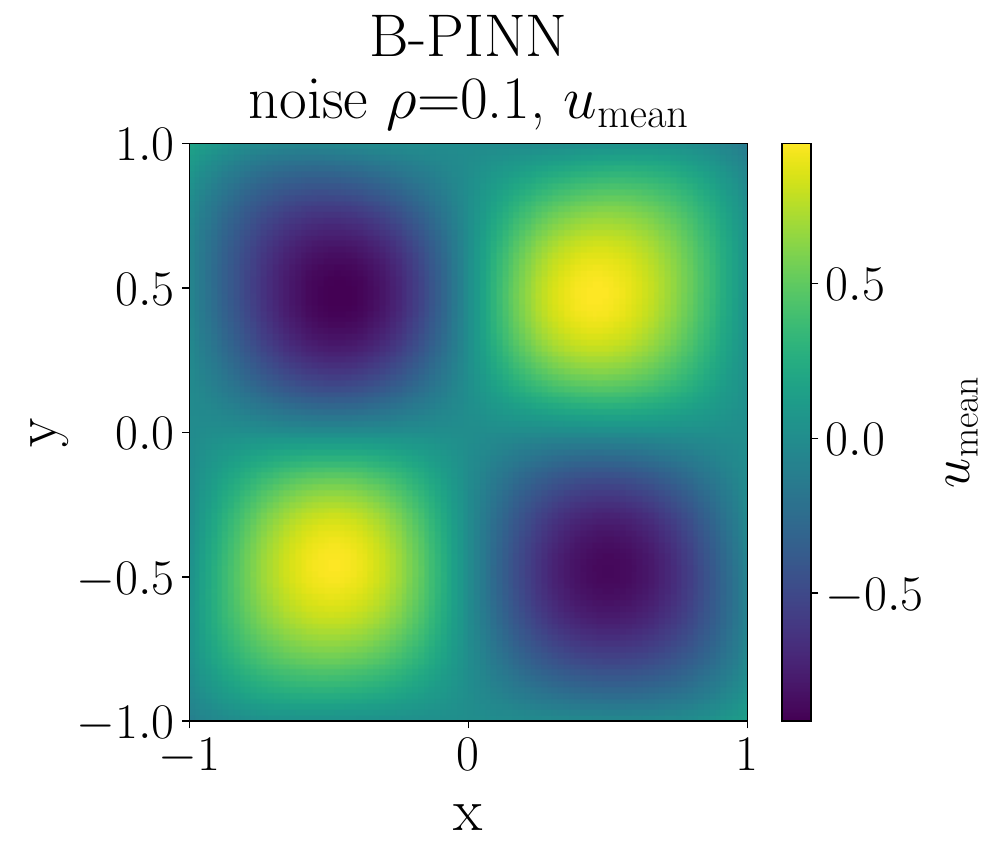}
    \end{subfigure}
    \par\vspace{0.4em}
    \begin{subfigure}[b]{0.32\textwidth}\centering
        \incfig[width=\textwidth]{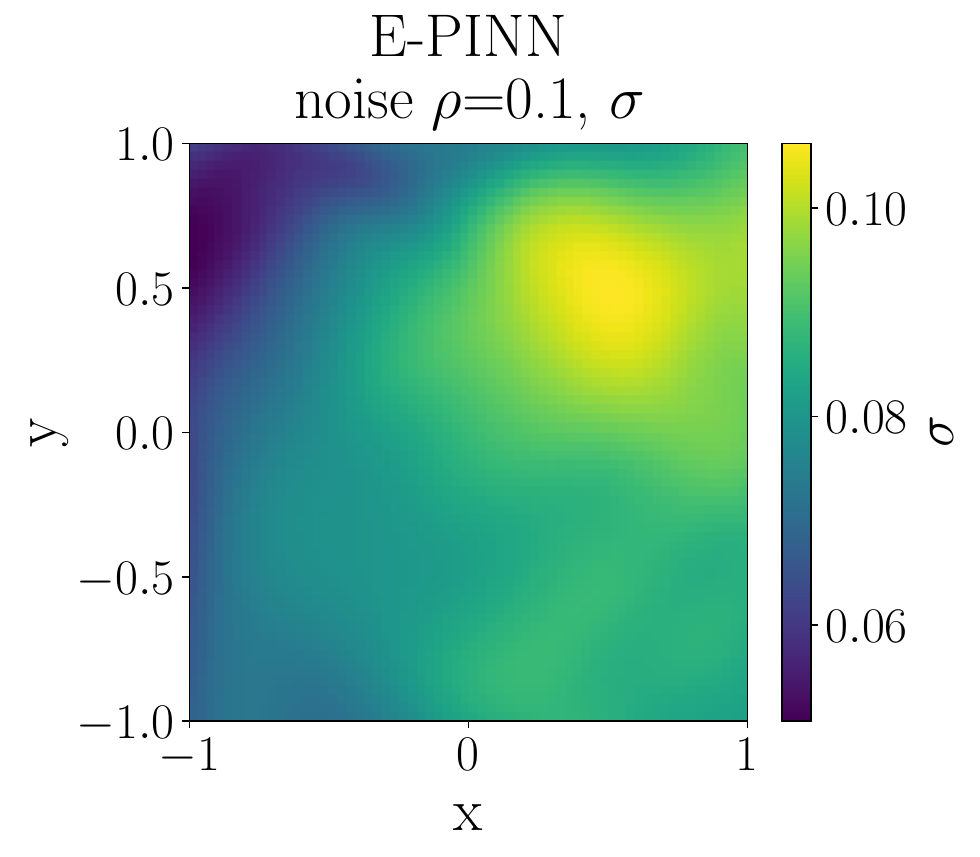}
    \end{subfigure}\hfill
    \begin{subfigure}[b]{0.32\textwidth}\centering
        \incfig[width=\textwidth]{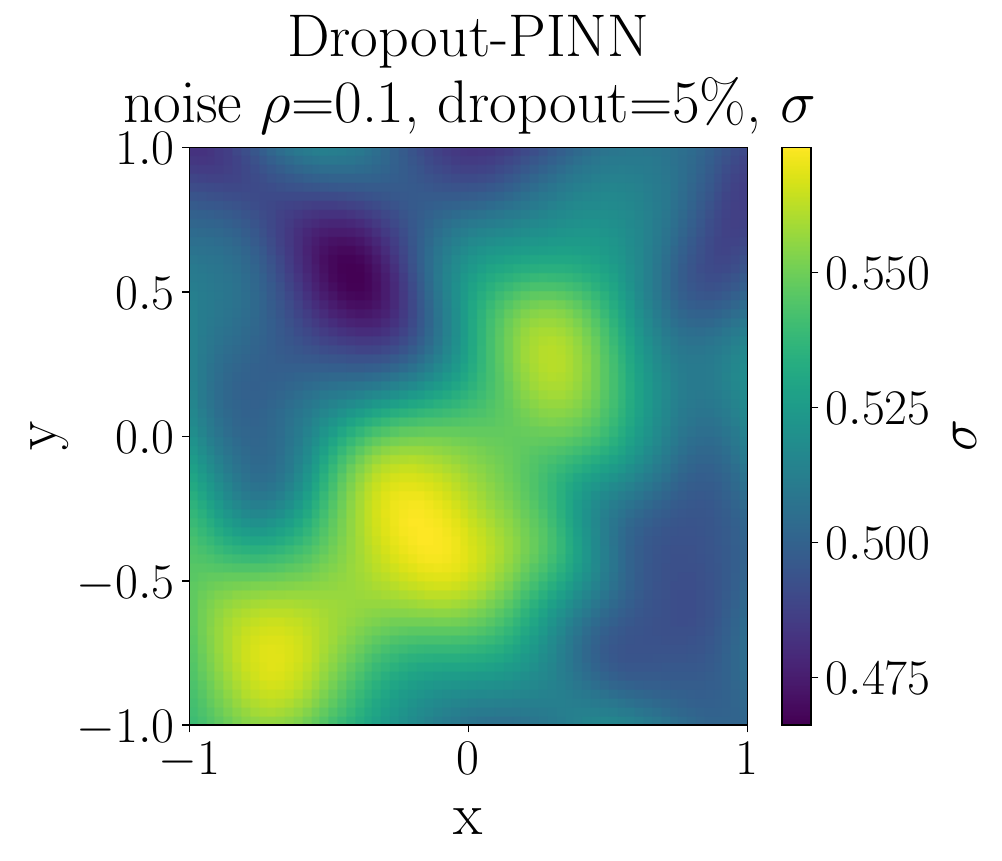}
    \end{subfigure}\hfill
    \begin{subfigure}[b]{0.32\textwidth}\centering
        \incfig[width=\textwidth]{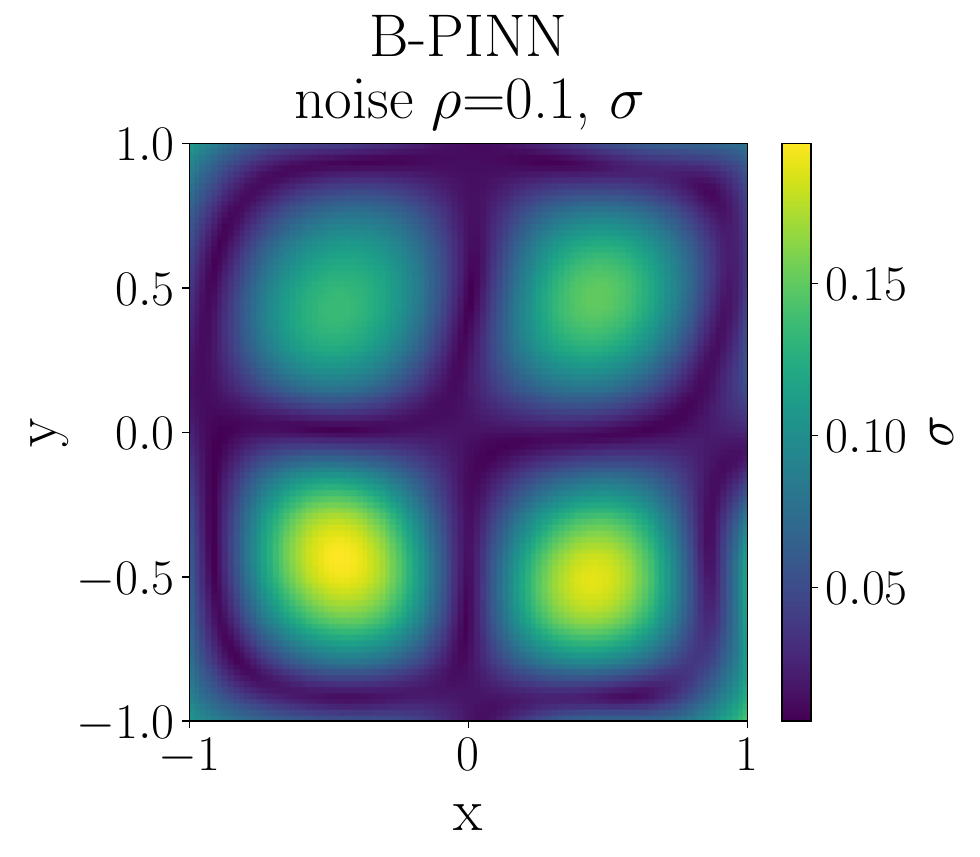}
    \end{subfigure}
    \par\vspace{0.4em}
    \begin{subfigure}[b]{0.32\textwidth}\centering
        \incfig[width=\textwidth]{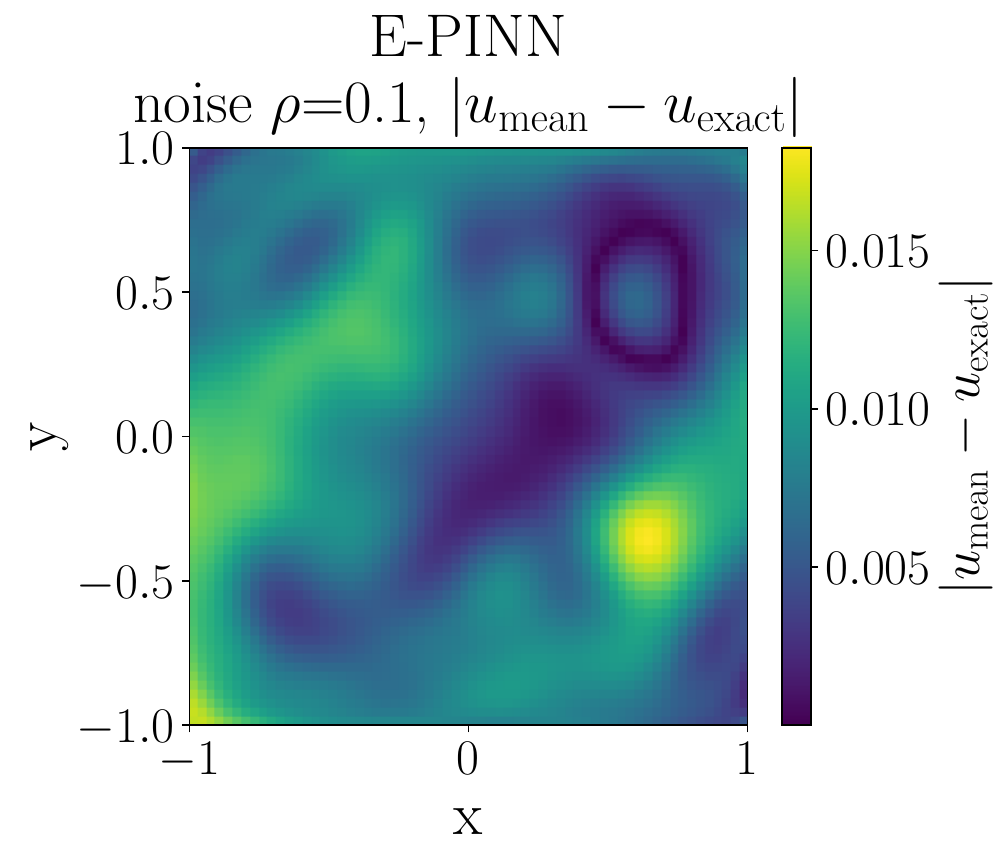}
    \end{subfigure}\hfill
    \begin{subfigure}[b]{0.32\textwidth}\centering
        \incfig[width=\textwidth]{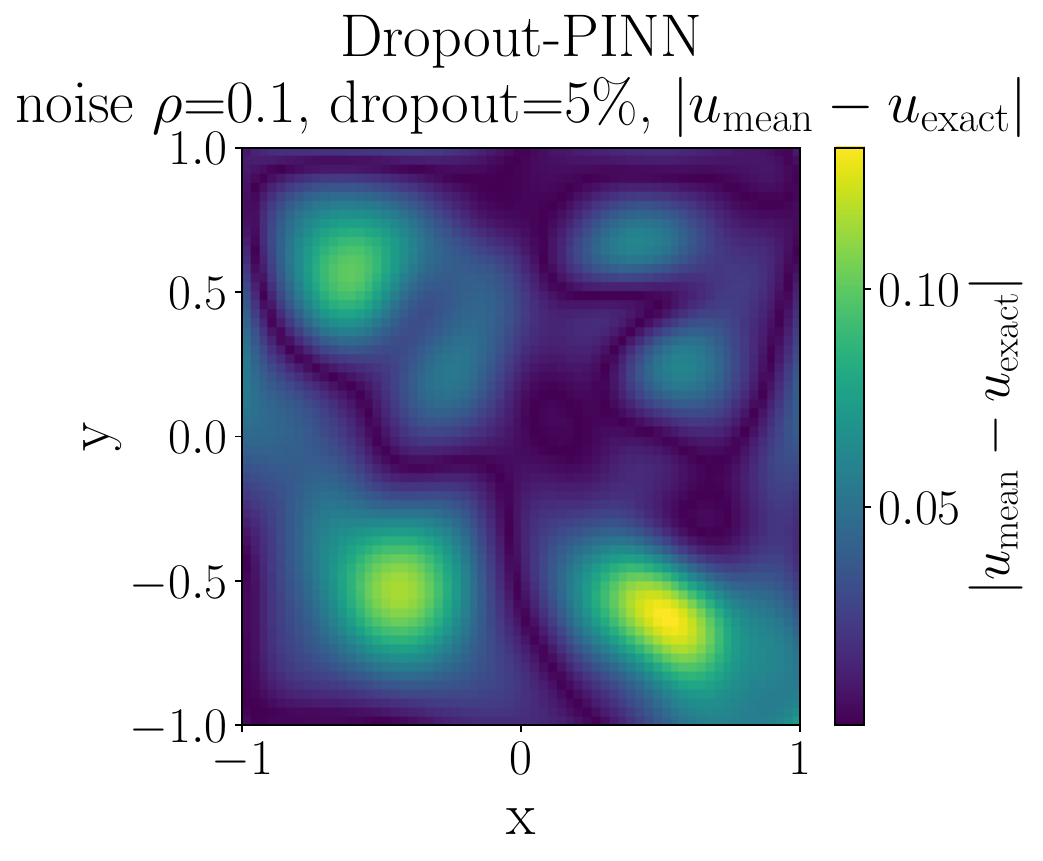}
    \end{subfigure}\hfill
    \begin{subfigure}[b]{0.32\textwidth}\centering
        \incfig[width=\textwidth]{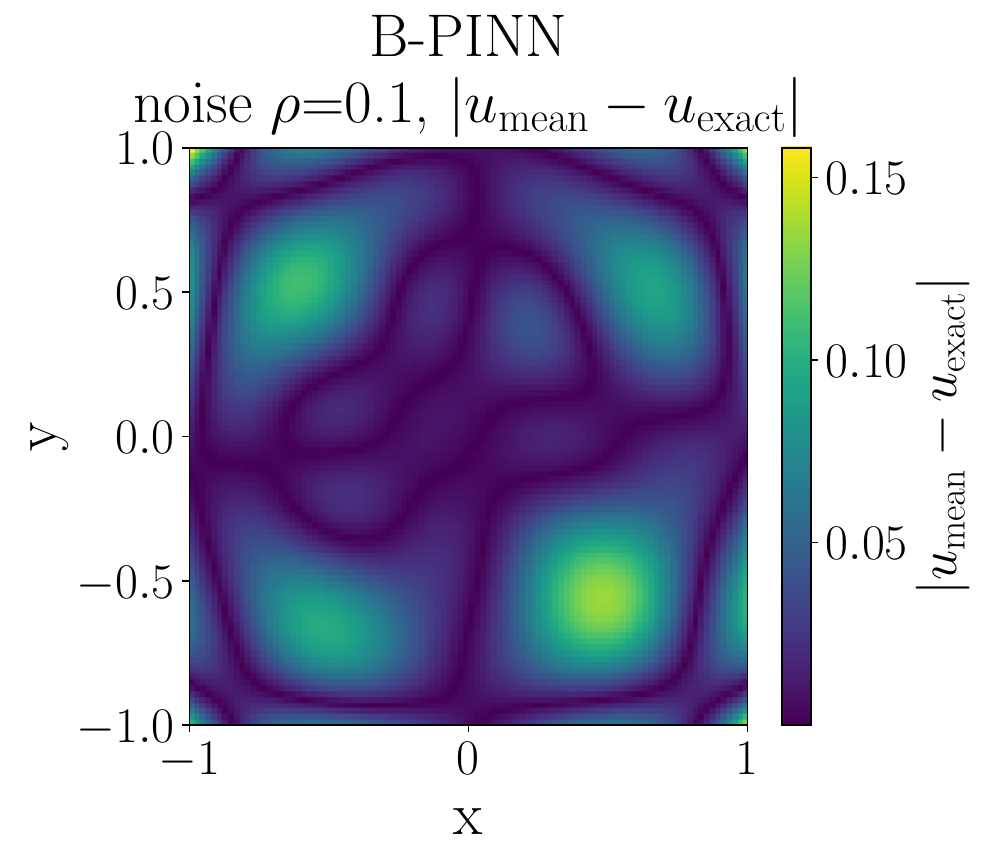}
    \end{subfigure}
    \caption{2D nonlinear Poisson ($\rho=0.10$): top, mean $u$; middle, epistemic standard deviation; bottom, absolute error. Columns: E-PINN (left), Dropout-PINN (center), B-PINN (right).}
    \label{fig:2D_nlP_grid_0.10_app}
\end{figure}

Figures~\ref{fig:2D_nlP_grid_0.10_app} provides the $\rho=0.10$ comparison (mean, standard deviation, and error) for completeness.

\clearpage

\clearpage
\subsection*{Additional 1D Poisson figures}
\begin{figure}[H]
    \centering
    \begin{subfigure}[b]{0.45\textwidth}
        \centering
        \incfig[width=\textwidth]{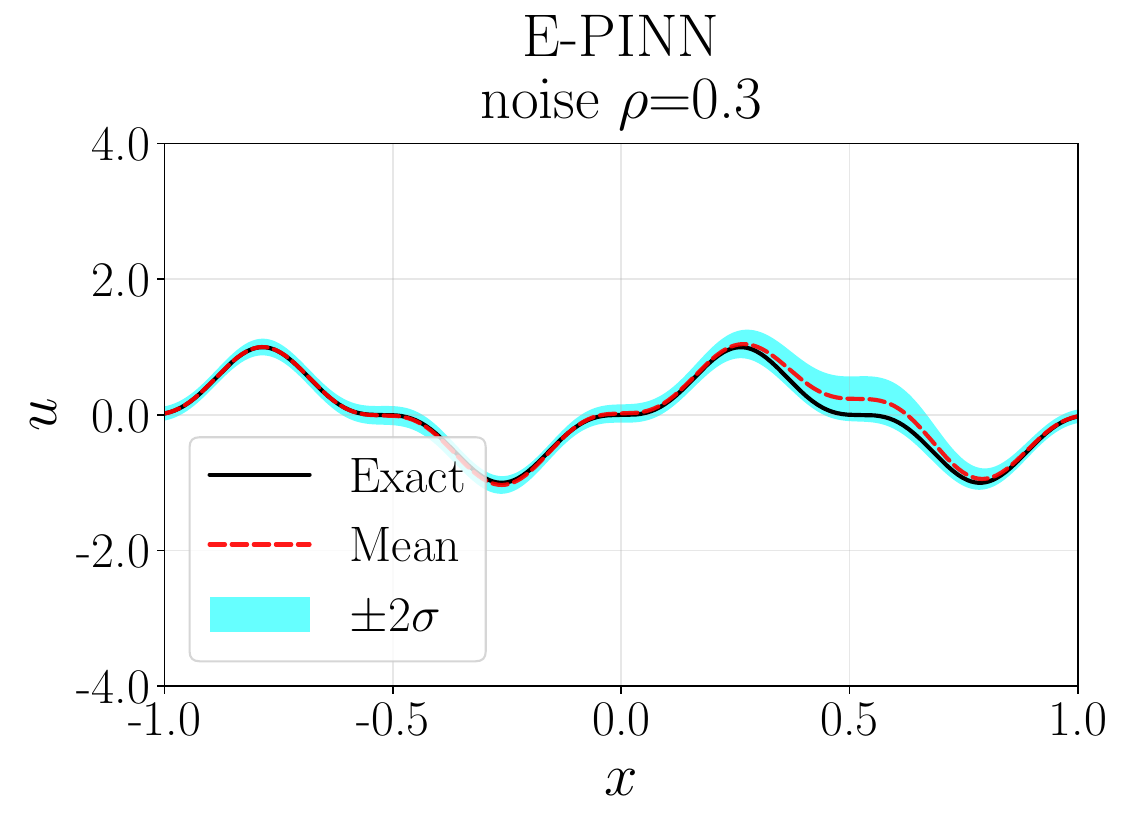}
        \caption{E-PINN}
    \end{subfigure}
    \hfill
    \begin{subfigure}[b]{0.45\textwidth}
        \centering
        \incfig[width=\textwidth]{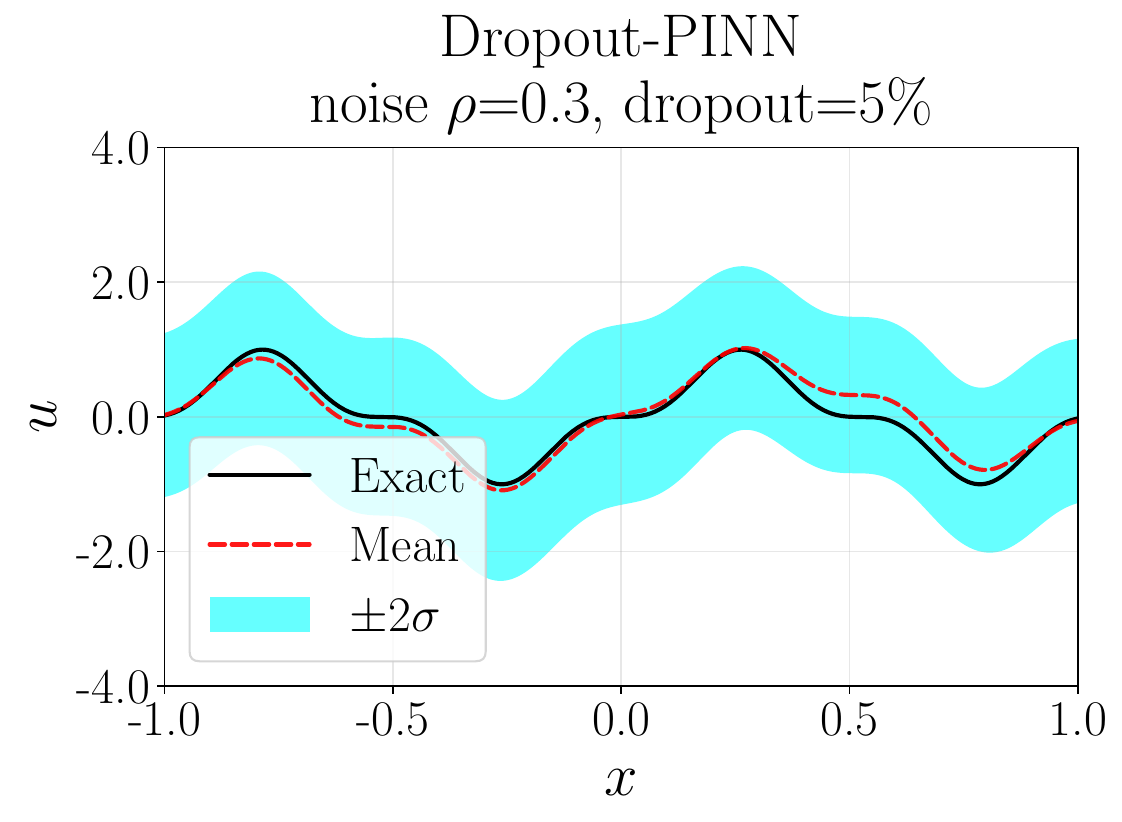}
        \caption{Dropout-PINN (5\%)}
    \end{subfigure}
    \begin{subfigure}[b]{0.45\textwidth}
        \centering
        \incfig[width=\textwidth]{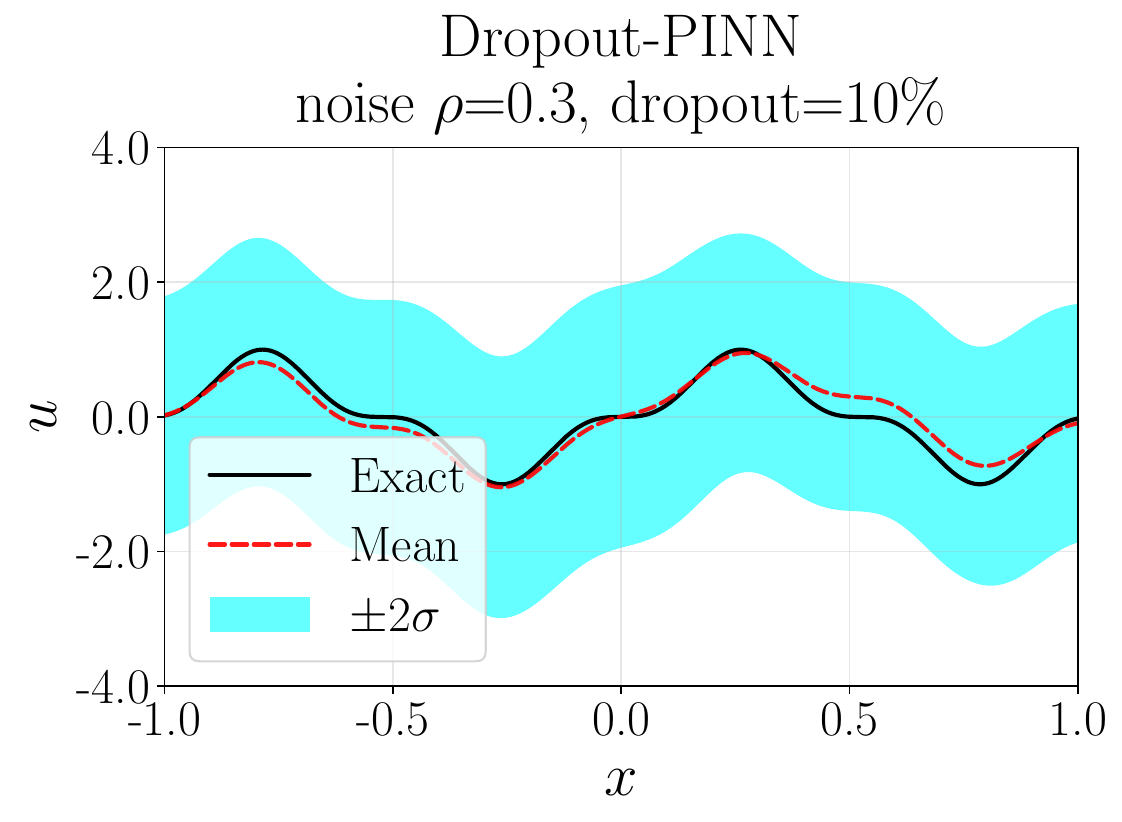}
        \caption{Dropout-PINN (10\%)}
    \end{subfigure}
    \hfill
    \begin{subfigure}[b]{0.45\textwidth}
        \centering
        \incfig[width=\textwidth]{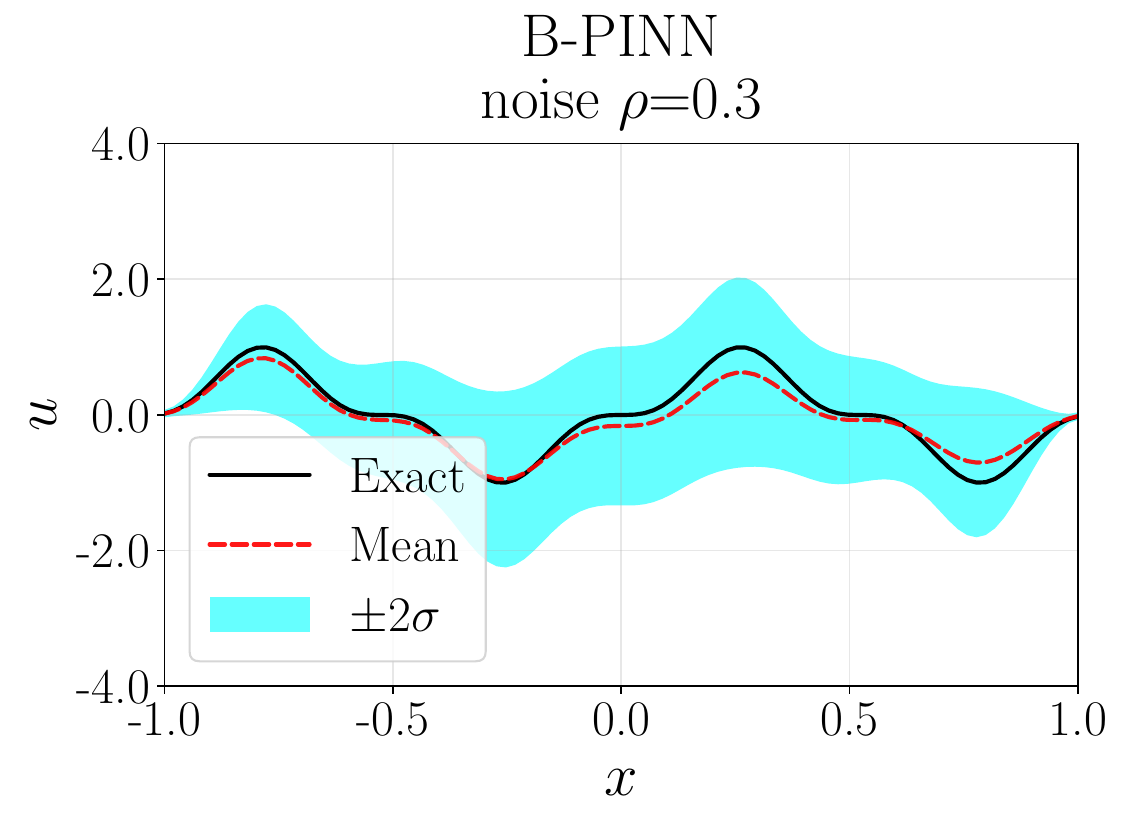}
        \caption{B-PINN}
    \end{subfigure}
    \caption{1D Poisson (Gaussian noise, $\sigma = 0.30\,\lVert u\rVert_{\infty}$): mean and $\pm 2\sigma$ epistemic bands for $u(x)$. Methods: (a) E-PINN, (b) Dropout-PINN (5\%), (c) Dropout-PINN (10\%), (d) B-PINN.}
    \label{fig:app_poisson_030}
\end{figure}
\clearpage

\clearpage
\subsection*{Additional 1D nonlinear Poisson figures}
\begin{figure}[H]
    \centering
    \begin{subfigure}[b]{0.45\textwidth}
        \centering
        \incfig[width=\textwidth]{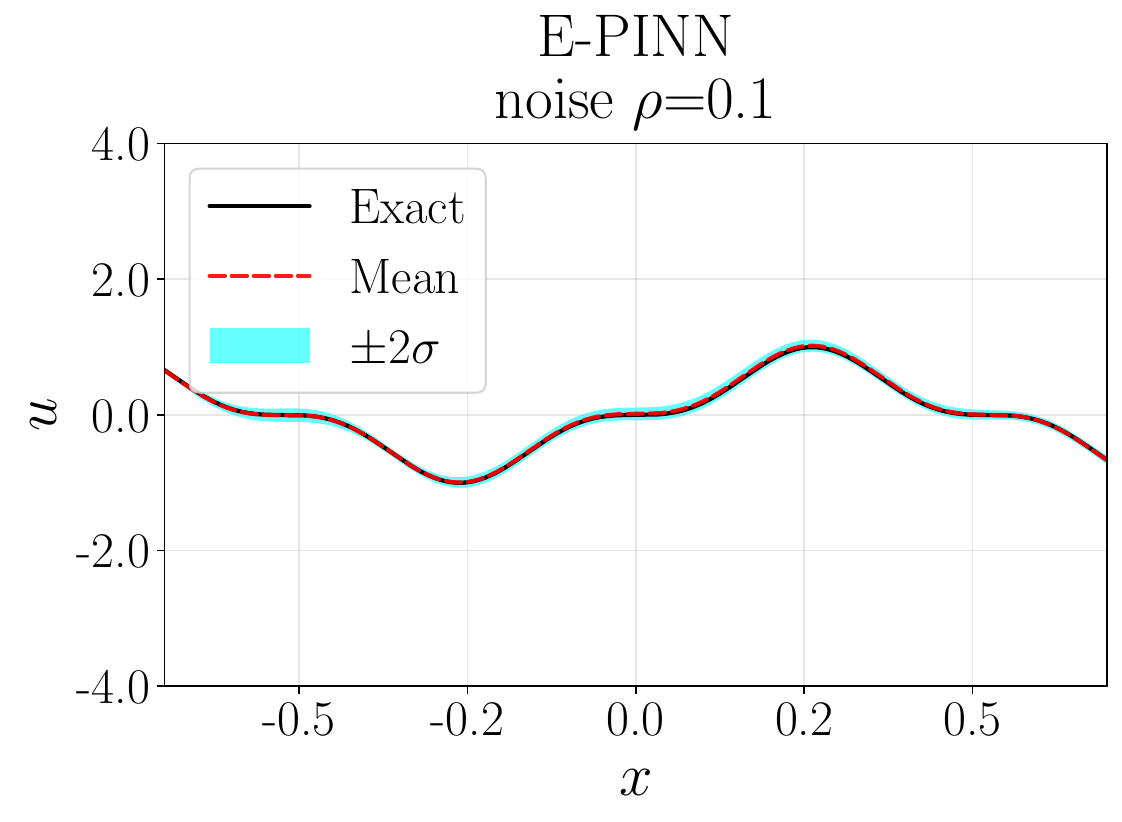}
        \caption{E-PINN}
    \end{subfigure}
    \hfill
    \begin{subfigure}[b]{0.45\textwidth}
        \centering
        \incfig[width=\textwidth]{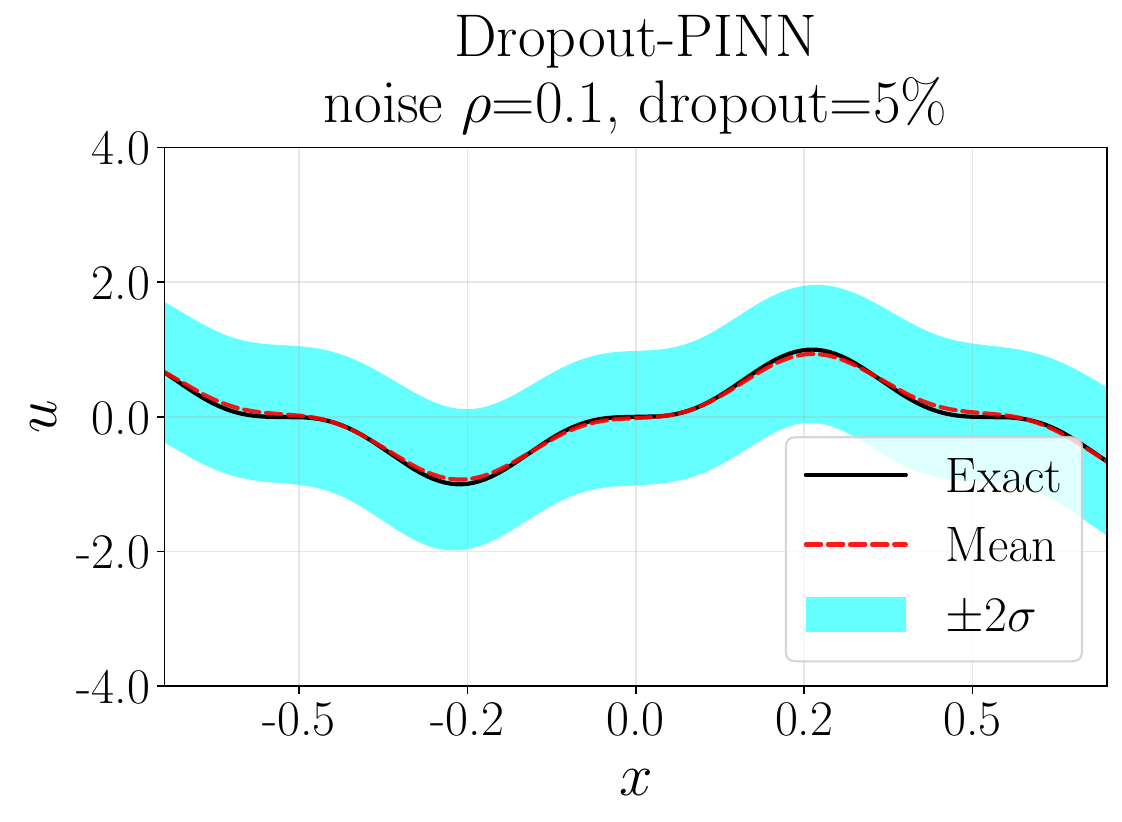}
        \caption{Dropout-PINN (5\%)}
    \end{subfigure}
    \begin{subfigure}[b]{0.45\textwidth}
        \centering
        \incfig[width=\textwidth]{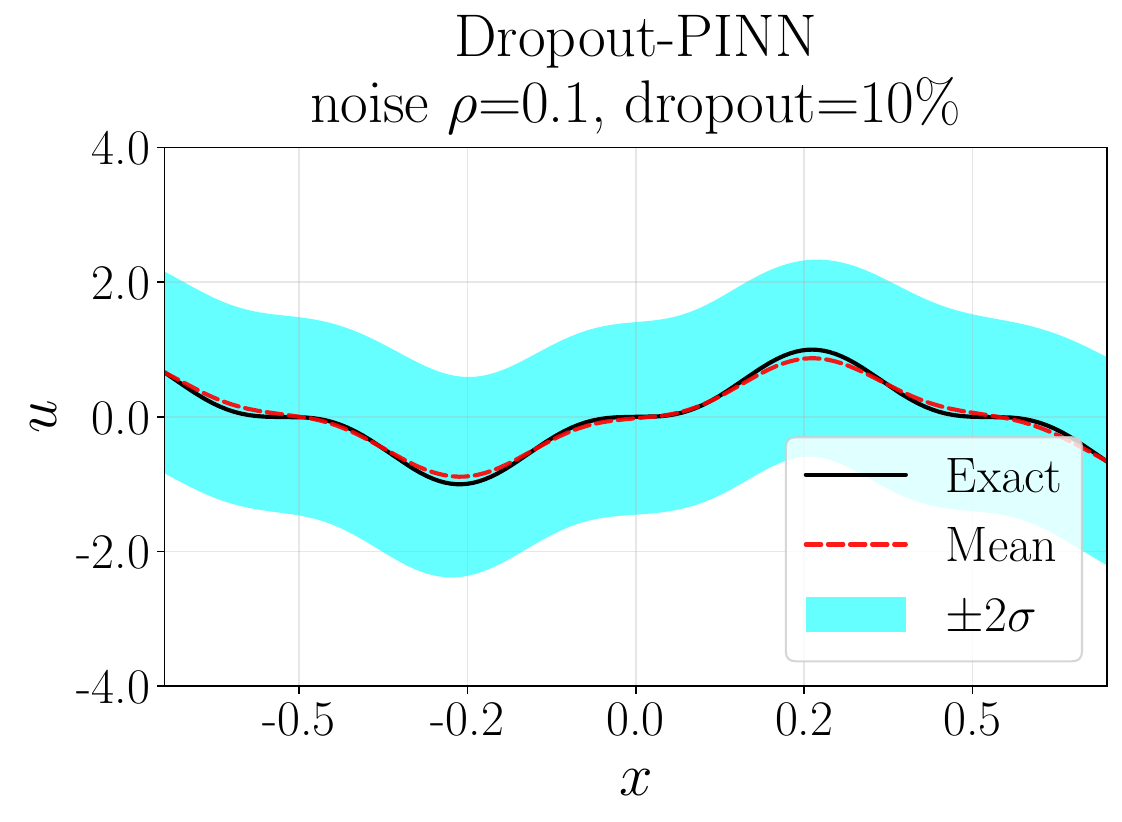}
        \caption{Dropout-PINN (10\%)}
    \end{subfigure}
    \hfill
    \begin{subfigure}[b]{0.45\textwidth}
        \centering
        \incfig[width=\textwidth]{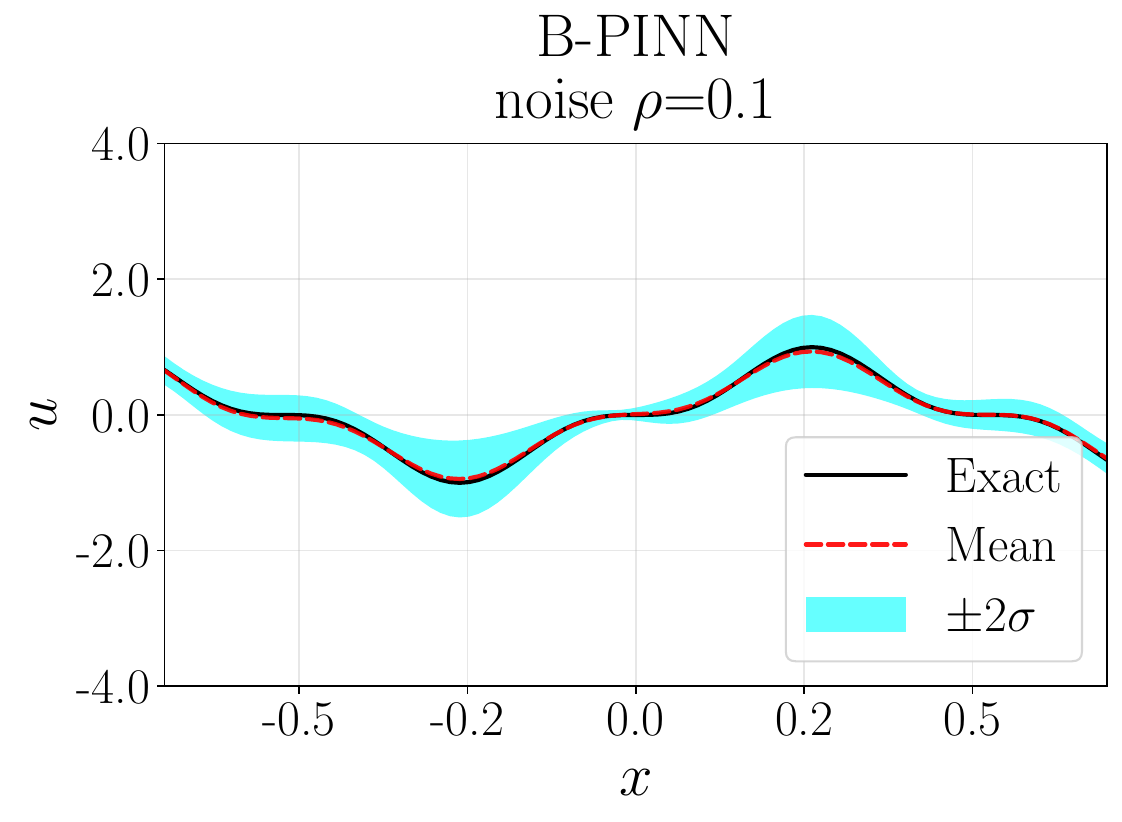}
        \caption{B-PINN}
    \end{subfigure}
    \caption{1D nonlinear Poisson (Gaussian noise, $\sigma = 0.10\,\lVert u\rVert_{\infty}$): mean and $\pm 2\sigma$ epistemic bands for $u(x)$. Methods: (a) E-PINN, (b) Dropout-PINN (5\%), (c) Dropout-PINN (10\%), (d) B-PINN.}
    \label{fig:app_nonlin_poisson_010}
\end{figure}
\clearpage

\clearpage
\subsection*{Additional 1D porous medium figures}
\begin{figure}[H]
    \centering
    \begin{subfigure}[b]{0.45\textwidth}
        \centering
        \incfig[width=\textwidth]{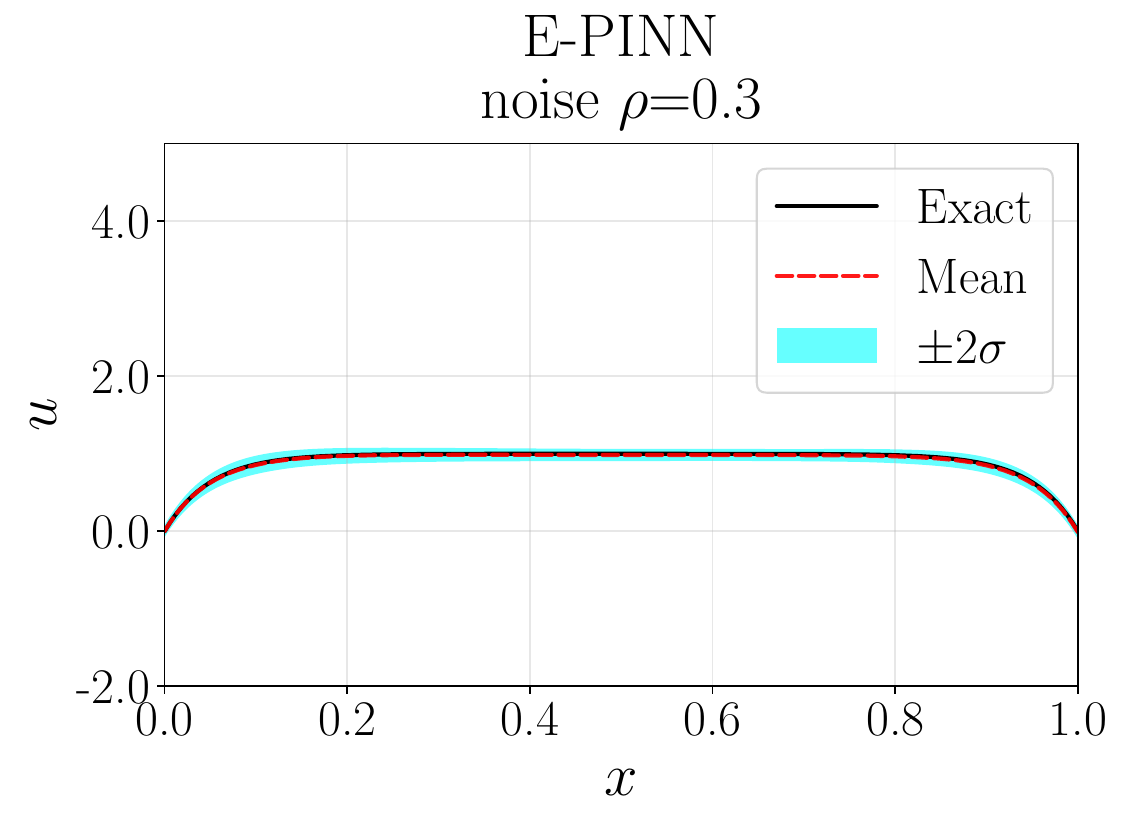}
        \caption{E-PINN}
    \end{subfigure}
    \hfill
    \begin{subfigure}[b]{0.45\textwidth}
        \centering
        \incfig[width=\textwidth]{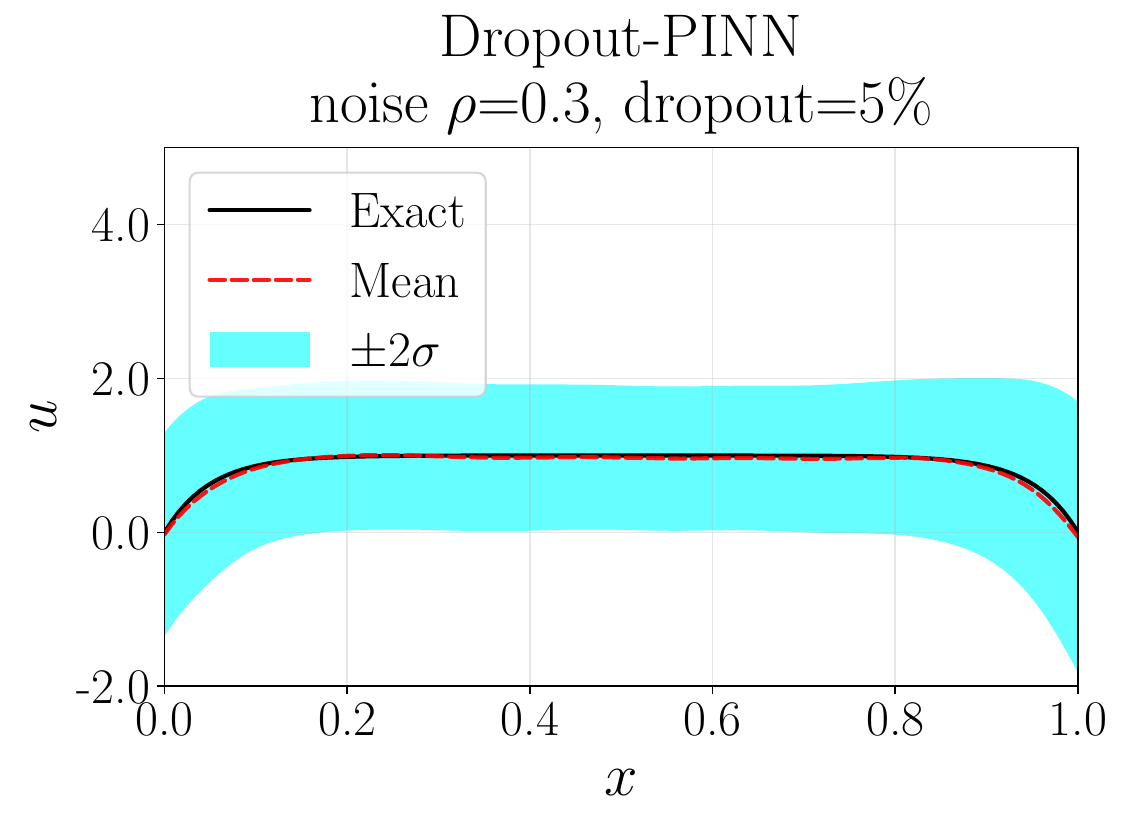}
        \caption{Dropout-PINN (5\%)}
    \end{subfigure}
    \begin{subfigure}[b]{0.45\textwidth}
        \centering
        \incfig[width=\textwidth]{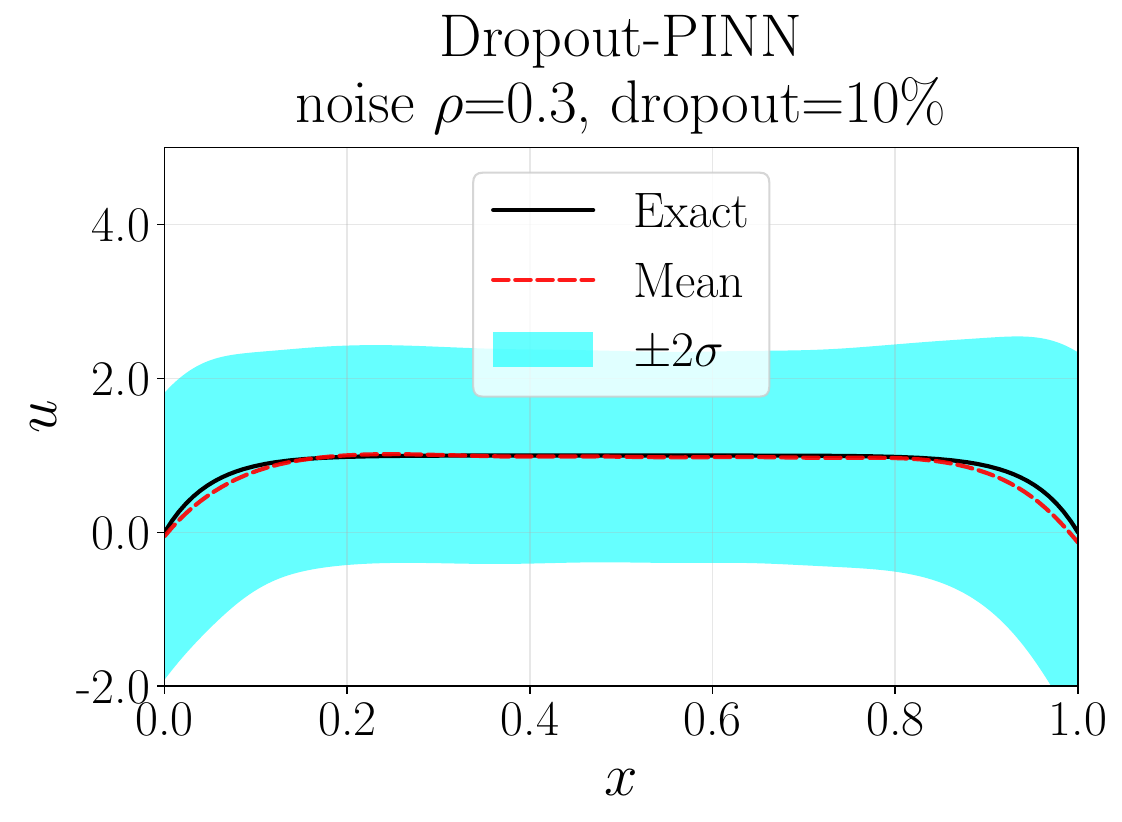}
        \caption{Dropout-PINN (10\%)}
    \end{subfigure}
    \hfill
    \begin{subfigure}[b]{0.45\textwidth}
        \centering
        \incfig[width=\textwidth]{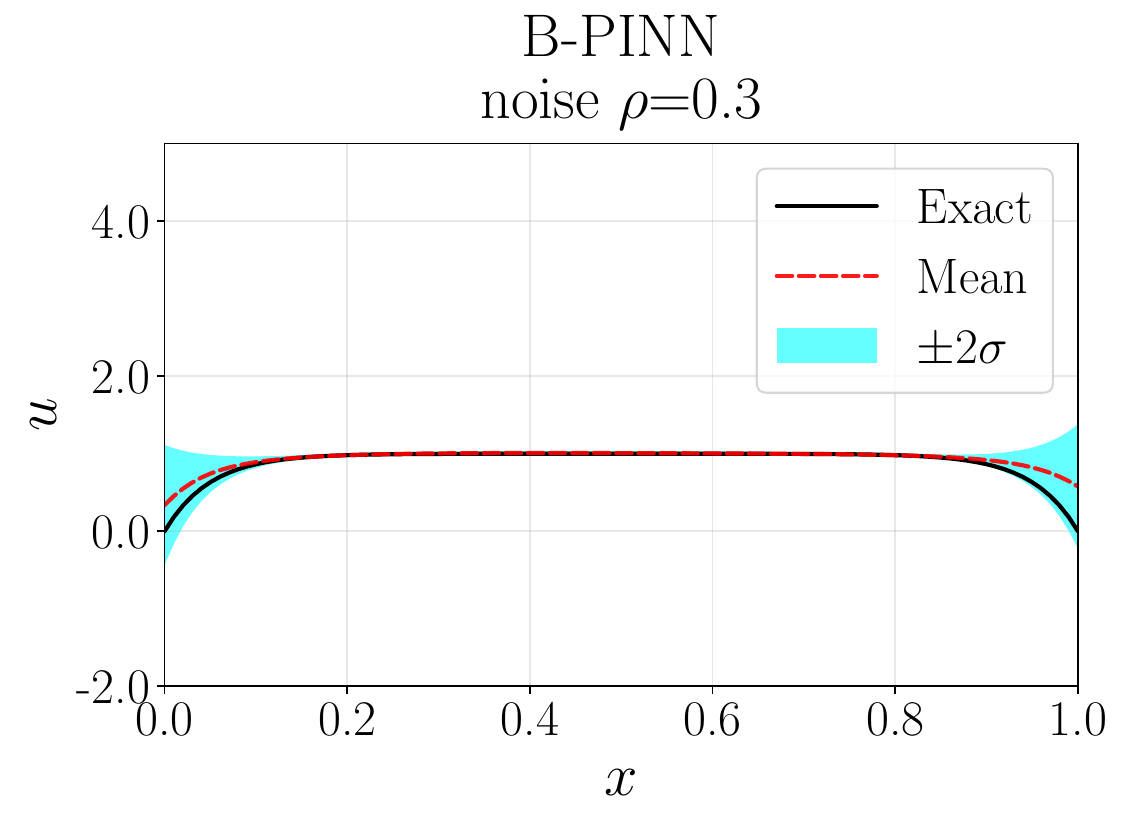}
        \caption{B-PINN}
    \end{subfigure}
    \caption{1D porous medium (Gaussian noise, $\sigma = 0.30\,\lVert u\rVert_{\infty}$): mean and $\pm 2\sigma$ epistemic bands for $u(x)$. Methods: (a) E-PINN, (b) Dropout-PINN (5\%), (c) Dropout-PINN (10\%), (d) B-PINN.}
    \label{fig:app_porous_030}
\end{figure}
\clearpage

\clearpage
\subsection*{Additional 1D heat equation figures}
\begin{figure}[H]
    \centering
    \begin{subfigure}[b]{0.32\textwidth}
        \centering
        \incfig[width=\textwidth]{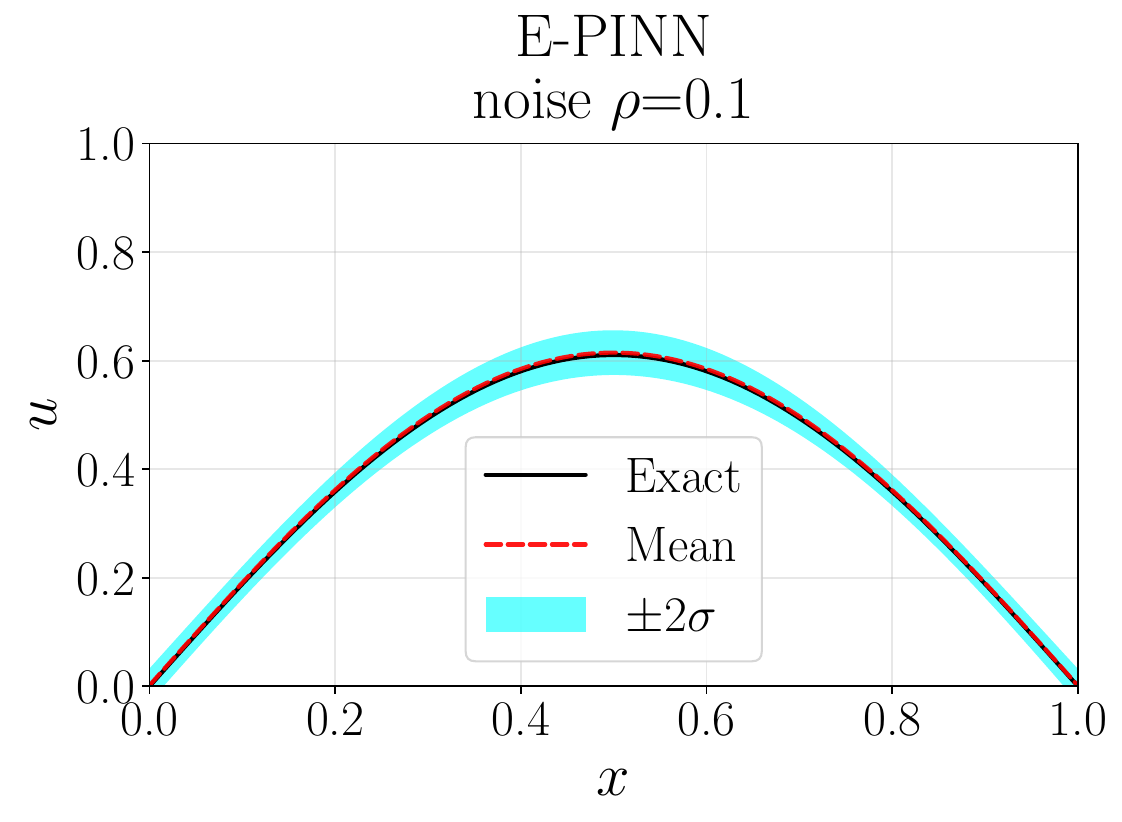}
        \caption{E-PINN}
    \end{subfigure}\hfill
    \begin{subfigure}[b]{0.32\textwidth}
        \centering
        \incfig[width=\textwidth]{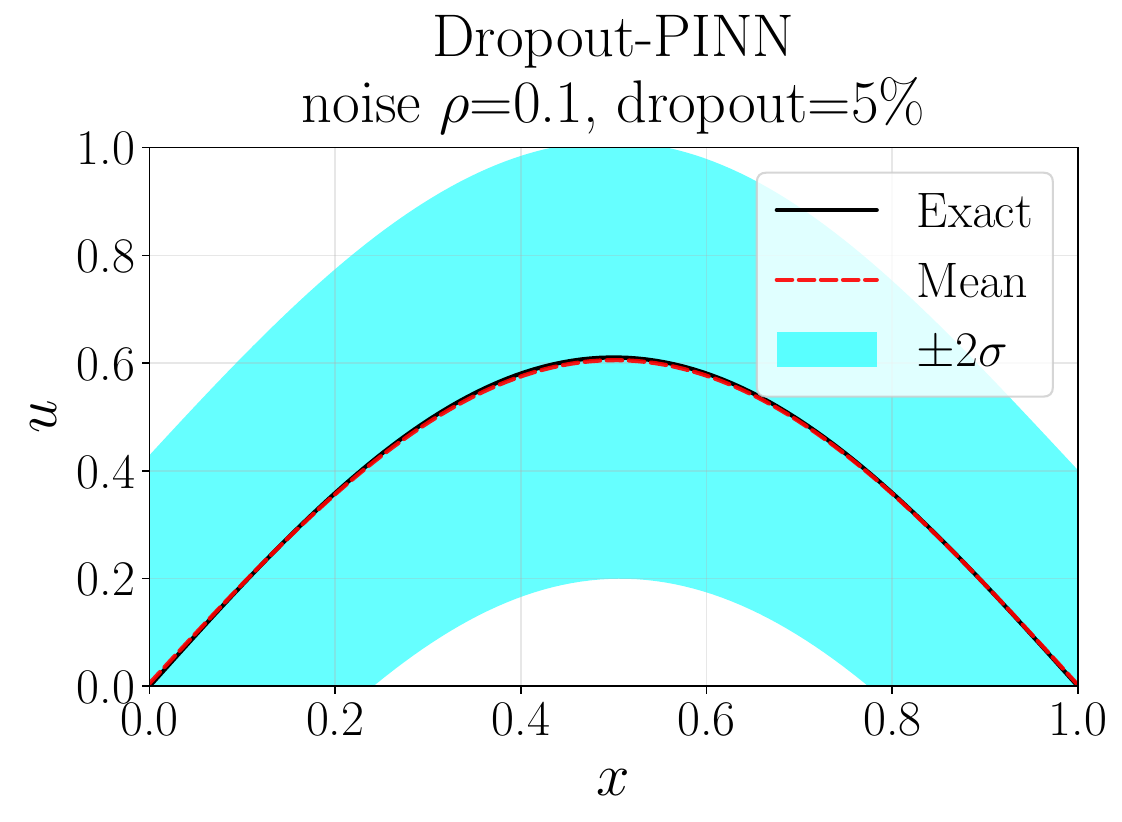}
        \caption{Dropout-PINN (5\%)}
    \end{subfigure}\hfill
    \begin{subfigure}[b]{0.32\textwidth}
        \centering
        \incfig[width=\textwidth]{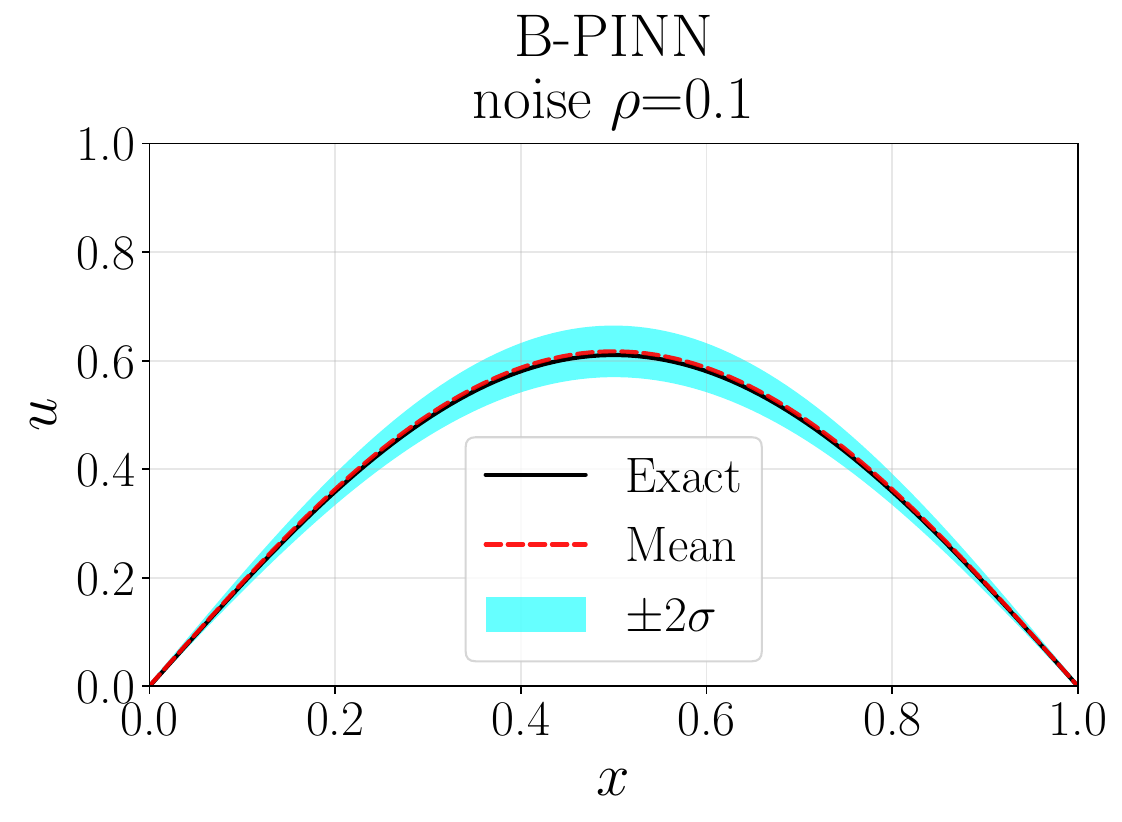}
        \caption{B-PINN}
    \end{subfigure}
    \caption{1D heat equation (inverse; $\rho=0.10$, $t=0.50$): mean and $\pm 2\sigma$ epistemic bands for $u(x,t)$ (instantaneous slice). Methods: (a) E-PINN, (b) Dropout-PINN (5\%), (c) B-PINN.}
    \label{fig:app_heat_t050_010}
\end{figure}
\clearpage

\begin{figure}[H]
    \centering
    \begin{subfigure}[b]{0.32\textwidth}
        \centering
        \incfig[width=\textwidth]{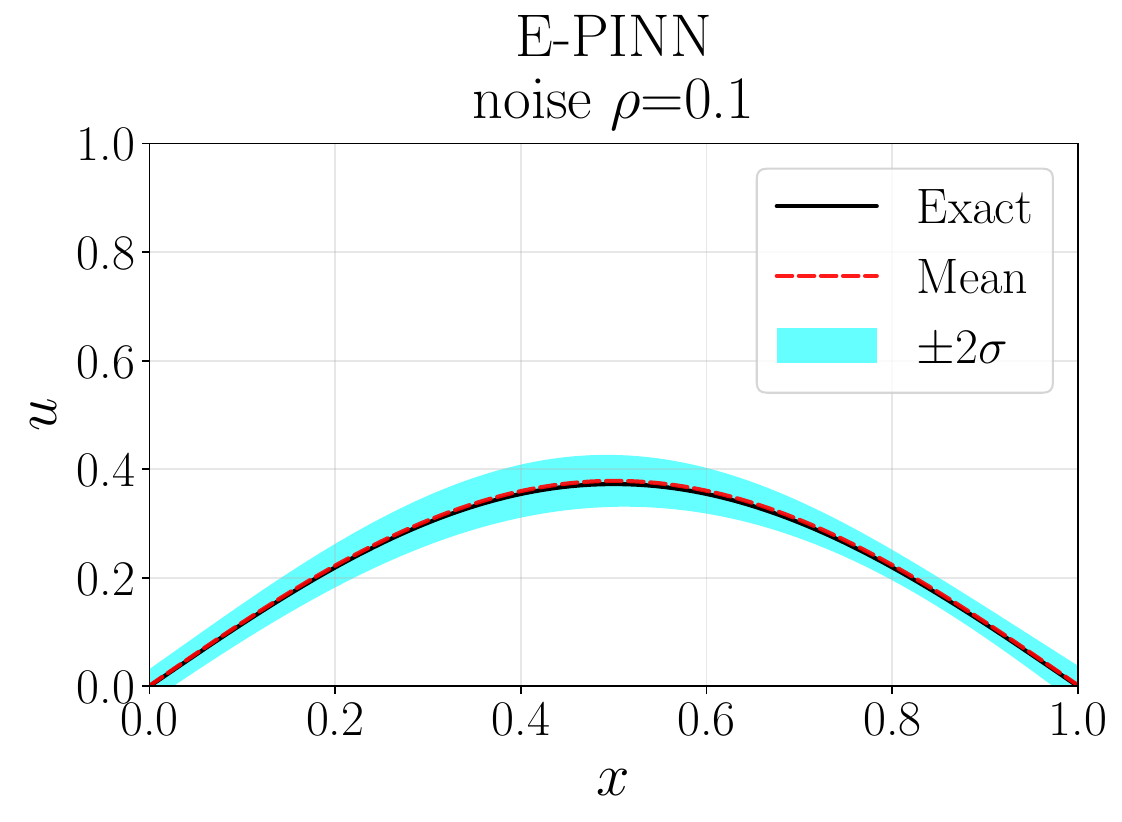}
        \caption{E-PINN}
    \end{subfigure}\hfill
    \begin{subfigure}[b]{0.32\textwidth}
        \centering
        \incfig[width=\textwidth]{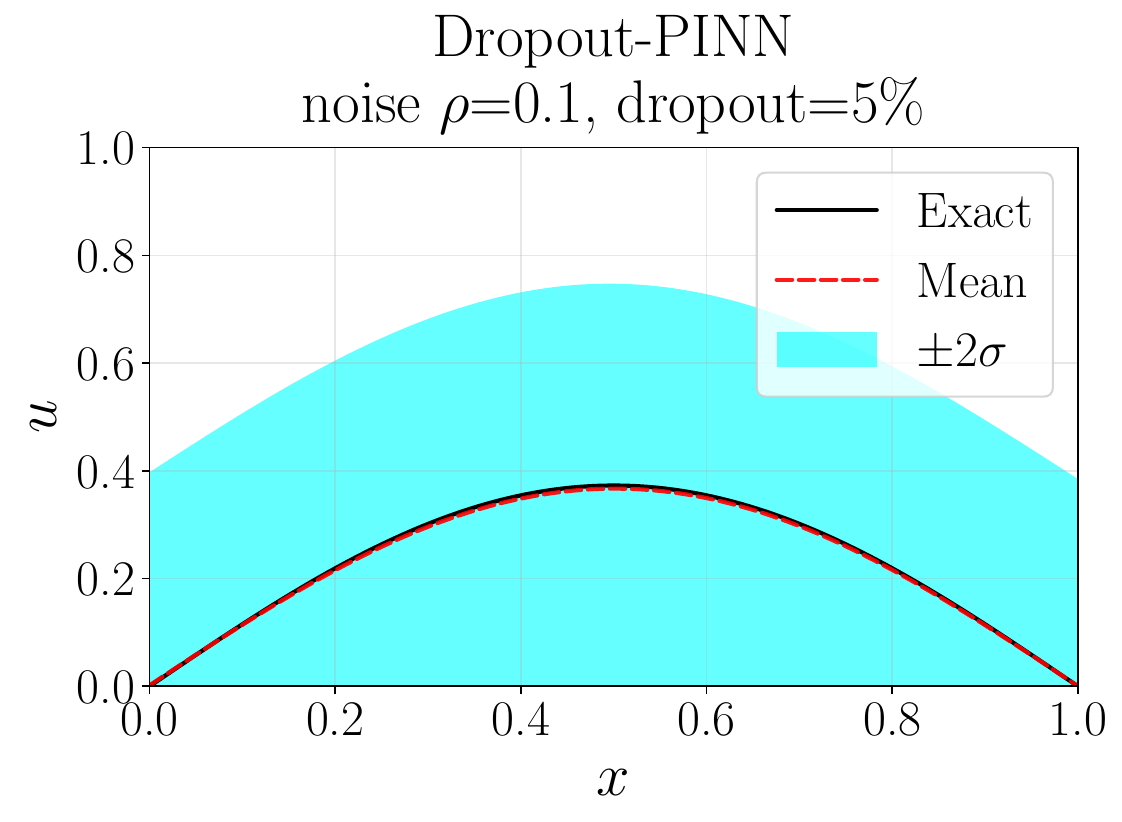}
        \caption{Dropout-PINN (5\%)}
    \end{subfigure}\hfill
    \begin{subfigure}[b]{0.32\textwidth}
        \centering
        \incfig[width=\textwidth]{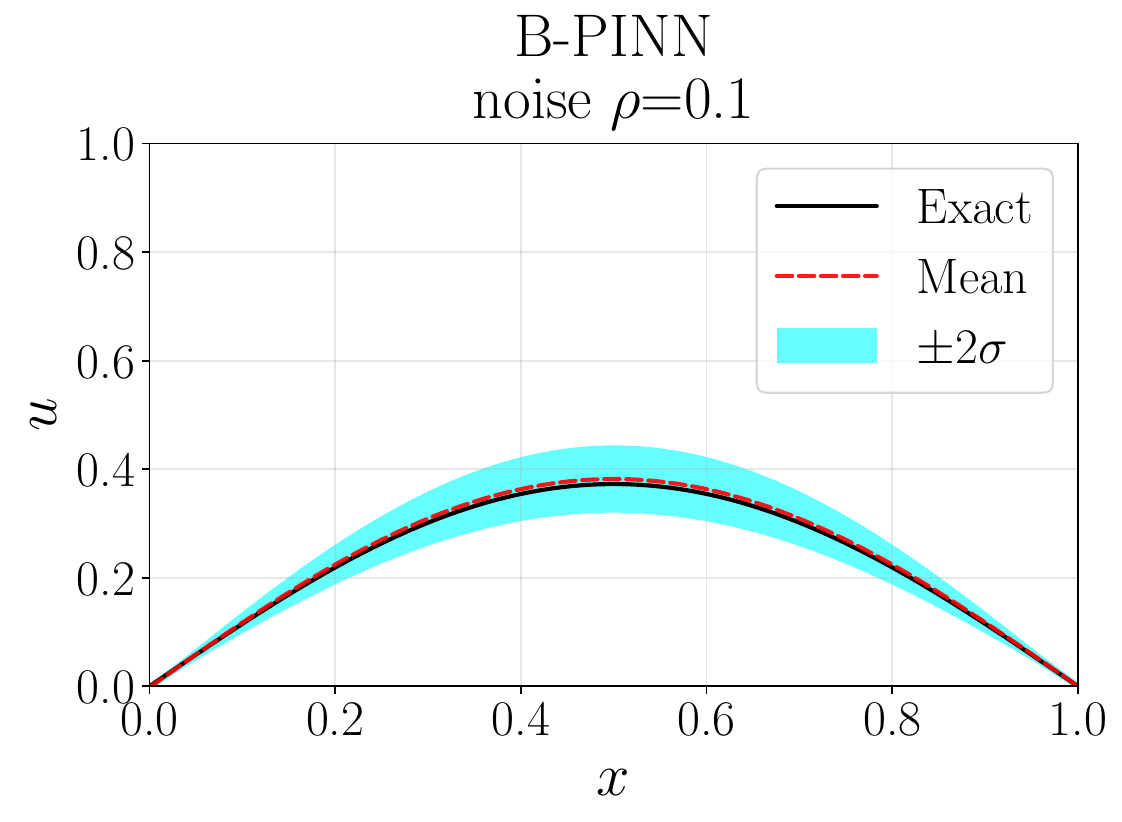}
        \caption{B-PINN}
    \end{subfigure}
    \caption{1D heat equation (inverse; $\rho=0.10$, $t=1.00$): mean and $\pm 2\sigma$ epistemic bands for $u(x,t)$ (instantaneous slice). Methods: (a) E-PINN, (b) Dropout-PINN (5\%), (c) B-PINN.}
    \label{fig:app_heat_t100_010}
\end{figure}
\clearpage

\bibliographystyle{unsrtnat}
\bibliography{refs.bib}
\end{document}